%% file: clean.tex
\documentclass{article}

\usepackage{PRIMEarxiv}

\usepackage[utf8]{inputenc} 
\usepackage[T1]{fontenc}    
\usepackage{hyperref}       
\usepackage{url}            
\usepackage{booktabs}       
\usepackage{amsfonts}       
\usepackage{nicefrac}       
\usepackage{microtype}      
\usepackage{lipsum}
\usepackage{lscape}
\usepackage{longtable}
\usepackage{booktabs}
\usepackage{fancyhdr}       
\usepackage{colortbl}
\usepackage{makecell}
\usepackage[dvipsnames]{xcolor}
\usepackage{multirow}
\usepackage[acronym]{glossaries}
\usepackage[nonumberlist]{glossaries-extra}
\makeglossaries
\setabbreviationstyle[acronym]{long-short}
\input{Abbreviations}

\usepackage{graphicx}       
\graphicspath{{media/}}     
\usepackage{ltcaption}
\usepackage{pdflscape}  
\usepackage{pgfplots}
\usepackage{filecontents}
\usepackage{pgfcalendar}
\usepackage{datetime}
\pgfplotsset{compat=1.17}
\usetikzlibrary{patterns, shapes.arrows, decorations.pathreplacing}

\title{A Comprehensive Evaluation of Large Language Models on Mental Illnesses}

\author{ Abdelrahman Hanafi$^\dagger$, 
Mohammed Saad$^\dagger$, Noureldin Zahran$^\dagger$, Radwa J. Hanafy$^\dagger,\ddagger$ and  Mohammed E. Fouda$^\dagger$ \\
$\dagger$Compumacy for Artificial Intelligence solutions, Cairo, Egypt.\\
$\ddagger$ Department of Behavioural Health- Saint Elizabeths Hospital, Washington DC, 20032.\\
fouda@compumacy.com
}

\begin{document}
\maketitle

\begin{abstract}
{ \glspl{LLM} have shown promise in various domains, including healthcare, with significant potential to transform mental health applications by enabling scalable and accessible solutions. This study aims to provide a comprehensive evaluation of 33 \gls{LLM}s, ranging from 2 billion to 405+ billion parameters, in performing key mental health tasks using social media data across six datasets. To our knowledge, this represents the largest-scale systematic evaluation of modern \glspl{LLM} for mental health applications. Models such as GPT-4, Llama 3, Claude, Gemma, Gemini, and Phi-3 were assessed for their  \gls{ZS} and \gls{FS} capabilities across three tasks: binary disorder detection, disorder severity evaluation, and psychiatric knowledge assessment. Key findings revealed that models like GPT-4 and Llama 3 exhibited superior performance in binary disorder detection, achieving accuracies up to 85\% on certain datasets, while \gls{FS} learning notably enhanced disorder severity evaluations, reducing the \gls{MAE} by 1.3 points for the Phi-3-mini model. Recent models, such as Llama 3.1 405b, demonstrated exceptional psychiatric knowledge assessment accuracy at 91.2\%, while prompt engineering played a crucial role in improving performance across tasks. However, the ethical constraints imposed by many \gls{LLM} providers limit their ability to respond to sensitive queries, hampering comprehensive performance evaluations. This work highlights both the capabilities and limitations of \gls{LLM}s in mental health contexts, offering valuable insights for future applications in psychiatry.
}
\end{abstract}


\section{Introduction}
\label{sec:introduction}
\gls{AI} is rapidly transforming the landscape of mental healthcare, offering innovative solutions to address the growing global burden of mental illness. According to the World Health Organization, mental disorders accounted for 5.1\% of the global disease burden in 2019, with depressive disorders affecting 280 million people worldwide, and an estimated 703,000 people {dying} by suicide \cite{world2022world}. The economic consequences of mental health conditions are also significant. In the United States, the cumulative cost of mental health inequities is projected to reach a staggering US\$14 trillion between 2024 and 2040. These costs include direct medical expenses, emergency department utilization, productivity losses, and premature deaths \cite{deloitte2024projected}.

The integration of \gls{AI} in psychiatry has enabled a wide array of applications, spanning from the crucial tasks of early detection and diagnosis to predicting treatment outcomes and providing therapeutic interventions. \gls{AI} models have been employed to analyze diverse data modalities, including neuroimaging data such as \gls{fMRI} and \gls{EEG}, physiological signals like \gls{ECG}, genetic information, and demographic data \cite{chen2022modern}. 

The advent of \gls{NLP} in mental health research opened up new possibilities. Early \gls{NLP} applications focused on analyzing text from clinical interviews and self-report questionnaires. These applications used techniques like the Linguistic Inquiry and Word Count to quantify linguistic patterns associated with various mental health conditions \cite{le2021machine}. The emergence of transformer-based models, such as \gls{BERT} \cite{devlin2018bert}, marked a significant turning point in \gls{NLP} for mental health. These models, trained on massive amounts of text data, can automatically learn complex linguistic representations, eliminating the need for manual feature engineering. This has led to substantial improvements in tasks like sentiment analysis, suicide risk assessment, and the detection of mental health conditions from social media posts \cite{ji2021mentalbert}.

The advancement to \gls{LLM}s marks a significant leap forward in the field of \gls{NLP}. \gls{LLM}s, such as OpenAI's GPT-4 \cite{achiam2023gpt}, Google's Gemini \cite{team2023gemini, reid2024gemini}, Meta's Llama \cite{touvron2023Llama}, and many others have been trained on vast corpora of text data from diverse sources across the internet, encompassing billions of words and a wide range of human knowledge. This extensive training enables \gls{LLM}s to understand and generate human-like text with a remarkable degree of fluency and coherence.

Given their extensive training, \gls{LLM}s have immense potential in the realm of mental health and psychiatry. They can be employed for initial consultations, providing patients with preliminary information about mental health disorders and guiding them toward appropriate care pathways. \gls{LLM}s can quickly analyze transcripts from patient sessions, offering initial insights and supporting medical practitioners in making more informed decisions. Their ability to process large volumes of patient history and previous medication data can streamline administrative tasks, ensuring that practitioners have quick access to comprehensive patient profiles. Additionally, \gls{LLM}s can serve as virtual counselors, offering support and guidance to individuals experiencing poor mental states, thus extending the reach of mental health services to those who may not have immediate access to professional help. These models can also assist in monitoring social media for signs of mental distress, enabling timely interventions and potentially saving lives. As these technologies continue to evolve, their integration into mental healthcare promises to enhance the efficiency, accessibility, and quality of care provided to patients \cite{obradovich2024opportunities}.

{ Previous research on \gls{LLM}s in mental health has explored five primary directions (as detailed in Section \ref{sec:related_work}): 1) Initial evaluations of GPT-3.5-Turbo's capabilities in detecting depression (86\% F1) and suicide risk (37\% F1) from social media \cite{lamichhane2023evaluation, amin2023will}, 2) Fine-tuning smaller models like Llama 2 and FLAN-T5 to match specialized architectures such as Mental-RoBERTa \cite{yang2024mentaLlama, xu2024mental}, 3) Developing chatbots for patient simulations \cite{chen2023llm} and systems integrating clinical guidelines like \gls{DSM-V} \cite{qin2023read}, 4) Creating evaluation benchmarks like PsyEval for standardized testing \cite{jin2023psyeval}, and 5) Literature reviews synthesizing ethical and practical considerations \cite{obradovich2024opportunities, hua2024large}. While these studies demonstrate progress, recent analyses note most focus on older models like GPT-3.5 rather than modern architectures, with limited transparency in prompt engineering methodologies \cite{omar2024applications}.}

{
While the potential of \gls{LLM}s in mental health care is evident, several research gaps remain to be addressed. Firstly, existing evaluations of \gls{LLM}s in mental health applications are limited in scope, with most studies testing only few models. This narrow focus restricts understanding of how performance varies across architectures, scales, and training approaches, making it difficult to determine which designs are most effective for real-world clinical use. Secondly, there is a lack of focus on prompt engineering to investigate the effect of different prompts on disorder detection tasks. While a few studies have touched upon this aspect, a more systematic and comprehensive exploration of prompt engineering techniques is needed to optimize \gls{LLM} performance in this domain. Thirdly, many studies have overlooked smaller yet important factors such as \gls{LLM} variation across runs and the effect of small prompt modifications across models. Additionally, there is often a lack of full transparency in reporting methodologies, even when code is provided, hindering reproducibility and further research.

}
\vspace{0.5em}

{
To address these limitations, our study makes the following advances:
\begin{itemize}
\item \textbf{Comprehensive Model Evaluation:} First study to evaluate 33 cutting-edge \gls{LLM}s (including GPT-4o, Llama 3.1 405B, Claude 3.5, and Phi-3) across parameter scales from 2B to 405B+ - significantly expanding beyond prior work limited to GPT-3.5/4 and Llama 2.

\item \textbf{Extensive Prompt Engineering:} Systematic analysis of \gls{ZS} vs. \gls{FS} performance with severity scoring templates, demonstrating \gls{MAE} reductions up to 1.3 points through prompt optimization (e.g., in Phi-3-mini).

\item \textbf{Methodology Transparency:} Full disclosure of all prompt templates used across 6 datasets with temperature/seed controls - enabling exact reproducibility missing in previous studies.

\item \textbf{Dataset Investigation:} Comprehensive investigation of published human-annotated mental health datasets, both private and public, ensuring the quality and reliability of data used in experiments by prioritizing datasets labeled by experts or trained volunteers.

\item \textbf{Practical Challenges Identification:} Identified key barriers including high costs (GPT-4/Claude), \gls{API} restrictions (Gemini), and prompt sensitivity issues. Suggested future approaches: CoT prompting, RAG integration, and fine-tuning on clinical data.

\end{itemize}
}

{

This paper is organized as follows: \textbf{Section \ref{sec:related_work}} discusses the related work and usage of \gls{LLM}s in the mental illness domain. \textbf{Section \ref{sec:methedology}} outlines the methodology employed in this study, including the experimental design, the datasets and models used, and the creation of prompt templates. \textbf{Section \ref{sec:discussion}} presents the results of our experiments, analyzing the performance of various \gls{LLM}s on three key tasks: binary disorder classification, disorder severity evaluation, and psychiatric knowledge assessment. This section begins with an examination of performance variability across models. \textbf{Section \ref{sec:implications}} discusses the broader implications of these findings, focusing on model selection considerations, the role of prompt engineering, and the balance between model safety and accuracy in mental health applications. \textbf{Section \ref{sec:additional_investigations}} details supplementary experiments conducted alongside the main analysis, such as exploring the impact of repeating instructions in prompts and a knowledge reinforcement approach. \textbf{Section \ref{sec:limitations}} acknowledges the limitations of the current study, including dataset constraints and model-related challenges. \textbf{Section \ref{sec:future}} suggests potential directions for future research in this area, such as exploring different model architectures and expanding to other mental health tasks. \textbf{Section \ref{sec:conclusion}} concludes the paper. 
}

\section{Related Works}
\label{sec:related_work}

Research on \gls{LLM}s in mental health has rapidly expanded across several directions. Current work in the literature focuses on five main areas: basic evaluation of existing models (Sec. \ref{subsec:evaluating}), performance enhancement through fine-tuning (Sec. \ref{subsec:fine_tuning}), development of practical applications including data augmentation and chatbots (Sec. \ref{subsec:data_augmentation}), creation of standardized benchmarks (Sec. \ref{subsec:benchmarks}), and systematic literature reviews (Sec. \ref{subsec:surveys}). While these studies demonstrate promising results, they also highlight the need for extensive validation before clinical deployment.

\subsection{Evaluating \gls{LLM}s on Mental Health Tasks}
\label{subsec:evaluating}

One of the earliest works in this domain is the study by \cite{lamichhane2023evaluation}, which presents a simple evaluation of GPT-3.5-Turbo's \gls{ZS} binary classification performance on stress, depression, and suicidality detection tasks using social media posts. The resulting F1 scores (73\% for stress, 86\% for depression, and 37\% for suicide), were particularly bad in the suicidality severity prediction task. The authors of \cite{amin2023will} extend this evaluation by assessing GPT-3.5-Turbo's text classification abilities on a broader range of affective computing problems, including big-five personality prediction and sentiment analysis, in addition to suicide tendency detection. Their findings reveal that while GPT-3.5-Turbo provided decent results, comparable to Word2Vec and bag-of-words baselines, it did not outperform the fine-tuned RoBERTa model on the given tasks. However, the authors did not prompt GPT-3.5-Turbo through an \gls{API}, they instead did it manually using the web interface, only resetting the context every 25 prompts, which might have affected their results. 

\subsection{Fine-tuning \gls{LLM}s for mental health tasks}
\label{subsec:fine_tuning}

Building upon the evaluation studies, \cite{yang2024mentaLlama} and \cite{xu2024mental} explore fine-tuning \gls{LLM}s for improved performance in mental health tasks. \cite{yang2024mentaLlama} introduce the MentaLlama model, which is fine-tuned on the IMHI dataset, a large dataset of social media posts labeled with mental health conditions and explanations. The authors created the IMHI dataset by taking multiple existing mental health datasets and generating additional explanatory labels using GPT-3.5-Turbo. The MentaLlama-chat-13B model surpasses or approaches state-of-the-art discriminative methods in prediction correctness on 7 out of 10 test sets and generates explanations on par with GPT-3.5-Turbo thus providing interpretability. 

The authors of \cite{xu2024mental} conducted a comprehensive evaluation of multiple \gls{LLM}s, including Alpaca, Alpaca-LoRA, FLAN-T5, Llama 2, GPT-3.5-Turbo, and GPT-4, on various mental health prediction tasks using online text data. Their work explored \gls{ZS} prompting, \gls{FS} prompting, and instruction fine-tuning techniques. The authors fine-tuned Alpaca and FLAN-T5 on six mental health prediction tasks across four datasets from Reddit. Results show that instruction fine-tuning significantly improves the performance of \gls{LLM}s on these tasks. The fine-tuned models, Mental-Alpaca and Mental-FLAN-T5, outperformed much larger models like GPT-3.5-Turbo and GPT-4 on specific tasks, and performed on par with the state-of-the-art task-specific model Mental-RoBERTa \cite{ji2021mentalbert}. The study also includes an exploratory case study on these models’ reasoning capability, which further suggests both the promising future and the important limitations of \gls{LLM}s.

\subsection{\gls{LLM}s for Data Augmentation and Chatbot Development}
\label{subsec:data_augmentation}

\cite{qin2023read}, \cite{bucur2023utilizing}, and \cite{chen2023llm} explore the application of \gls{LLM}s beyond prediction tasks, focusing on data augmentation and chatbot development. \cite{qin2023read} { Developed} Chat-Diagnose, an explainable and interactive \gls{LLM}-augmented system for depression detection in social media. Chat-Diagnose incorporates a tweet selector to filter excessive posts, an image descriptor to convert images into text, and professional diagnostic criteria (\gls{DSM-V}) to guide the diagnostic process. The system utilizes the \gls{CoT} technique to provide explanations and diagnostic evidence. In the full-training setting, it leverages answer heuristics from a traditional depression detection model. Chat-Diagnose achieves \gls{SoTA} performance in various settings, including \gls{ZS}, \gls{FS}, and full-training scenarios.

\cite{bucur2023utilizing} investigates the use of GPT-3.5-Turbo for generating synthetic Reddit posts simulating depression symptoms from the \gls{BDI}-II questionnaire, aiming to enhance semantic search capabilities. The study finds that using original \gls{BDI}-II responses as queries is more effective, suggesting that the generated data might be too specific for this task.

\cite{chen2023llm} focuses on developing chatbots that simulate psychiatrists and patients in clinical diagnosis scenarios, specifically for depressive disorders. The study follows a three-phase design, involving collaboration with psychiatrists, experimental studies, and evaluations with real psychiatrists and patients. The development process emphasizes an iterative refinement of prompt design and evaluation metrics based on feedback from both psychiatrists and patients. A key finding of the study is the comparison between real and simulated psychiatrists, revealing differences in questioning strategies and empathy behaviors.

\subsection{Benchmarks for Evaluating \gls{LLM}s in Psychiatry}
\label{subsec:benchmarks}

In contrast to the aforementioned studies, the authors of \cite{jin2023psyeval} took a different approach. Instead of evaluating existing \gls{LLM}s on established datasets, they sought to create a comprehensive benchmark specifically designed to assess the capabilities of \gls{LLM}s in the mental health domain. PsyEval (the benchmark) includes tasks such as mental health question-answering, where \gls{LLM}s are evaluated on their ability to provide accurate and informative responses to queries related to mental health conditions, symptoms, and treatments. The benchmark also features tasks focused on diagnosis prediction, both using data from social media or simulated {dialogues} between patients and clinicians. Additionally, PsyEval incorporates tasks that assess \gls{LLM}s' ability to provide empathetic and safe psychological counseling in simulated conversations.

\subsection{Literature Reviews}
\label{subsec:surveys}
In addition to the contribution papers mentioned above, several surveys have been conducted to summarize and analyze the landscape of \gls{LLM}s in mental health care. The authors of \cite{obradovich2024opportunities} provide a comprehensive overview of the opportunities and risks associated with \gls{LLM}s in psychiatry, discussing their potential to enhance mental health care through improved diagnostic accuracy, personalized care, and streamlined administrative processes. The authors highlight the ability of \gls{LLM}s to efficiently analyze patient data, summarize therapy sessions, and aid in complex diagnostic problem-solving. They also discuss the potential of \gls{LLM}s to automate certain managerial decisions within hospital systems, such as personnel schedules and equipment needs. However, the authors also acknowledge the risks associated with \gls{LLM}s, such as labor substitution, the potential for reduced human-to-human socialization, and the amplification of existing biases. They conclude by advocating for the development of pragmatic frameworks, including red-teaming and multi-stakeholder-oriented safety and ethical guidelines, to ensure the safe and responsible deployment of \gls{LLM}s in psychiatry.

The authors of \cite{omar2024applications} conducted a systematic review of the applications of \gls{LLM}s, such as GPT-3.5-Turbo, in the field of psychiatry. The review identified 16 studies that directly examined the use of \gls{LLM}s in psychiatry, with applications ranging from clinical reasoning and social media analysis to education. The authors found that \gls{LLM}s like GPT-3.5-Turbo and GPT-4 showed promise in assisting with tasks such as diagnosing mental health issues, managing depression, evaluating suicide risk, and supporting education in the field. However, the review also highlighted limitations, such as difficulties with complex cases and potential underestimation of suicide risks. 

The article by \cite{hua2024large} presents a scoping review of \gls{LLM}s in mental health care, focusing on their applications and outcomes. The review identified 34 publications that met their inclusion criteria, including studies compiling new datasets, studies focused on model development, studies focused on employing and evaluating \gls{LLM}s, and studies examining the potential opportunities, risks and ethical considerations associated with utilizing \gls{LLM}s in a psychiatric context. The authors provide a detailed overview of the models and datasets used in these studies. They found that most studies utilized GPT-3.5 or GPT-4, with some fine-tuning on mental health-specific datasets like MentaLlama and Mental-LLM. The datasets used in these studies were primarily sourced from social media platforms and online mental health forums, with a few utilizing clinical data or simulated conversations. The authors also discuss the key challenges associated with \gls{LLM}s, such as data availability and reliability, effective handling of mental states, and appropriate evaluation methods.

\section{Methodology Implementation}
\label{sec:methedology}

To investigate the potential of \gls{LLM}s in mental health tasks, we designed a series of experiments to evaluate their performance across various scenarios. This section details our experimental setup, including the chosen datasets, models, prompting strategies, and evaluation metrics.

\subsection{Experimental Setup}
\label{subsec:task_formulation}

{To ensure the robustness and interpretability of our results, we first conducted a performance variability experiment to evaluate the inherent performance fluctuations of each \gls{LLM}. In this experiment, we assessed the inherent performance variability of each \gls{LLM} on a representative task. The experiment involved repeatedly evaluating each model on the same dataset (DEPTWEET) five times using consistent, standard parameters (e.g., temperature 0). This approach allows us to approximate the natural performance fluctuations and establish a baseline for judging whether observed performance enhancements in the main tasks are meaningful improvements or simply due to random variability. The experiment was conducted on 1000 fairly sampled instances from the DEPTWEET dataset.}

Following this initial variability assessment, our experiments were designed to assess \gls{LLM} capabilities across three primary tasks: binary disorder detection, disorder severity evaluation, and psychiatric knowledge assessment. We conducted \gls{ZS} experiments for all three tasks, where the \gls{LLM}s were given only the input data (social media post or multiple-choice question) and the task description, without any prior examples. For the disorder severity prediction task, we also conducted \gls{FS} experiments, where the \gls{LLM}s were provided with three examples of posts and their corresponding severity ratings before being asked to predict the severity of a new post.

\subsubsection{Task 1: \gls{ZS} Binary Disorder Detection}

\gls{ZS} learning is a machine learning paradigm where the model makes predictions without having seen any labeled examples of the specific task during training. Instead, the model relies on its pre-existing knowledge and the task description to make inferences. This approach is particularly useful when labeled data is scarce or when deploying models to new tasks quickly.

In this analysis, \gls{ZS} learning was employed to evaluate the \gls{LLM}'s inherent understanding of psychiatric disorders. Each \gls{LLM} was tasked with determining whether a social media user exhibits a specific mental disorder based solely on their post and a brief task description. We focused on three prevalent disorders: depression, suicide risk, and stress. This setup allowed us to assess the model's ability to identify these disorders based on its pre-trained knowledge and general language understanding.

{ \subsubsection{Task 2: \gls{FS} Disorder Severity Evaluation}}

\gls{FS} learning, on the other hand, is a technique where the model is provided with a small number of labeled examples before being tasked with making predictions. This approach helps the model to better understand the context and complexities of the task.

This task involved evaluating the severity (e.g. from 0 to 4) of a specific mental disorder in a user solely from a social media post written by them. We evaluated \gls{LLM}s on depression and suicide risk, employing both \gls{ZS} and \gls{FS} approaches. In the \gls{ZS} scenario, the \gls{LLM} received only the social media post and the task description. In the \gls{FS} scenario, the \gls{LLM} was provided with three examples of posts from other users with corresponding severity ratings before assessing the new post. By comparing the results of these two approaches, we aimed to determine how much additional context in the form of examples improved the model's ability to accurately gauge disorder severity.

\gls{FS} prompting was only employed for severity evaluation because it is a more complex and challenging task compared to binary disorder detection, thereby providing a better demonstration of the potential improvement gained through \gls{FS} learning.

\subsubsection{Task 3:  Psychiatric Knowledge Assessment}

This task was designed to test the \gls{LLM}'s knowledge of basic psychiatric concepts. The \gls{LLM} was presented with multiple-choice questions related to psychiatry. This task evaluates the \gls{LLM}'s foundational understanding of psychiatric concepts and its ability to provide accurate factual information typically found in textbooks.

{
\subsubsection{Initial Exploration of \gls{LLM} fine-tuning}

While this study primarily focuses on evaluating the \gls{ZS} and \gls{FS} capabilities of \glspl{LLM} in their pre-trained state, fine-tuning presents a complementary approach to further adapt these models for specific mental health tasks. To gain initial insights into this, we conducted preliminary fine-tuning experiments using GPT-4o mini and Phi-3.5-mini, employing parameter-efficient techniques like LoRA (Low-Rank Adaptation) for resource-conscious adaptation. For Phi-3.5-mini, we specifically utilized LoRA in 4-bit quantization (Q4) to further enhance efficiency.

In these preliminary fine-tuning experiments, we specifically fine-tuned both \glspl{LLM} for Task 2 (Disorder Severity Assessment) using the four severity-annotated datasets described in Section \ref{subsec:datasets} (DepSeverity, DEPTWEET, RED SAM, and SAD). For GPT-4o-mini, fine-tuning was also initially explored on a 1600-sample balanced subset of the DEPTWEET dataset to assess in-domain performance.  It is important to note that Phi-3.5-mini was released after the main zero-shot and few-shot experiments of this study were completed\footnote{Consequently, Phi-3.5-mini is not included in the main zero-shot and few-shot analysis presented in Sections \ref{sec:discussion}.}. For all fine-tuning processes, we utilized fair random sampling on these datasets, adhering to the sampling strategy outlined in Section \ref{subsec:sampling}. We present and discuss the results of these initial fine-tuning experiments in Section \ref{subsec:finetuning}.
}
\subsection{Datasets}
\label{subsec:datasets}

We deliberately avoided datasets with automated labeling methods, such as those collecting posts from specific mental health subreddits (e.g., r/Depression) and matching them to control posts from unrelated subreddits. Additionally, we steered clear of semi/weak-labeled datasets that include posts from users who self-reported a specific disorder using phrases like "I am depressed." These approaches tend to introduce some level of unreliability, as they lack human verification to determine if a specific post or user is exhibiting mental health-related issues.

Instead, we prioritized datasets labeled by experts or crowd-sourced workers/volunteers who were trained by experts. This ensures the quality and reliability of the data used for our experiments, providing a more accurate foundation for assessing the \gls{LLM}'s performance in detecting and evaluating mental health disorders.

We considered various mental disorders but ultimately focused on depression, suicide risk, and stress due to their prevalence and the relative availability of high-quality datasets. Other disorders were not included in this study due to challenges in data collection and annotation. The sensitivity of mental health data often limits the public availability of relevant social media posts, and stricter data usage agreements imposed by social media platforms\footnote{The landscape of social media data collection has changed significantly in recent years. In 2023, platforms like Reddit \cite{Reddit_API_Changes_2023} and X (formerly Twitter) \cite{Twitter_TOS_2023, Twitter_API_Changes_2023} introduced substantial costs for accessing their \gls{API}s, which were previously used to create many research datasets. Furthermore, stricter data usage agreements have been implemented, making some previously popular datasets, such as the popular SMHD \cite{cohan2018smhd}, unsharable due to updated terms of service.} have further restricted access to previously used datasets. Additionally, less prevalent disorders typically have fewer human-annotated datasets available, making it difficult to compile a sufficiently large and diverse dataset for robust model evaluation.

We summarize our dataset findings in Table \ref{tab:datasets}. While not exhaustive, this table includes datasets that met our criteria for quality and relevance. However, some datasets were not directly used in this study either due to time and resource constraints or due to inaccessibility.

While our focus was on disorder classification and evaluation, we acknowledge the existence of other relevant datasets that could evaluate \gls{LLM}s' performance on tasks like empathy, dialogue safety, emotional support provision, and counselling. 

{
Table \ref{tab:dataset_summary} summarizes the key characteristics of the datasets used in our experiments, including their clinical focus, task type, and scale. Below we provide detailed descriptions of these datasets, with particular attention to their annotation processes and label distributions.}

\begin{table}[ht]
\caption{Summary of all the datasets used in this study.}
\label{tab:dataset_summary}
\centering
\begin{tabular}{@{}lllll@{}}
\toprule
\textbf{Dataset} & \textbf{Disorder} & \textbf{Task Type} & \textbf{Original Size} & \textbf{Test Size} \\ 
\midrule
Dreaddit & Stress & Binary Classification & 3,553 & 715 \\
DepSeverity & Depression & Binary/Severity Classification & 3,553 & 1,000 \\
SDCNL & Suicide & Binary Classification & 1,895 & 379 \\
SAD & Stress & Binary/Severity Classification & 6,850 & 1,000 \\
DEPTWEET & Depression & Binary/Severity Classification & 40,191 & 1,000 \\
RED SAM & Depression & Binary/Severity Classification & 16,632 & 1,000 \\
MedMCQA & Psychiatry & Question Answering & 4,442 & 1,000 \\
\bottomrule
\end{tabular}
\par
\smallskip
\textbf{Note:}
Test sizes reflect actual samples used after applying our sampling strategy (Section \ref{subsec:sampling}).
\end{table}


{ Datasets containing severity metrics like DepSeverity and SAD were binarized by assigning a "False" label (0) to posts/users with the minimum severity score and a "True" label (1) to those with any higher severity score.}

\input{Summary_table}

The following subsections detail the composition and annotation characteristics of each dataset:

\subsubsection{Dreaddit \cite{turcan2019dreaddit}}
Dreaddit is a dataset for stress analysis collected from Reddit using the PRAW \gls{API}. It contains 187,444 posts from 10 subreddits across five domains: interpersonal conflict (abuse, social), mental illness (anxiety, \gls{PTSD}), and financial need (financial). For annotation, posts were divided into five-sentence segments, and crowdworkers on Amazon Mechanical Turk labeled them as "Stress," "Not Stress," or "Can't Tell." Segments with low agreement or labeled as "Can't Tell" were discarded, resulting in a final dataset of 3,553 labeled segments. The labels indicate whether the author of the segment is expressing stress and a negative attitude towards it.

\subsubsection{DepSeverity \cite{naseem2022early}}
DepSeverity is a dataset designed for the identification of depression severity levels in social media posts. It contains 3,553 Reddit posts labeled across four levels of depression severity: minimal, mild, moderate, and severe. The dataset was created by relabeling an existing binary-class dataset using the Depressive Disorder Annotation scheme, which is based on six clinical resources and aligns with common depression levels in the psychological domain.

The annotation process involved two professional annotators who independently labeled each post. In cases of disagreement, a third annotator was consulted, and the majority vote was used to assign the final label.

\subsubsection{SDCNL \cite{haque2021deep}}
SDCNL is a dataset designed for the classification of suicidal ideation versus depression in social media posts. It contains 1,895 posts from two Reddit subreddits: r/SuicideWatch and r/Depression. The dataset was created to address the lack of research on distinguishing between these two mental health conditions, which is crucial for providing appropriate interventions.

This is the only dataset in our study that was not labeled by humans. Instead, it employed an unsupervised approach to correct initial labels based on the subreddit of origin.

\subsubsection{SAD \cite{mauriello2021sad}}
SAD (Stress Annotated Dataset) is a dataset designed for the identification of daily stressors in SMS-like conversational systems. It contains 6,850 sentences labeled across nine categories of stressors, derived from stress management literature, live conversations from a prototype chatbot system, crowdsourced data, and targeted web scraping from LiveJournal. The dataset was created to improve chatbot interactions by enabling the identification and classification of stressors in text, facilitating appropriate interventions and empathetic responses. The stress categories include: Work (1043), School (1043), Financial Problem (152), Emotional Turmoil (238), Social Relationships (99), Family Issues (101), Health, Fatigue, or Physical Pain (103), Everyday Decision Making (101), and Other (123). { However, in this analysis, we focused on the level of stress severity without considering its source.}

The annotation process involved crowd workers on Amazon Mechanical Turk, who labeled sentences with stressor categories and severity ratings on a 10-point Likert scale, ranging from "not stressful" (1) to "extremely stressful" (10). The dataset includes metadata such as stressor ID, sentence text, data source, labels, severity ratings, and binary fields indicating stress and COVID-19 relation. The dataset is publicly available on GitHub for research and application development purposes.

\subsubsection{DEPTWEET \cite{kabir2023deptweet}}
DEPTWEET is a dataset designed for the detection of depression severity levels in social media texts. It contains 40,191 tweets labeled across four levels of depression severity: non-depressed, mild, moderate, and severe. The dataset was created by collecting tweets based on depression-related keywords and then having trained crowdworkers annotate them according to a typology based on the \gls{DSM-V} and \gls{PHQ}-9 clinical assessment tools.

The annotation process involved 111 crowdworkers who were pre-screened and trained on the labeling typology. Each tweet was annotated by at least three annotators, and the final label was determined by majority voting. A confidence score was also assigned to each label, reflecting the level of agreement among annotators. The dataset includes additional metadata such as tweet ID, reply count, retweet count, and like count.

 \subsubsection{MedMCQA \cite{pal2022medmcqa}}
 MedMCQA is a large-scale dataset designed for medical domain question answering, { {and in this study, it was specifically used for the third task, psychiatric knowledge assessment} (Section \ref{subsec:task_3})}. focusing on a wide range of medical subjects and topics relevant to medical entrance exams. It contains 194k high-quality multiple-choice questions (\gls{MCQ}s) covering 2.4k healthcare topics and 21 medical subjects, including a small portion (4442) related to psychiatry. Each sample in the dataset includes a question, correct answer(s), other options, and a detailed explanation of the solution. The dataset is designed to be challenging, requiring a deep understanding of both language and medical knowledge to answer the questions correctly. It is intended to serve as a benchmark for evaluating and improving the performance of AI models in medical question answering.

\subsubsection{RED SAM \cite{sampath2022data, kayalvizhi2022findings}}
RED SAM (Reddit Social Media Depression Levels) is a dataset designed for the detection of depression levels in social media posts. It comprises 16,632 Reddit posts labeled across three categories: "Not Depressed," "Moderately Depressed," and "Severely Depressed." The dataset was created to address the need for identifying not only the presence of depression but also its severity, which is crucial for effective treatment and intervention.

The dataset creation process involved collecting posts from relevant subreddits on mental health topics, such as "r/MentalHealth," "r/depression," "r/loneliness," "r/stress," and "r/anxiety." The collected posts were then annotated by two domain experts who followed specific guidelines to determine the level of depression reflected in each post. The annotation guidelines considered factors like the length of the post, the expression of emotions, and the presence of specific themes or language patterns.

\subsubsection{CSSRS-Suicide (Excluded) \cite{gaur2019reddit}}
{ CSSRS-Suicide} is a dataset designed for assessing the severity of suicide risk in individuals based on their social media posts. It contains 500 Reddit users from the r/SuicideWatch subreddit, labeled across five levels of suicide risk severity: Supportive, Indicator, Ideation, Behavior, and Attempt. The dataset was created to enable early intervention and prevention of suicide by identifying individuals who may be at risk.

The annotation process involved four practicing clinical psychiatrists who followed the guidelines outlined in the Columbia Suicide Severity Rating Scale (C-SSRS). The dataset includes the Reddit posts of the 500 users, along with their corresponding suicide risk severity labels. 

{ We} originally planned to include the CSSRS-Suicide dataset in our analysis, However, as the dataset is made up of posts from 500 people, with the number of posts included for each user varying greatly, from 1 in the lower {end} and more than 100 in the upper {end}. This made it difficult to evaluate the dataset fairly, with some users including more information than others. Furthermore, evaluating the users with too many posts would prove quite costly and exceed the context window of many of the smaller \gls{LLM}s tested. Upon trying to include only one post per user (e.g. by selecting only the first post for each user), the results were significantly worse and as such the dataset was discarded.

\subsection{Sampling}
\label{subsec:sampling}
As all testing was conducted through \gls{API}s, and for the bigger models, such as GPT-4, the costs could escalate rapidly, we employed two strategies for cost saving. First, we utilized the test dataset only if it was pre-split. Second, if the dataset exceeded 1000 examples, we employed a fair random sampling of 1000 instances. This sample size was chosen based on Hoeffding's inequality (Equation \ref{eqn:1}) \cite{hoeffding1994probability}, which provides a general upper bound for the probability {that the difference between the empirical mean  $\nu$  (i.e., the sample average) and the true mean $\mu$ (i.e., the population average) exceeds a predefined margin of error $\epsilon$, given a sample size $N$ 

\begin{equation}
\label{eqn:1}
P\left( \left| \nu - \mu \right| > \varepsilon \right) \leq 2e^{-2\varepsilon^2 N}
\end{equation}

Here, $N = 1000$ represents the number of sampled instances, and $\epsilon = 0.05$ corresponds to the maximum allowable deviation (5\% discrepancy) between the sample and population metrics. Substituting these values into Equation \ref{eqn:1}, the probability of the empirical mean deviating from the true mean by more than 5\% is bounded by $2e^{-2(0.05)^2 \times 1000} \approx 0.0135.$ This implies a confidence level exceeding 98.65\% that the observed error rate remains within 5\% of the true error rate.}

\subsection{Models}
\label{subsec:models}

To ensure a comprehensive assessment of \gls{LLM} capabilities across diverse architectures and sizes, we selected a wide range of models for our experiments. We included both large, state-of-the-art models known for their performance on various natural language processing tasks, as well as smaller, more accessible models that could potentially be deployed in resource-constrained environments.

Our model selection encompassed both closed-source models, accessible through APIs, and open-source models, which offer greater transparency and flexibility for research purposes. We included models from various families, including OpenAI's GPT series, Anthropic's Claude models, Mistral AI's models, Google's models, Meta's Llama series, and finally, Phi-2 and MentaLlama, a model specifically fine-tuned on mental health-related data. Additionally, we attempted to include Gemini 1.5 Pro in our evaluation{, however,} due to rate limit constraints and filtering mechanisms challenges mentioned in { Section \ref{sec:limitations}}, we were not able to include it.

Table \ref{tab:models} provides a summary of the models used in our experiments, along with their respective sizes, { the \gls{API}s through which we accessed them, and the costs for running our experiments.}

\begin{table}[]
\caption{Summary of large language models evaluated in this analysis { and cost of running our experiments.}}
\label{tab:models}
\centering{
\begin{tabular}{@{}llllll@{}}
\toprule
\multicolumn{1}{l}{\textbf{Model}} &
\multicolumn{1}{l}{\textbf{Size (Parameters)}} &
\multicolumn{1}{l}{\textbf{Source}} &
\multicolumn{1}{l}{\textbf{\gls{API} Service Provider}} &
\multicolumn{1}{l}{{\textbf{Experiment Costs (USD)}}} \\ \midrule
Claude 3 Haiku \cite{anthropic2024claude} & Proprietary & Anthropic & Claude \gls{API} & {\$2.3} \\
Claude 3 Sonnet \cite{anthropic2024claude} & Proprietary & Anthropic & Claude \gls{API} & {\$9.8} \\
Claude 3 Opus \cite{anthropic2024claude} & Proprietary & Anthropic & Claude \gls{API} & {\$33.8} \\
Claude 3.5 Sonnet \cite{anthropic2024claude35} & Proprietary & Anthropic & Claude \gls{API} & {\$6.8} \\
Gemma 2b \cite{team2024gemma} & 2B & Google & Nvidia NIM & {\$0.8} \\
Gemma 7b \cite{team2024gemma} & 7B & Google & Groq & {\$0.0} \\
Gemma 2 9b \cite{team2024gemma2} & 9B & Google & Nvidia NIM & {\$1.5} \\
Gemma 2 27b \cite{team2024gemma2} & 27B & Google & Nvidia NIM & {\$3.8} \\
Gemini 1.0 Pro \cite{team2023gemini} & Proprietary & Google & Gemini \gls{API} & {\$0.0} \\
GPT-4o mini \cite{openai2024gpt4omini} & Proprietary & OpenAI & OpenAI \gls{API} & {\$0.8} \\
GPT-3.5-Turbo \cite{openai2023gpt35turbo} & 175B & OpenAI & OpenAI \gls{API} & {\$3.8} \\
GPT-4o \cite{openai2024gpt4o} & Proprietary & OpenAI & OpenAI \gls{API} & {\$11.3} \\
GPT-4-Turbo \cite{achiam2023gpt} & Proprietary & OpenAI & OpenAI \gls{API} & {\$22.5} \\
GPT-4 \cite{achiam2023gpt} & Proprietary & OpenAI & OpenAI \gls{API} & {\$65.3} \\
Llama 3 8b \cite{dubey2024llama} & 8B & Meta & Groq & {\$0.0} \\
Llama 3.1 8b \cite{dubey2024llama} & 8B & Meta & Together AI & {\$1.5} \\
Llama 2 13b \cite{touvron2023llama2} & 13B & Meta & Together AI & {\$1.5} \\
Llama 2 70b \cite{touvron2023llama2} & 70B & Meta & Together AI & {\$6.0} \\
Llama 3 70b \cite{dubey2024llama} & 70B & Meta & Groq & {\$0.0} \\
Llama 3.1 70b \cite{dubey2024llama} & 70B & Meta & Together AI & {\$4.5} \\
Llama 3.1 405b \cite{dubey2024llama} & { 405B} & Meta & Together AI & {\$25.5} \\
Mistral 7b \cite{jiang2023mistral} & 7B & Mistral AI & Together AI & {\$1.5} \\
Mistral NeMo \cite{mistral2024nemo} & 12B & Mistral AI & Mistral AI \gls{API} & {\$1.5} \\
Mixtral 8x7b \cite{jiang2024mixtral} & 45B (13B) & Mistral AI & Groq & {\$0.0} \\
Mixtral 8x22b \cite{jiang2024mixtral} & 141B (39B) & Mistral AI & Together AI & {\$6.8} \\
Mistral Medium \cite{mistral2023la} & Proprietary & Mistral AI & Mistral AI \gls{API} & {\$17.3} \\
Mistral Large \cite{mistral2024large} & Proprietary & Mistral AI & Mistral AI \gls{API} & {\$18.0} \\
Mistral Large 2 \cite{mistral2024large2} & Proprietary & Mistral AI & Mistral AI \gls{API} & {\$15.0} \\
Phi-2 \cite{javaheripi2023phi2} & 2.7b & Phi-1 Labs & Nvidia NIM & {\$0.8} \\
Phi-3-mini \cite{abdin2024phi} & 3.8B & Phi-1 Labs & Nvidia NIM & {\$0.8} \\
Phi-3-small \cite{abdin2024phi} & 7B & Phi-1 Labs & Nvidia NIM & {\$1.5} \\
Phi-3-medium \cite{abdin2024phi} & 14B & Phi-1 Labs & Nvidia NIM & {\$2.3} \\
Qwen 2 72b \cite{yang2024qwen2} &  72B & Alibaba Cloud & Together AI & {\$4.5} \\
\bottomrule
\end{tabular} \\
}

\textbf{Note 1}: The Mixtral 8x7b and 8x22b models have two sizes because the total size and the number of parameters utilized at any point in inference are different.

{
\textbf{Note 2}: Gemini 1.0 Pro API and Groq models were available for free at the time these experiments were conducted. These are approximate costs, accounting only for the final runs presented in this paper and do not include any experimentation costs, which would drastically increase the overall expenditure.
}
\end{table}

For some of the side experiments discussed in Section \ref{sec:additional_investigations}, a subset of the available models was utilized due to cost constraints. However, the selected subset was carefully chosen to ensure representation from all relevant model families, maintaining the diversity of the study.


\subsection{Prompt Templates}
\label{subsec:prompt_templates}

The prompt, or instruction, given to an \gls{LLM} plays a { significant} role in its performance on a given task. Prior research has shown that even slight variations in prompt wording can significantly impact the quality of \gls{LLM} outputs \cite{salinas2024butterfly}. Therefore, we dedicated considerable effort to crafting effective prompts tailored to each task in our study.

To optimize \gls{LLM} performance on social media post analysis, we employed prompts that encouraged the models to adopt the role of a psychiatrist. Role-playing has proven effective in eliciting relevant and contextually appropriate responses from \gls{LLM}s \cite{kong2023better, zheng2023helpful}. We also utilized the concept of prompt templates, which consist of a fixed structure with variable elements that can be adapted for different tasks. This approach ensured consistency across experiments while minimizing prompt-based bias.

We developed multiple prompt templates for each task category: binary disorder detection, disorder severity evaluation, and psychiatric knowledge assessment. These templates varied in their structure and level of explicit instruction, aiming to accommodate the diverse capabilities of the \gls{LLM}s included in our study.

\subsubsection{Binary Disorder Detection Prompt Templates}
In the \gls{ZS} binary disorder detection task, we employed four distinct prompt templates, designated BIN-1 to BIN-4. By utilizing multiple prompt templates, we aimed to assess the impact of both open-ended and structured prompts on the performance of various tested \gls{LLM}s. This approach allowed us to evaluate \gls{LLM}s' sensitivity, variability, and adaptability to diverse prompting styles, highlighting their strengths and weaknesses in different scenarios.

As shown in { Table} \ref{tab:binary_prompts}, Prompts BIN-1 and BIN-2 utilized a straightforward structure, first instructing the \gls{LLM} to assume the role of a psychiatrist and then directly asking whether the poster of a given social media post exhibited a specific disorder. We included a final instruction to elicit a "yes" or "no" response for easier parsing of the \gls{LLM} output.

\begin{table}[]
\caption{Task 1 (Binary Disorder Classification) Prompt Templates}
\label{tab:binary_prompts}
\centering
\begin{tabular}{@{}l@{}} 
\toprule
\textbf{Prompts BIN-1 and BIN-2} \\ 
\midrule
As a psychologist, analyze the provided social media post(s). \\ 
\textbf{Is the poster depressed?}/ \textbf{Does the poster exhibit clear symptoms of depression?} \\
(Answer with Yes or No only without explaining your reasoning) \\ 
\midrule
\textbf{Prompt BIN-3} \\ 
\midrule
\makecell[l]{
[Task] \\
Act as a psychologist. \textbf{Label} the writer of the provided social media post \textbf{as 'depressed'} \\ 
\textbf{or 'not depressed'} according to provided guidelines. \\
{[Guidelines]} \\
- Concise Response: Respond only with 'Yes' (depressed) or 'No' (not depressed). \\
- No Explanations: Don't provide explanations for your assessment. \\
- Ambiguity: If the post is unclear, choose the most probable label. \\
{[Post]}
} \\
\midrule
\textbf{Prompt BIN-4} \\ 
\midrule
Analyze the following social media post to determine if the writer \textbf{exhibits clear} \\ 
\textbf{symptoms of depression} according to provided guidelines. \\
\\
... Same as BIN-3 \\
\\
\bottomrule
\end{tabular}
\end{table}

Prompts BIN-3 and BIN-4 introduced a more structured format, explicitly separating the prompt into "Task," "Guidelines," and "Post" sections.  The guidelines emphasized the desired output format (yes/no), discouraged explanations, and instructed the \gls{LLM} to choose the most probable label in cases of ambiguity. This structure aimed to improve the \gls{LLM}s' adherence to the task instructions and reduce the incidence of irrelevant or unparseable responses. The inspiration for this structure was from Gemini, as this is the structure used for its training \cite{team2023gemini}.

\subsubsection{Disorder Severity Evaluation Prompt Templates}
\label{sebsec:severity_prompts}

\begin{figure}[htbp]
\centering
\includegraphics[width=0.75\linewidth]{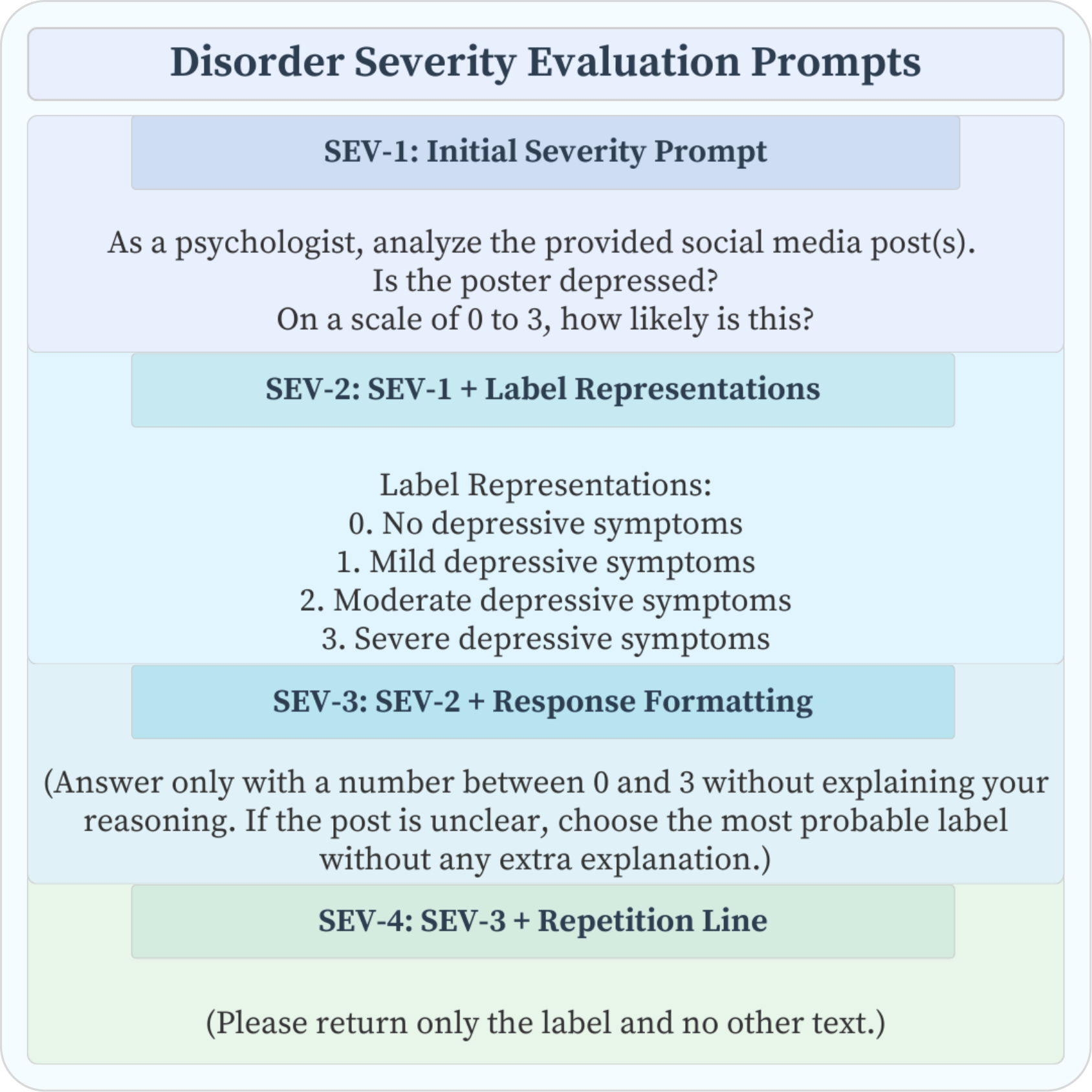}
\caption{Disorder Severity Evaluation Task Evolution}
\label{fig:severity_prompts}
\end{figure}

For the disorder severity evaluation task, we iteratively refined the prompt template based on initial observations from the binary task and the performance of various \gls{LLM}s. We began with a baseline prompt, SEV-1, which was a modified version of the binary prompt template BIN-1. The key modification in SEV-1 was the addition of a severity scale, representing varying levels of disorder severity. Figure \ref{fig:severity_prompts} illustrates all four Severity prompts (SEV-1 to SEV-4).

During the design of this initial prompt, we observed that most \gls{LLM}s performed better when provided with the full context of each severity level, rather than receiving no context or only the extreme values (0 and 4).

Subsequent refinements aimed to address the tendency of some expensive models to provide verbose explanations despite being instructed to give concise answers. We systematically introduced modifications to the prompt, resulting in three additional versions: SEV-2, SEV-3, and SEV-4.

\begin{enumerate}
    \item SEV-2: Modified the presentation of the severity scale from free text to a numbered list to enhance clarity.
    \item SEV-3: Incorporated an ambiguity clause (similar to the one used in BIN-3 and BIN-4) to guide responses when the post was unclear.
    \item SEV-4 :Reiterated the instruction to provide a concise answer without explanation. This simple repetition proved surprisingly effective, especially for Mistral models. The added line is henceforth referred to as \textit{repetition line}
\end{enumerate}

To evaluate the effectiveness of these modifications, we conducted experiments on a sample of 1000 posts from the DEPTWEET dataset, comparing the performance of different \gls{LLM}s across the various prompt versions (P1-P4). The results, presented in Table \ref{tab:sev-prompt-engineering-experiments-results}, demonstrate a general trend of improved performance from P1 to P4, particularly for models like Gemini 1.0 Pro and Llama 3 70b. Notably, even models that did not show significant improvement experienced only minor decreases in performance. Hence, prompt SEV-4 was chosen as the main prompt for the severity task.

\begin{table}[] \caption{Performance metrics and invalid responses of different models across the various severity prompts} \label{tab:sev-prompt-engineering-experiments-results} 
\centering
\begin{tabular}{@{}lcccc@{}} \toprule \textbf{Model/Prompt} & \textbf{P1} & \textbf{P2} & \textbf{P3} & \textbf{P4} \\ \midrule \multirow{3}{*}{GPT-3.5-Turbo} & MAE: 0.713 & MAE: \textcolor{OliveGreen}{-0.001} & MAE: \textcolor{OliveGreen}{-0.029} & MAE: \textcolor{OliveGreen}{-0.066} \\ & BA: 43.6\% & BA: \textcolor{BrickRed}{-1.4\%} & BA: \textcolor{BrickRed}{-1.7\%} & BA: \textcolor{BrickRed}{-0.4\%} \\ & IR: 0 & IR: 55 & IR: 0 & IR: 0 \\ \midrule \multirow{3}{*}{Gemini 1.0 Pro} & MAE: 0.629 & MAE: \textcolor{OliveGreen}{-0.039} & MAE: \textcolor{OliveGreen}{-0.095} & MAE: \textcolor{OliveGreen}{-0.133} \\ & BA: 42.7\% & BA: \textcolor{OliveGreen}{+2.8\%} & BA: \textcolor{OliveGreen}{+7.6\%} & BA: \textcolor{OliveGreen}{+11.5\%} \\ & IR: 0 & IR: 0 & IR: 0 & IR: 0 \\ \midrule \multirow{3}{*}{Claude 3 Haiku} & MAE: 0.97  & MAE: \textcolor{OliveGreen}{-0.041} & MAE: \textcolor{OliveGreen}{-0.069} & MAE: \textcolor{OliveGreen}{-0.092} \\ & BA: 29.8\% & BA: \textcolor{OliveGreen}{+2.3\%} & BA: \textcolor{OliveGreen}{+3.2\%} & BA: \textcolor{OliveGreen}{+4.5\%} \\ & IR: 70 & IR: 22 & IR: 18  & IR: 4 \\ \midrule \multirow{3}{*}{Mistral Medium} & MAE: 0.608 & MAE: \textcolor{OliveGreen}{+0.001} & MAE: \textcolor{OliveGreen}{-0.051} & MAE: \textcolor{OliveGreen}{-0.052} \\ & BA: 46.2\% & BA: \textcolor{OliveGreen}{+2.2\%} & BA: \textcolor{OliveGreen}{+5.8\%} & BA: \textcolor{OliveGreen}{+5.7\%} \\ & IR: 38     & IR: 39 & IR: 3  & IR: 0 \\ \midrule \multirow{3}{*}{Gemma 7b} & MAE: 0.892 & MAE: \textcolor{BrickRed}{+0.027}  & MAE: \textcolor{BrickRed}{+0.058}  & MAE: \textcolor{BrickRed}{+0.031}  \\ & BA: 35.5\% & BA: \textcolor{BrickRed}{-1\%}     & BA: \textcolor{BrickRed}{-1.9\%}  & BA: \textcolor{BrickRed}{-1.3\%}  \\ & IR: 4  & IR: 4 & IR: 7 & IR: 0 \\ \midrule \multirow{3}{*}{Llama 3 8b}     & MAE: 1.093 & MAE: \textcolor{BrickRed}{+0.146}  & MAE: \textcolor{OliveGreen}{-0.043} & MAE: \textcolor{OliveGreen}{-0.324} \\ & BA: 32\%   & BA: \textcolor{BrickRed}{-5.7\%}   & BA: \textcolor{BrickRed}{-1.3\%}  & BA: \textcolor{OliveGreen}{+5\%}   \\ & IR: 137 & IR: 117 & IR: 57 & IR: 7 \\ \midrule \multirow{3}{*}{Llama 3 70b}    & MAE: 0.949 & MAE: \textcolor{OliveGreen}{-0.062} & MAE: \textcolor{OliveGreen}{-0.105} & MAE: \textcolor{OliveGreen}{-0.275} \\ & BA: 29\%   & BA: \textcolor{OliveGreen}{+4.3\%}  & BA: \textcolor{OliveGreen}{+6.0\%}  & BA: \textcolor{OliveGreen}{+13.3\%} \\ & IR: 23 & IR: 22 & IR: 17 & IR: 0 \\ \bottomrule
\end{tabular} 
\\
\textbf{Notes:}
BA: Balanced Accuracy, MAE: Mean Absolute Error, IR: Invalid Responses \end{table}

For the \gls{FS} disorder severity evaluation task, the SEV-4 prompt was initially used as a foundation. Subsequently, three hand-selected examples were added to the prompt as a guidance (The chosen examples are not sent to the \gls{LLM}s for inference). Table \ref{tab:severity_fs_prompt_4} clearly details these prompt modifications.

\begin{table}[h!]
\centering
\begin{minipage}{0.4\textwidth} 
\centering
\caption{Prompt SEV-4 (\gls{FS})}
\label{tab:severity_fs_prompt_4}
\begin{tabular}{@{}l@{}} 
\toprule
\textbf{Prompt} \\ 
\midrule
\textbf{SEV-4 (ZS) prompt} \\
Some examples for guidance: \\
Example 1: \\
Post: <post contents> \\
Response: Yes/No \\
Example 2: \\
Post: <post contents> \\
Response: Yes/No \\
Example 3: \\
Post: <post contents> \\
Response: Yes/No \\
The post to be analyzed is: \\
\bottomrule
\end{tabular}
\end{minipage}
\hspace{0.02\textwidth} 
\begin{minipage}{0.4\textwidth} 
\centering
\caption{Prompt KNOW-1}
\label{tab:knowledge_prompt}
\begin{tabular}{@{}p{0.9\textwidth}@{}} 
\toprule
\textbf{Prompt KNOW-1} \\ 
\midrule
As a psychologist, please analyze the following MCQ question and select the most appropriate answer. \\
Do not provide any explanation for your choice. \\
Indicate your choice with a single English letter. \\
(Provide only one answer) \\
\bottomrule
\end{tabular}
\end{minipage}
\end{table}

\subsubsection{Psychiatric Knowledge Assessment Prompt Template}

For the psychiatric knowledge assessment task, we employed a simple prompt template (KNOW-1) that instructed the \gls{LLM} to answer a multiple-choice question with only the corresponding letter (A, B, C, or D) and no explanation. There was no need for any iterative experimentation. The prompt is illustrated in Table \ref{tab:knowledge_prompt}

\subsection{Evaluation Metrics}
\label{subsec:evaluation_metrics}

To assess the performance of the \gls{LLM}s across our defined tasks, we primarily employed accuracy as our evaluation metric. For most tasks, we applied fair random sampling to the datasets, ensuring balanced class distributions. This balanced sampling naturally leads to the calculation of \gls{BA}, which is the average of recall obtained on each class. \gls{BA} is particularly well-suited for classification tasks where class imbalance may be present, as it provides a more equitable assessment of the model's ability to correctly identify both positive and negative instances compared to traditional accuracy.

However, in cases where fair random sampling was not feasible due to limitations in the dataset, such as the SAD dataset with its sparse distribution of severity classes, we utilized regular random sampling with a random state of 42 and evaluated performance using traditional accuracy. Similarly, for the MedMCQA dataset, which involves multiple-choice question answering without distinct class labels, we also relied on accuracy as the primary evaluation metric.

For the disorder severity evaluation task, we employed two metrics to assess the performance of \gls{LLM}s: \gls{BA} and \gls{MAE}. \gls{BA}, in this context, refers to the proportion of posts for which the \gls{LLM} predicted the exact severity level assigned by human annotators. It serves as a measure of how often the model gets the severity classification exactly right. On the other hand, \gls{MAE} quantifies the average magnitude of the errors in the predicted severity ratings. It provides a complementary perspective on model performance by assessing how close, on average, the predicted severity levels are to the true levels, even if they are not exactly correct. A model could potentially improve in one metric while worsening in the other.

\subsection{Parsing Model Outputs}
{
Parsing, the process of interpreting and extracting meaningful information from \gls{LLM} outputs, proved to be a non-trivial aspect of our methodology. While seemingly straightforward for tasks requiring simple responses like "yes/no" or numerical ratings, the reality was more complex. \gls{LLM}s, particularly larger and more sophisticated models, often deviate from explicit instructions regarding output format, frequently providing explanations or incorporating the requested answer within a longer response.

To manage this output variability and facilitate the extraction of relevant information, we developed custom parsers designed for each task.

\begin{enumerate}
    \item Binary Disorder Detection: Our parser searched for the text "yes" or "no" in the \gls{LLM} output. If only one option was found, it was taken as the model's answer. If both were present, or neither was found, the output was considered to be an invalid response.

    \item  Disorder Severity Evaluation: The parser searched for numerical values within the specified severity range. If a single valid number was found, it was accepted as the answer. If multiple numbers were found, the number was outside the valid range, or no number was found, then output was considered to be an invalid response. 

    \item Psychiatric Knowledge Assessment: Parsing multiple-choice answers proved more challenging, particularly for the letter "a," which could be either the answer or part of a word like "an." We employed an iterative approach, refining the parser's regular expressions by examining invalid outputs. This iterative process revealed that \gls{LLM}s tend to follow a limited set of patterns in their responses, making it feasible to achieve reliable parsing with relatively few rules. Interestingly, only eight rules were required to parse the majority of outputs across all \gls{LLM}s.
\end{enumerate}
}

\subsection{Additional Considerations}
\label{subsec:additional_considerations}
In addition to the evaluation metrics and parsing strategies discussed previously, there are a few additional considerations worth noting. { First, despite our efforts in prompt engineering, certain \gls{LLM}s, such as Mistral Medium and Claude 3 Opus, struggled to adhere to instructions across tasks. During the design of the Severity prompts (Section~\ref{sebsec:severity_prompts}), we observed that including a repetition line (\textit{``(Please return only the label (Yes or No) and no other text)''}) improved compliance. To address similar issues in the binary classification task, we created a modified prompt, BIN1.1, by appending this repetition line to the original binary prompt (see Table~\ref{tab:binary_1.1_prompt}). This adjustment improved adherence to output formatting requirements in subsequent experiments. We also conducted a small investigation into the effect of this repetition line on model performance (see Section~\ref{subsec:repetition_line_experiment}).}


Second, our analysis included several smaller, specialized experiments designed to evaluate specific aspects of \gls{LLM} performance or explore particular research questions. Due to their focused nature, the details of these experiments, including their implementation, results, and discussion, are presented separately in Section \ref{sec:additional_investigations} to maintain clarity and coherence in the main body of our analysis.

Third, during evaluation, we compared two approaches: setting a maximum output token limit of 2 tokens versus allowing the models to respond freely without truncation. The rationale behind the former approach was that it would likely include the keyword for the diagnosis we were seeking. Conversely, for models that did not adhere well to prompt instructions, the diagnosis might appear later in their response. Our findings revealed that the longer a model's response, the more likely it was to become invalid due to the possibility of it using 'yes' or 'no' when providing explanations for its decision. Consequently, we decided to impose a response length limit of 2 tokens for all models tested. { A detailed analysis of the invalid response rates across different models, which informed this decision, is presented in Section \ref{subsec:invalid_response_analysis}.} 


\section{Results \& Discussion}
\label{sec:discussion}
{
In this section, we present the results of our experiments, analyzing the performance of various \gls{LLM}s across the three tasks defined in Section \ref{sec:methedology}. We first present the results of our performance variability assessment (Section \ref{subsec:performance_variability_results}), which provides insights into the inherent reliability of each model and helps contextualize the significance of performance differences observed in subsequent tasks. Next, we delve into the results of the \gls{ZS} binary disorder detection task (Section \ref{subsec:task_1}), followed by the \gls{ZS}/\gls{FS} disorder severity evaluation task (Section \ref{subsec:task_2}), and the \gls{ZS} psychiatric knowledge assessment task (Section \ref{subsec:task_3}). We then present an initial exploration of fine-tuning for the severity evaluation task (Section \ref{subsec:finetuning}). Finally, we conclude with a brief discussion of the study's limitations (Section \ref{subsec:limitations_discussion}), which will be explored in greater detail in Section \ref{sec:limitations}.
}

In our evaluation of the \gls{LLM}s across the three distinct tasks, we observed a wide range of performance capabilities and several noteworthy trends. In the \gls{ZS} binary disorder detection task, GPT-4, GPT-4o, Mistral NeMo{,} and Llama 2 70b consistently outperformed other models. Additionally, the choice of prompt significantly influenced model performance, with some models favoring more structured prompts while others performed better with open-ended ones. Notable mentions in this category include the Gemma 7b model which achieved 76.70\% (BIN-4) accuracy on Dreaddit, which is significantly higher than the previous BIN-1 best of 74.50\% { achieved by Mistral NeMo. Mistral NeMo is also notable for competing with models much larger than its size.}

The \gls{ZS}/\gls{FS} disorder severity evaluation task revealed that most models benefited from \gls{FS} learning, where they were provided with a few examples before making predictions. However, the degree of improvement varied across models and datasets, highlighting the need for careful consideration of both model architecture and data characteristics. Finally, in the \gls{ZS} psychiatric knowledge assessment task, we found that the more recent models generally performed better, compared to older, much larger models (e.g. Llama 3 70b and Phi-3-medium competing with GPT-4, GPT-4-Turbo, Claude 3.5 Sonnet, and others). However, with models released around the same time period, model size seemed to mostly be the deciding factor (Llama 3.1 models are excluded from this finding due to their tendency to filter/not respond to our prompts due to their sensitive mental health nature).

{ 
\subsection{Performance Variability Results}
\label{subsec:performance_variability_results}
Table \ref{tab:performance_variability} presents the results of our performance variability assessment, as described in Section \ref{subsec:task_formulation}. The table shows the standard deviation of accuracy for each model across five repeated evaluations on the DEPTWEET dataset.

The results indicate that most models exhibited low performance variability, with standard deviations below 0.5\%. Specifically, models such as Gemma 2 9b, Gemma 2 27b, GPT-3.5-Turbo, Llama 2 70b, Llama 3 70b, Phi-3-small, Phi-3-medium, and Qwen 2 72b fall within this range. Mixtral 8x22b showed a higher variability with a standard deviation of 1.6\%. Models unlisted in the table demonstrated 0\% standard deviation.

These results are crucial for interpreting the significance of performance differences observed in the subsequent task results. The consistently low variability of most models suggests that performance changes exceeding approximately 0.5\% (or 1.6\% for Mixtral 8x22b) are likely due to experimental manipulations (e.g., prompt variations) and not random model fluctuations. This allows us to more confidently analyze the impact of different factors on model performance in the tasks that follow.

\begin{table}[h!]
\caption{Performance Variability on DEPTWEET Across Five Runs}
\label{tab:performance_variability}
\centering
\begin{tabular}{@{}lr@{}}
\toprule
\textbf{Model} & \textbf{Standard Deviation (\%)} \\ \midrule
Gemma 2 9b & 0.05 \\
Gemma 2 27b & 0.1 \\
GPT-3.5-Turbo & 0.4 \\
Llama 2 70b & 0.2 \\
Llama 3 70b & 0.1 \\
Mixtral 8x22b & 1.6 \\
Phi-3-small & 0.05 \\
Phi-3-medium & 0.16 \\
Qwen 2 72b & 0.04 \\
\bottomrule
\end{tabular}
\end{table}
}

\subsection{Binary Disorder Classification (Task 1)}
\label{subsec:task_1}
{The results for the \gls{ZS} binary disorder detection task are presented below. We first examine the performance of models using the BIN-1 prompt, followed by an analysis of how modifications in prompts BIN-2, BIN-3, and BIN-4 affected model performance.}

\subsubsection{Results of Prompt BIN-1}
\begin{figure}[htbp!]
\centering
\includegraphics[width=1\linewidth]{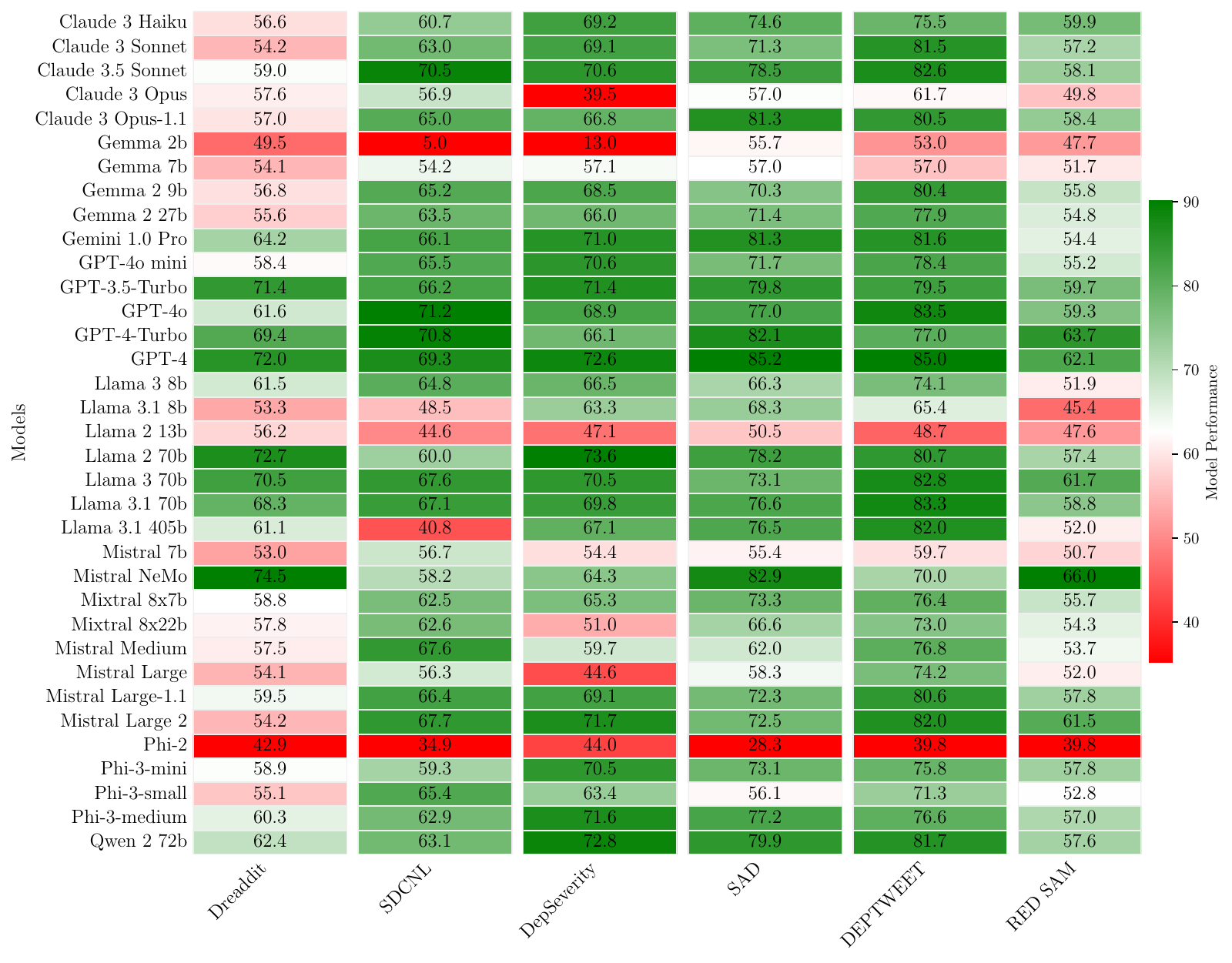}
\caption{Results of Prompt BIN-1 on Task 1}
\label{fig:binary_1_results}
\end{figure}

An analysis of Figure \ref{fig:binary_1_results} reveals that the performance of each model varies across datasets. However, a general trend emerges:  OpenAI stands out with three of the six best-performing models. Specifically, GPT-4 achieves top scores of 85.20\% on the SAD dataset and 85.00\% on the DEPTWEET dataset, while GPT-4o excels with 71.20\% on the SDCNL dataset. In addition to OpenAI's success, Llama 2 70b performs best in the DepSeverity dataset with 73.58\%, Mistral NeMo leads in both Dreaddit Test and RED SAM with 74.50, and 66.00\% respectively.

OpenAI models consistently demonstrate strong performance. GPT-4 achieves the highest accuracy among the tested models, reaching approximately 85\% on both the DEPTWEET and SAD datasets. GPT-4-Turbo and GPT-4o generally follow closely behind GPT-4, though with some performance fluctuations. GPT-4-Turbo typically lags slightly, outperforming GPT-4 in two instances (63.7\% in RED SAM vs. 62.1\%, and 70.8\% in SDCNL vs. 69.3\%). GPT-4o exhibits more significant fluctuations, sometimes exceeding all other variants (e.g., 71.2\% in SDCNL) but also deviating from GPT-4 by up to 9.8\% in SDCNL. GPT-3.5-Turbo consistently follows closely behind GPT-4, deviating by approximately 5\% at worst, which is notable considering its much smaller size. The newest and most cost-efficient model, GPT-4o mini, generally performs worse compared to GPT-3.5-Turbo.

The Llama 2 70b model demonstrates notable performance, achieving 73.58\% in DepSeverity, unexpectedly outperforming the Llama 3 70b (70.5\%) and even surpassing GPT-4, which achieves 72.6\%. The newer Llama 3 models exhibit greater consistency, with deviations of only 3-10\% between the 8b and 70b variants, compared to 10-20\% deviations in the Llama 2 family. However, the Llama 3.1 family of models was a significant disappointment, with the Llama 3.1 8b model performing considerably worse compared to the earlier version, except in two datasets (SAD \& DEPTWEET), where the performance is generally comparable. A similar observation applies to the Llama 3.1 70b model, which is generally closer in performance to the earlier version { and} actually outperforms the Llama 3 70b model in the SAD and DEPTWEET datasets. Lastly, the Llama 3.1 405b model was a major disappointment, performing even worse than the Llama 3.1 70b model across all six datasets.

Mistral 7b shows moderate performance, achieving an accuracy of 55\% (± 2.8\%) across datasets, barely surpassing the baseline accuracy of a random classifier. Although it outperforms Llama 2 13b, it is surpassed by Llama 3 8b. The Mistral Medium model demonstrates similar performance to Mistral 7b. Mistral Large initially encounters difficulties due to instruction-following issues but shows significant improvement with the BIN-1.1 prompt, achieving a 5-25\% increase in accuracy depending on the dataset, with the exception of Dreaddit. The Mistral NeMo model stands out as the best-performing model on the Dreaddit Test dataset with 74.5\%, outperforming GPT-4's 72\%. Beyond Dreaddit, Mistral NeMo performs better than Mistral 7b, but is generally outpaced by Llama 3 8b. As for Mistral Large 2, its performance is much better than Mistral Large 1, even mostly outperforming Mistral Large 1 with the BIN-1.1 prompt. However, it performs worse overall compared to the much smaller Mistral NeMo model

The Claude family presents varied results. Claude 3 Haiku often performs similarly to or even surpasses Claude 3 Sonnet, particularly in Dreaddit (56.6\% vs. 54.2\%) and SAD (74.6\% vs. 71.3\%). Claude 3 Sonnet shows a significant advantage over the Haiku model only in DEPTWEET (81.5\% vs. 75.5\%). Claude 3 Opus, Anthropic's flagship model, also struggles initially (due to the same instruction-following problems) but improves with BIN-1.1, though only marginally outperforming Haiku and Sonnet in most cases, except SAD (81.3\% vs. 74.6\% and 71.3\%). The newer Claude 3.5 Sonnet model outperforms all of its siblings, including the Claude 3 Opus model (even when using the BIN-1.1 prompt) in all but two instances (78.5\% vs 81.3\% in SAD, and 58.1\% vs 58.4\% in RED SAM). Overall, the Claude family demonstrates inconsistency in model performance compared to other families. However, the Claude 3.5 Sonnet model is overall the best-performing model from this family.

Gemini 1.0 Pro generally trails GPT-3.5-Turbo, only slightly outperforming it in the DEPTWEET dataset. This performance is notable given GPT-3.5-Turbo's proximity to GPT-4's capabilities. Gemma 7b performs similarly to Mistral 7b, while Gemma 2b and Phi-2 lag significantly, even falling below baseline performance, with Gemma 2b reaching only 5\% and 13\% accuracy in SDCNL and DepSeverity, respectively.

Looking at the newest model families—Gemma 2 and Phi-3—there is clear inconsistency in the results obtained between the Gemma 2 9b and the Gemma 2 27b models. In 5 out of 6 instances, the Gemma 2 9b model outperforms its larger sibling, with the largest difference being in the DEPTWEET dataset (80.4\% vs 77.9\%). Overall, the two models are close in performance; however, it is unexpected that the smaller model is consistently more performant.

As for the Phi-3 family, we observe a similar trend of inconsistency. The Phi-3-mini model consistently outperforms the larger Phi-3-small model in 5 out of the 6 datasets, and even outperforms the Phi-3-medium model in the RED SAM dataset (57.8\% vs 57\%). The Phi-3-medium model generally outperforms its smaller siblings except for two instances.

Lastly, the Qwen 2 72b model performs well across most datasets without any specifically noteworthy performance. 

\begin{figure}[htbp!]
\centering
\includegraphics[width=1\linewidth]{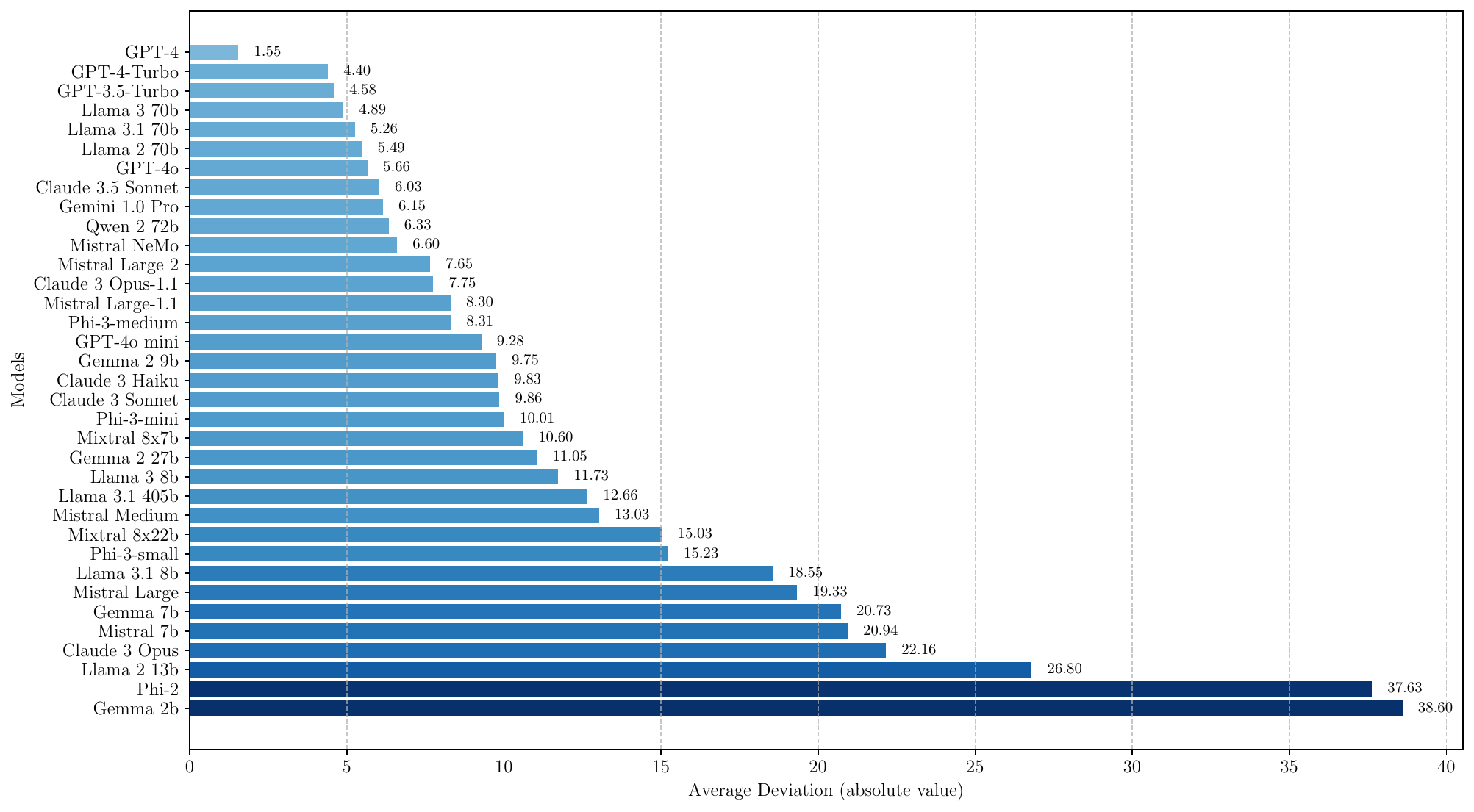}
\caption{Average Deviation of Model Accuracy from the Best Performing Model's Accuracy on each dataset of Task 1 (Dreaddit, SDCNL, DepSeverity, SAD, DEPTWEET, RED SAM) using Prompt BIN-1. Deviation is calculated for each dataset as the absolute difference in accuracy between a given model and the best performing model on that specific dataset. These deviations are then averaged across all datasets to produce the final values shown.}
\label{fig:binary_1_results_deviation}
\end{figure}

Figure \ref{fig:binary_1_results_deviation} summarizes model performance deviations, indicating how far each model is from the best-performing model for each dataset. This allows for an easy comparison between models and provides a general idea of their overall performance. A clear observation is that the graph is separated into clusters of similarly performing models.

To start, GPT-4 excels with an exceptionally low average deviation of 1.55\%. This places it at the top of the performance spectrum. Following closely behind, we have a cluster of high performers, including GPT-4-Turbo, GPT-3.5-Turbo, Llama 3 70b, Llama 3.1 70b, Llama 2 70b, GPT-4o, Claude 3.5 Sonnet, Gemini 1.0 Pro, Qwen 2 72b, and Mistral NeMo, with deviations ranging from 4.40\% to 6.60\%. It is noteworthy that the GPT-3.5-Turbo model slightly outperforms the GPT-4o model (4.58\% vs. 5.66\% average deviation), while being only very slightly outperformed by the GPT-4-Turbo model (4.40\%). This highlights the exceptional performance of the older model.

Next, models such as Mistral Large 2, Claude 3 Opus, and Mistral Large, both using the BIN-1.1 prompt, along with Phi-3-medium, demonstrate similar performance with an average deviation of around 8\% (± 0.3\%). Considering the Phi-3-medium model's relatively small size of 14 billion parameters, its performance is particularly impressive and indicates a high level of optimization.

The next cluster includes moderately performing models with average deviations ranging from 9.28\% to 11.73\%. This cluster includes GPT-4o mini, Gemma 2 9b/27b, Claude 3 Haiku/Sonnet, Phi-3-mini, Mixtral 8x7b, and Llama 3 8b.

The remaining clusters include models with average deviations of more than 12\%, such as Mistral Medium, Phi-3-Small, Mistral Large, and Claude 3 Opus (with the prompt BIN-1), Gemma 7b, Mistral 7b, Llama 2 13b, Phi-2, Gemma 2b, and others. Most disappointing in this category are the Llama 3.1 405b and, to a much lesser extent, the Llama 3.1 8b, which performs poorly despite being newer and of much larger models.

Overall, several models stand out for their notable (whether positive or negative) performance on the six binary datasets. OpenAI stands out for its performance once more, with the top three models belonging to it, and the next three belonging to Meta. Sadly, Llama 3.1 70b performs slightly worse than the older Llama 3 70b model, but this is much better than the Llama 3.1 405b model, which was a major disappointment.

Mistral NeMo slightly outperforms the Mistral Large 2 model despite their significant size difference (14b vs. 123b). Phi-3-medium and Gemma 2 9b also perform admirably, holding their own against much larger models. Overall, Mistral NeMo emerges as the top model in the sub-14 billion parameter category, followed by Phi-3-medium. The Llama 3 70b model leads in the sub-70 billion parameter category, followed closely by the Llama 3.1 70b model. The Phi-3-mini model also stands out, performing comparably to Claude 3 Haiku/Sonnet despite being only 3.8 billion parameters in size.

\subsubsection{Results of Prompts BIN-2/3/4}
Table \ref{tab:binary_prompt_merged_results} presents the changes in model performance when moving from prompt BIN-2 to BIN-3, and BIN-4 respectively. Comparing the improvement when going from prompt BIN-1 to prompt BIN-2, we can observe the effect of changing the main task from checking whether the poster suffers from a disorder (i.e., 'is the user stressed') to checking whether the poster shows or exhibits symptoms of a disorder ('Does this poster exhibit clear symptoms of stress'). This rephrasing results in different outcomes, specifically creating a less strict boundary for a true (i.e., stressed) label.

Models such as Gemini 1.0 Pro, Mistral Medium, Mistral Large 2, Llama 2 13b, Llama 3.1 8b/70b/405b, Mixtral 8x7b/22b, Mistral 7b, Gemma 7b, Gemma 2 9b/27b, Phi-3-mini/small, and Qwen 2 72b exhibited improvements moving from prompt BIN-1 to BIN-2, particularly on the Dreaddit and DepSeverity datasets. These improvements were more pronounced in models with initially lower performance, generally stabilizing around the 65-70\% accuracy range. For instance, Gemini 1.0 Pro, which started with an accuracy above 60\%, improved to 72.20\% on the Dreaddit Test. Mixtral 8x22b, Gemma 7b, and Mistral Large 2 demonstrated substantial improvements, nearing the 70\% accuracy mark, with increases of 12.29\%, 13.68\%, and 14.80\%, respectively. This performance approaches that of the previously best-performing model, Mistral NeMo, which had achieved 74.50\% accuracy on the Dreaddit Test with BIN-1. The Phi-3-mini and Phi-3-small models showed moderate improvements of 4.80\% and 2.30\%, respectively, while Qwen 2 72b improved by 5.70\%. In contrast, the SDCNL dataset generally did not exhibit similar improvements, potentially due to lower labeling quality, which may have hindered the models' ability to generalize improvements.

When examining the transitions from BIN-1 to BIN-3 and BIN-4, fewer models exhibited improved performance. Notably, models such as GPT-3.5-Turbo, Claude 3 Haiku, Llama 3 8b, Llama 2 70b, Llama 3.1 405b, Mistral NeMo, Phi-3-medium, and Qwen 2 72b showed declines ranging from 2\% to over 20\%, depending on the model and dataset. This trend indicates a general preference for the BIN-2 prompt, likely due to its clearer{,} less structured approach.

Among the models showing notable improvements, Gemini 1.0 Pro demonstrated significant gains on Dreaddit and DepSeverity, with increases of +8\% and +4\% (BIN-4), respectively. Gemma 7b also excelled, achieving a 22.58\% increase on Dreaddit with prompt BIN-4, reaching 76.70\% accuracy—surpassing the previous best of 74.50\% achieved by Mistral NeMo. On the DepSeverity dataset, Gemma 7b also improved significantly with a gain of 11.88\% (BIN-3). Gemma 2b, while not showing a clear preference between BIN-3 and BIN-4, exhibited fluctuations with significant improvements, notably a +45\% increase (BIN-4) on SDCNL, starting from an anomalous 5\%, and a +51\% increase (BIN-3) on DepSeverity from a baseline of 13\%.

Llama 2 13b displayed a strong preference for BIN-4 over BIN-3, showing improvements with BIN-4 of 9.41\% on Dreaddit and 13.80\% on DepSeverity, as opposed to the significant performance degradation with BIN-3 (-42\%, -26\%). Meanwhile, the Llama 3.1 8b model showed substantial improvements with both BIN-3 and BIN-4, particularly with increases of +3.7\% (BIN-3) and 10.80\% (BIN-4) on Dreaddit, and a +6.8\% gain (BIN-4) on DepSeverity. The Llama 3.1 70b model showed minor improvements of approximately 1-2\% with BIN-4 across datasets.

The Mistral and Mixtral models consistently improved with each prompt, with a clear preference for BIN-4. Mistral Medium achieved a notable 15.40\% improvement on DepSeverity with BIN-4, reaching a total accuracy of 75.10\%. Mixtral 8x7b and Mixtral 8x22b also showed substantial improvements, with increases of 10.75\% on Dreaddit and 21\% on DepSeverity, respectively. Mistral Large 2 similarly preferred BIN-4, improving by 10.40\% on Dreaddit and 2.50\% on DepSeverity. In contrast, Mistral NeMo exhibited declines with both BIN-3 and BIN-4, showing significant decreases in performance on Dreaddit and DepSeverity.

The Phi-3 models displayed mixed results, with Phi-3-mini showing a slight improvement of 0.80\% on Dreaddit with BIN-4, but declines in other metrics. Phi-3-small exhibited moderate improvements with BIN-2 and BIN-3, and a notable 5.50\% increase with BIN-4. Phi-3-medium, however, showed declines across all prompts.

Gemma 2 9b and Gemma 2 27b displayed consistent improvements, particularly with BIN-4 across all datasets. The highest accuracy of 71.70\% on DepSeverity with BIN-4 suggests a favorable response to this prompt. Gemma 2 27b achieved accuracies of 62.40\%, 65.40\%, and 71.70\% on Dreaddit, SDCNL Test, and DepSeverity, respectively.

Overall, most models showed a preference for the BIN-4 prompt, achieving the highest accuracy improvements, especially on the Dreaddit and DepSeverity datasets. This preference suggests that a structured and detailed prompt format generally enhances performance. The shift from BIN-1 to BIN-2 also led to significant improvements, highlighting the benefit of rephrasing tasks to focus on symptoms rather than direct diagnoses.

{Regarding the overall best-performing models}, Gemma 7b as the top performer on the Dreaddit dataset with 76.70\% accuracy (BIN-4), surpassing Mistral NeMo's 74.50\%. Other notable performances include Gemini 1.0 Pro, Mixtral 8x7b, and Mixtral 8x22b, which achieved accuracies of 72.20\%, 69.50\%, and 70.10\% on BIN-4 and BIN-2, respectively. For the SDCNL dataset, Llama 3 70b matched GPT-4o's 71.20\% with BIN-2, while Gemma 7b reached 66.10\% with prompt BIN-2. On the DepSeverity dataset, Mistral Medium achieved an accuracy of 75.10\% (BIN-4), surpassing Llama 2 70b's 73.58\% (previous best on BIN-1). Several models, including Phi-3-small, Mixtral 8x7b, Mixtral 8x22b, Llama 3 70b, Gemma 2b, and Gemma 7b, approached or exceeded the 70\% accuracy threshold, with Qwen 2 72b achieving the best performance at 76.60\% (BIN-4).

\begin{table}[h!]
\caption{Merged Results of Prompt BIN-2/BIN-3/BIN-4 on Task 1}
\label{tab:binary_prompt_merged_results}
\centering
\resizebox{\textwidth}{!}{
\begin{tabular}{@{}lrrrrrrr@{}}
\toprule
\textbf{Model} & \textbf{Dreaddit Test} & \textbf{SDCNL Test} & \textbf{DepSeverity} \\ \midrule
Claude 3 Haiku & \textcolor{black}{56.60} / \textcolor{OliveGreen}{+5.20} / \textcolor{BrickRed}{-6.60} / \textcolor{BrickRed}{-5.60} / 61.80\textsuperscript{\textbf{\footnotesize{2}}} & \textcolor{black}{60.70} / \textcolor{BrickRed}{-0.80} / \textcolor{BrickRed}{-6.90} / \textcolor{BrickRed}{-6.40} / 60.70\textsuperscript{\textbf{\footnotesize{1}}} & \textcolor{black}{69.20} / \textcolor{BrickRed}{-0.10} / \textcolor{BrickRed}{-15.80} / \textcolor{BrickRed}{-15.20} / 69.20\textsuperscript{\textbf{\footnotesize{1}}} \\
Gemma 2b & \textcolor{black}{49.49} / \textcolor{BrickRed}{-3.09} / \textcolor{OliveGreen}{+10.21} / \textcolor{OliveGreen}{+0.51} / 59.70\textsuperscript{\textbf{\footnotesize{3}}} & \textcolor{black}{5.0} / \textcolor{OliveGreen}{+2.00} / \textcolor{OliveGreen}{+39.00} / \textcolor{OliveGreen}{+45.00} / 50.00\textsuperscript{\textbf{\footnotesize{4}}} & \textcolor{black}{13.0} / \textcolor{BrickRed}{-4.00} / \textcolor{OliveGreen}{+51.00} / \textcolor{OliveGreen}{+39.00} / 64.00\textsuperscript{\textbf{\footnotesize{3}}} \\
Gemma 7b & \textcolor{black}{54.12} / \textcolor{OliveGreen}{+13.68} / \textcolor{OliveGreen}{+11.88} / \textcolor{OliveGreen}{+22.58} / \underline{76.70}\textsuperscript{\textbf{\footnotesize{4}}} & \textcolor{black}{54.21} / \textcolor{BrickRed}{-5.31} / \textcolor{OliveGreen}{+11.89} / \textcolor{OliveGreen}{+0.89} / 66.10\textsuperscript{\textbf{\footnotesize{3}}} & \textcolor{black}{57.10} / \textcolor{OliveGreen}{+11.88} / \textcolor{OliveGreen}{+11.89} / \textcolor{OliveGreen}{+6.20} / 68.99\textsuperscript{\textbf{\footnotesize{3}}} \\
Gemma 2 9b & \textcolor{black}{56.8} / \textcolor{OliveGreen}{+1.30} / \textcolor{OliveGreen}{+3.80} / \textcolor{OliveGreen}{+4.20} / 61.00\textsuperscript{\textbf{\footnotesize{4}}} & \textcolor{black}{65.2} / \textcolor{BrickRed}{-6.80} / \textcolor{BrickRed}{-5.70} / \textcolor{BrickRed}{-2.20} / 65.20\textsuperscript{\textbf{\footnotesize{1}}} & \textcolor{black}{68.5} / \textcolor{BrickRed}{-0.40} / \textcolor{OliveGreen}{+1.80} / \textcolor{OliveGreen}{+3.20} / 71.70\textsuperscript{\textbf{\footnotesize{4}}} \\
Gemma 2 27b & \textcolor{black}{55.6} / \textcolor{OliveGreen}{+3.00} / \textcolor{OliveGreen}{+4.10} / \textcolor{OliveGreen}{+6.80} / 62.40\textsuperscript{\textbf{\footnotesize{4}}} & \textcolor{black}{63.5} / \textcolor{BrickRed}{-1.00} / \textcolor{BrickRed}{-1.30} / \textcolor{OliveGreen}{+1.90} / 65.40\textsuperscript{\textbf{\footnotesize{4}}} & \textcolor{black}{66.0} / \textcolor{OliveGreen}{+4.10} / \textcolor{OliveGreen}{+4.10} / \textcolor{OliveGreen}{+5.70} / 71.70\textsuperscript{\textbf{\footnotesize{4}}} \\
Gemini 1.0 Pro & \textcolor{black}{64.2} / \textcolor{OliveGreen}{+5.70} / \textcolor{OliveGreen}{+1.10} / \textcolor{OliveGreen}{+8.00} / 72.20\textsuperscript{\textbf{\footnotesize{4}}} & \textcolor{black}{66.1} / \textcolor{BrickRed}{-5.50} / \textcolor{BrickRed}{-0.40} / \textcolor{BrickRed}{-2.20} / 66.10\textsuperscript{\textbf{\footnotesize{1}}} & \textcolor{black}{69} / \textcolor{OliveGreen}{+3.00} / \textcolor{OliveGreen}{+3.00} / \textcolor{OliveGreen}{+4.00} / 73.00\textsuperscript{\textbf{\footnotesize{4}}} \\
GPT-3.5-Turbo & \textcolor{black}{71.4} / \textcolor{BrickRed}{-5.50} / \textcolor{BrickRed}{-18.50} / \textcolor{BrickRed}{-17.10} / 71.40\textsuperscript{\textbf{\footnotesize{1}}} & \textcolor{black}{66.2} / \textcolor{BrickRed}{-2.00} / \textcolor{BrickRed}{-14.30} / \textcolor{BrickRed}{-13.20} / 66.20\textsuperscript{\textbf{\footnotesize{1}}} & \textcolor{black}{71.4} / \textcolor{OliveGreen}{+0.10} / \textcolor{BrickRed}{-14.60} / \textcolor{BrickRed}{-10.00} / 71.50\textsuperscript{\textbf{\footnotesize{2}}} \\
Llama 3 8b & \textcolor{black}{61.48} / \textcolor{OliveGreen}{+0.12} / \textcolor{BrickRed}{-10.18} / \textcolor{BrickRed}{-4.68} / 61.60\textsuperscript{\textbf{\footnotesize{2}}} & \textcolor{black}{64.82} / \textcolor{BrickRed}{-17.22} / \textcolor{BrickRed}{-16.42} / \textcolor{BrickRed}{-10.52} / 64.82\textsuperscript{\textbf{\footnotesize{1}}} & \textcolor{black}{66.50} / \textcolor{OliveGreen}{+1.30} / \textcolor{BrickRed}{-7.50} / \textcolor{BrickRed}{-5.20} / 67.80\textsuperscript{\textbf{\footnotesize{2}}} \\
Llama 3.1 8b & \textcolor{black}{53.30} / \textcolor{OliveGreen}{+4.00} / \textcolor{OliveGreen}{+3.70} / \textcolor{OliveGreen}{+10.80} / 64.10\textsuperscript{\textbf{\footnotesize{4}}} & \textcolor{black}{48.50} / \textcolor{BrickRed}{-2.20} / \textcolor{BrickRed}{-26.80} / \textcolor{BrickRed}{-5.00} / 48.50\textsuperscript{\textbf{\footnotesize{1}}} & \textcolor{black}{63.30} / \textcolor{OliveGreen}{+5.30} / \textcolor{BrickRed}{-0.30} / \textcolor{OliveGreen}{+6.80} / 70.10\textsuperscript{\textbf{\footnotesize{4}}} \\
Llama 2 13b & \textcolor{black}{56.19} / \textcolor{OliveGreen}{+5.11} / \textcolor{BrickRed}{-42.89} / \textcolor{OliveGreen}{+9.41} / 65.60\textsuperscript{\textbf{\footnotesize{4}}} & \textcolor{black}{44.58} / \textcolor{OliveGreen}{+11.02} / \textcolor{BrickRed}{-38.08} / \textcolor{BrickRed}{-2.68} / 55.60\textsuperscript{\textbf{\footnotesize{2}}} & \textcolor{black}{47.10} / \textcolor{OliveGreen}{+20.30} / \textcolor{BrickRed}{-26.20} / \textcolor{OliveGreen}{+13.80} / 67.40\textsuperscript{\textbf{\footnotesize{2}}} \\
Llama 2 70b & \textcolor{black}{72.70} / \textcolor{BrickRed}{-11.70} / \textcolor{BrickRed}{-21.50} / \textcolor{BrickRed}{-22.60} / 72.70\textsuperscript{\textbf{\footnotesize{1}}} & \textcolor{black}{59.5} / \textcolor{BrickRed}{-2.45} / \textcolor{BrickRed}{-7.55} / \textcolor{BrickRed}{-8.05} / 59.50\textsuperscript{\textbf{\footnotesize{1}}} & \textcolor{black}{73.58} / \textcolor{BrickRed}{-6.58} / \textcolor{BrickRed}{-21.08} / \textcolor{BrickRed}{-20.28} / 73.58\textsuperscript{\textbf{\footnotesize{1}}} \\
Llama 3 70b & \textcolor{black}{70.49} / \textcolor{BrickRed}{-5.29} / \textcolor{BrickRed}{-6.79} / \textcolor{OliveGreen}{+0.21} / 70.70\textsuperscript{\textbf{\footnotesize{4}}} & \textcolor{black}{67.58} / \textcolor{OliveGreen}{+3.62} / \textcolor{BrickRed}{-0.68} / \textcolor{BrickRed}{-2.48} / \underline{71.20}\textsuperscript{\textbf{\footnotesize{2}}} & \textcolor{black}{70.5} / \textcolor{OliveGreen}{+2.90} / \textcolor{OliveGreen}{+1.80} / \textcolor{OliveGreen}{+3.10} / 73.60\textsuperscript{\textbf{\footnotesize{4}}} \\
Llama 3.1 70b & \textcolor{black}{68.30} / \textcolor{OliveGreen}{+4.30} / \textcolor{BrickRed}{-5.70} / \textcolor{OliveGreen}{+0.70} / 72.60\textsuperscript{\textbf{\footnotesize{2}}} & \textcolor{black}{67.10} / \textcolor{OliveGreen}{+1.10} / \textcolor{BrickRed}{-5.90} / \textcolor{BrickRed}{-1.80} / 68.20\textsuperscript{\textbf{\footnotesize{2}}} & \textcolor{black}{69.80} / \textcolor{OliveGreen}{+0.60} / \textcolor{OliveGreen}{+1.00} / \textcolor{OliveGreen}{+2.20} / 72.00\textsuperscript{\textbf{\footnotesize{4}}} \\
Llama 3.1 405b & \textcolor{black}{61.10} / \textcolor{OliveGreen}{+0.00} / \textcolor{BrickRed}{-2.30} / \textcolor{OliveGreen}{+0.20} / 61.30\textsuperscript{\textbf{\footnotesize{4}}} & \textcolor{black}{40.80} / \textcolor{BrickRed}{-14.70} / \textcolor{BrickRed}{-38.60} / \textcolor{BrickRed}{-29.80} / 40.80\textsuperscript{\textbf{\footnotesize{1}}} & \textcolor{black}{67.10} / \textcolor{OliveGreen}{+0.80} / \textcolor{BrickRed}{-42.50} / \textcolor{BrickRed}{-13.70} / 67.90\textsuperscript{\textbf{\footnotesize{2}}} \\
Mistral 7b & \textcolor{black}{52.98} / \textcolor{BrickRed}{-0.08} / \textcolor{OliveGreen}{+0.22} / \textcolor{OliveGreen}{+0.22} / 53.20\textsuperscript{\textbf{\footnotesize{4}}} & \textcolor{black}{56.71} / \textcolor{OliveGreen}{+0.79} / \textcolor{BrickRed}{-6.71} / \textcolor{BrickRed}{-2.71} / 57.50\textsuperscript{\textbf{\footnotesize{2}}} & \textcolor{black}{54.40} / \textcolor{OliveGreen}{+0.30} / \textcolor{BrickRed}{-6.50} / \textcolor{OliveGreen}{+2.90} / 57.30\textsuperscript{\textbf{\footnotesize{4}}} \\
Mistral NeMo & \textcolor{black}{74.50} / \textcolor{OliveGreen}{+0.20} / \textcolor{BrickRed}{-23.50} / \textcolor{BrickRed}{-23.20} / 74.70\textsuperscript{\textbf{\footnotesize{2}}} & \textcolor{black}{58.20} / \textcolor{OliveGreen}{+5.90} / \textcolor{OliveGreen}{+2.90} / \textcolor{BrickRed}{-3.30} / 64.10\textsuperscript{\textbf{\footnotesize{2}}} & \textcolor{black}{64.30} / \textcolor{BrickRed}{-0.50} / \textcolor{BrickRed}{-7.90} / \textcolor{BrickRed}{-6.20} / 64.30\textsuperscript{\textbf{\footnotesize{1}}} \\
Mixtral 8x7b & \textcolor{black}{58.75} / \textcolor{OliveGreen}{+2.05} / \textcolor{OliveGreen}{+0.75} / \textcolor{OliveGreen}{+10.75} / 69.50\textsuperscript{\textbf{\footnotesize{4}}} & \textcolor{black}{62.47} / \textcolor{BrickRed}{-2.77} / \textcolor{OliveGreen}{+0.93} / \textcolor{OliveGreen}{+0.73} / 63.40\textsuperscript{\textbf{\footnotesize{3}}} & \textcolor{black}{65.30} / \textcolor{BrickRed}{-0.60} / \textcolor{OliveGreen}{+4.70} / \textcolor{OliveGreen}{+5.50} / 70.80\textsuperscript{\textbf{\footnotesize{4}}} \\
Mixtral 8x22b & \textcolor{black}{57.81} / \textcolor{OliveGreen}{+12.29} / \textcolor{BrickRed}{-3.81} / \textcolor{OliveGreen}{+2.19} / 70.10\textsuperscript{\textbf{\footnotesize{2}}} & \textcolor{black}{62.59} / \textcolor{OliveGreen}{+3.51} / \textcolor{OliveGreen}{+0.31} / \textcolor{OliveGreen}{+1.41} / 66.10\textsuperscript{\textbf{\footnotesize{2}}} & \textcolor{black}{51} / \textcolor{OliveGreen}{+13.00} / \textcolor{OliveGreen}{+14.00} / \textcolor{OliveGreen}{+21.00} / 72.00\textsuperscript{\textbf{\footnotesize{4}}} \\
Mistral Medium & \textcolor{black}{57.5} / \textcolor{OliveGreen}{+3.50} / \textcolor{OliveGreen}{+6.50} / \textcolor{OliveGreen}{+9.40} / 66.90\textsuperscript{\textbf{\footnotesize{4}}} & \textcolor{black}{67.6} / \textcolor{BrickRed}{-0.90} / \textcolor{OliveGreen}{+1.50} / \textcolor{BrickRed}{-1.50} / 69.10\textsuperscript{\textbf{\footnotesize{2}}} & \textcolor{black}{59.7} / \textcolor{OliveGreen}{+9.30} / \textcolor{OliveGreen}{+13.60} / \textcolor{OliveGreen}{+15.40} / 75.10\textsuperscript{\textbf{\footnotesize{4}}} \\
Mistral Large 2 & \textcolor{black}{54.20} / \textcolor{OliveGreen}{+14.80} / \textcolor{BrickRed}{-1.30} / \textcolor{OliveGreen}{+10.40} / 69.00\textsuperscript{\textbf{\footnotesize{2}}} & \textcolor{black}{67.70} / \textcolor{OliveGreen}{+0.90} / \textcolor{BrickRed}{-1.10} / \textcolor{BrickRed}{-1.70} / 68.60\textsuperscript{\textbf{\footnotesize{2}}} & \textcolor{black}{71.70} / \textcolor{BrickRed}{-5.60} / \textcolor{OliveGreen}{+1.20} / \textcolor{OliveGreen}{+2.50} / 74.20\textsuperscript{\textbf{\footnotesize{4}}} \\
Phi-3-mini & \textcolor{black}{58.9} / \textcolor{OliveGreen}{+4.80} / \textcolor{BrickRed}{-1.80} / \textcolor{OliveGreen}{+0.80} / 63.70\textsuperscript{\textbf{\footnotesize{2}}} & \textcolor{black}{59.3} / \textcolor{BrickRed}{-1.70} / \textcolor{OliveGreen}{+1.00} / \textcolor{BrickRed}{-6.50} / 60.30\textsuperscript{\textbf{\footnotesize{3}}} & \textcolor{black}{70.5} / \textcolor{BrickRed}{-2.10} / \textcolor{BrickRed}{-4.00} / \textcolor{BrickRed}{-6.40} / 70.50\textsuperscript{\textbf{\footnotesize{1}}} \\
Phi-3-small & \textcolor{black}{55.1} / \textcolor{OliveGreen}{+2.30} / \textcolor{OliveGreen}{+4.50} / \textcolor{OliveGreen}{+5.50} / 60.60\textsuperscript{\textbf{\footnotesize{4}}} & \textcolor{black}{65.4} / \textcolor{BrickRed}{-10.70} / \textcolor{BrickRed}{-0.30} / \textcolor{BrickRed}{-6.10} / 65.40\textsuperscript{\textbf{\footnotesize{1}}} & \textcolor{black}{63.4} / \textcolor{OliveGreen}{+2.90} / \textcolor{OliveGreen}{+9.20} / \textcolor{OliveGreen}{+7.90} / 72.60\textsuperscript{\textbf{\footnotesize{4}}} \\
Phi-3-medium & \textcolor{black}{60.3} / \textcolor{BrickRed}{-2.20} / \textcolor{BrickRed}{-2.90} / \textcolor{BrickRed}{-2.30} / 60.30\textsuperscript{\textbf{\footnotesize{1}}} & \textcolor{black}{62.9} / \textcolor{BrickRed}{-3.80} / \textcolor{OliveGreen}{+2.30} / \textcolor{BrickRed}{-7.40} / 65.20\textsuperscript{\textbf{\footnotesize{3}}} & \textcolor{black}{71.6} / \textcolor{BrickRed}{-1.00} / \textcolor{BrickRed}{-1.20} / \textcolor{OliveGreen}{+0.20} / 71.80\textsuperscript{\textbf{\footnotesize{4}}} \\
Qwen 2 72b & \textcolor{black}{62.40} / \textcolor{OliveGreen}{+5.70} / \textcolor{BrickRed}{-3.90} / \textcolor{BrickRed}{+4.10} / 68.10\textsuperscript{\textbf{\footnotesize{2}}} & \textcolor{black}{63.10} / \textcolor{BrickRed}{-1.40} / \textcolor{BrickRed}{-5.50} / \textcolor{BrickRed}{-3.00} / 63.10\textsuperscript{\textbf{\footnotesize{1}}} & \textcolor{black}{72.80} / \textcolor{OliveGreen}{+1.60} / \textcolor{BrickRed}{-0.10} / \textcolor{OliveGreen}{+3.80} / \underline{76.60}\textsuperscript{\textbf{\footnotesize{4}}} \\ \bottomrule
\end{tabular}
} \\ 
\textbf{Notes:} Results are formatted as: 
BIN-1 result / Change from BIN-2 / Change from BIN-3 / Change from BIN-4 / Highest Value across all prompts. The highest value across all models is \underline{underlined}. The final best value for each model is provided with a superscript \textsuperscript{\textbf{\footnotesize{1, 2, 3, or 4}}} indicating the corresponding prompt number from which the best result was obtained.
\end{table}

{
\subsubsection{Task 1 Invalid Response Rates}
\label{subsec:invalid_response_analysis}
As shown in Figure \ref{fig:invalid_response_percentage}, adherence to instruction formatting varied widely across the evaluated LLMs, with no clear correlation to parameter count. For example, the 8B Llama 3 8b (2.50\%) and 3.8B Phi-3-mini (0.51\%) matched or outperformed larger models like the 70B Llama 3 70b (0.37\%) and GPT-4 (0.02\%). Conversely, smaller architectures such as Phi-2 (2.7B, 26.98\%) and Gemma 2b (2B, 33.53\%) exhibited error rates far exceeding those of many large models, including Mistral Large (19.68\%) and Claude 3 Opus (20.49\%), which frequently appended explanations despite explicit formatting instructions.  

The lowest invalid response rates occurred in newer models: Mistral NeMo (12B, 0.00\%), Mistral Large 2 (0.00\%), GPT-4o mini (0.00\%), and GPT-4 (0.02\%). Mid-sized architectures like Phi-3-medium (14B, 0.07\%) and Gemma 2 27b (27B, 0.35\%) achieved near-zero error rates comparable to much larger systems such as Llama 3 70b (70B, 0.37\%) and GPT-3.5-Turbo (0.37\%).  

Meta’s Llama 3 series improves upon prior generations: Llama 3 70b (0.37\%) and 8b (2.50\%) outperform Llama 2 70b (0.43\%) and 13b (6.37\%). However, the newer Llama 3.1 70b (0.79\%) and Llama 3.1 405b (9.00\%) exhibit higher error rates than their base versions, likely owing to their new ethical guards (Section \ref{subsec:model_related_limitations}). Mistral’s Mistral 7b (7B, 12.29\%) and Mistral Large (19.68\%) are outperformed by successors Mistral NeMo and Mistral Large 2 (both 0.00\%).  

Claude models show substantial variability: Claude 3.5 Sonnet (0.80\%) and Claude 3 Sonnet (1.15\%) achieve lower rates than Claude 3 Haiku (15.90\%) and Claude 3 Opus (20.49\%). Mid-range performers include Gemini 1.0 Pro (1.37\%), Qwen 2 72b (1.46\%), GPT-4-Turbo (1.64\%), Gemma 7b (1.88\%), Phi-3-small (7B, 3.87\%), Llama 3.1 8b (4.82\%), and Mixtral 8x7b (6.08\%).  

The inverse relationship between model recency and error rates holds across most families, though exceptions exist. For example, Llama 3.1 405b (9.00\%) and Llama 3.1 8b (4.82\%) underperform relative to their release timing, diverging from the broader trend.  
}

\begin{figure}[ht]
\centering
\begin{tikzpicture}
    \begin{axis}[
        width=1\linewidth,
        height=0.6\linewidth,
        xlabel={Release Date},
        ylabel={{Rate} of Invalid Responses (Log Scale)},
        xmin=-250, xmax=3900, 
        ymin=0.005, ymax=100,
        ymode=log,
        log ticks with fixed point,
        xmajorgrids=true,
        grid style={line width=0.1pt, draw=gray!20, dash pattern=on 1pt off 1pt},
        xtick={-100, 100, 300, 500, 700, 900, 1100, 1300, 1500, 1700, 1900, 2100, 2300, 2500, 2700, 2900, 3100, 3300, 3500, 3700}, 
        xticklabels={2022-11-28, 2023-03-14, 2023-07-18, 2023-09-27, 2023-11-14, 2023-12-09, 2023-12-12, 2023-12-23, 2024-02-21, 2024-02-26, 2024-03-14, 2024-04-18, 2024-05-13, 2024-06-06, 2024-06-20, 2024-06-27, 2024-07-02, 2024-07-18, 2024-07-23, 2024-07-24},
        x tick label style={rotate=60, anchor=east}, 
        scatter/classes={
            claude={mark=pentagon,draw=cyan,fill=cyan},
            gemma={mark=o,draw=blue,fill=blue},       
            gpt={mark=square,draw=red,fill=red},      
            llama={mark=triangle,draw=green,fill=green}, 
            mistral={mark=diamond,draw=purple,fill=purple}, 
            phi={mark=star,draw=orange,fill=orange},  
            qwen={mark=x,draw=brown,fill=brown}       
        },
    ]
\addplot[only marks,scatter,scatter src=explicit symbolic] 
    coordinates {
        (1900,15.9)[claude]   
        (1900,1.15)[claude]   
        (2700,0.8)[claude]   
        (1900,20.49)[claude]   
        (1500,33.53)[gemma]   
        (1500,1.88)[gemma]   
        (2900,0.52)[gemma]   
        (2900,0.35)[gemma]   
        (1300,1.37)[gemma]   
        (3300,0.01)[gpt]     
        (-100,0.37)[gpt]     
        (2300,0.22)[gpt]     
        (700,1.64)[gpt]     
        (100,0.02)[gpt]        
        (2100,2.5)[llama]   
        (300,6.37)[llama]    
        (300,0.43)[llama]    
        (2100,0.37)[llama]   
        (3500,4.82)[llama]   
        (3500,0.79)[llama]   
        (3500,9)[llama]   
        (500,12.29)[mistral] 
        (3300,0.01)[mistral] 
        (900,6.08)[mistral] 
        (900,12.92)[mistral] 
        (1700,10.26)[mistral] 
        (1700,19.68)[mistral] 
        (3700,0.01)[mistral] 
        (1100,26.98)[phi]     
        (3100,0.51)[phi]     
        (3100,3.87)[phi]     
        (3100,0.07)[phi]     
        (2500,1.46)[qwen]    
    };


    \addplot[red, dashed, smooth, opacity=0.35, tension=0.2] coordinates {(-100, 0.37) (100, 0.02) (700, 1.64)  (2300, 0.22) (3300, 0.01)};
    \addplot[blue, dashed, smooth, opacity=0.35, tension=0.2] coordinates {(1300, 1.37) (1500, 1.88) (1500, 33.53) (2900, 0.52) (2900, 0.35)};
    \addplot[cyan, dashed, opacity=0.35] coordinates {(1900, 1.15) (1900, 15.9) (1900, 20.49) (2700, 0.8)};
    \addplot[orange, tension=0,smooth, dashed, opacity=0.35] coordinates {
    (1100,26.98)
    (3100,0.07)
    (3100,0.51)
    (3100,3.87)
        };
    \addplot[purple,dashed, smooth, opacity=0.35, tension=0.2] coordinates {
        (500,12.29)
        (900,6.08)
        (900,12.92)
        (1700,10.26)
        (1700,19.68)
        (3300,0.01)
        (3700,0.01)
    };
     \addplot[green, dashed, smooth, opacity=0.35, tension=0.2] coordinates {
        (300,0.43)
        (300,6.37)
        (2100,0.37)
        (2100,2.5)
        (3500,0.79)
        (3500,4.82)
        (3500,9)
    };

    \node[anchor=west, font=\tiny, rotate=30] at (axis cs:1900,15.9) {Claude 3 Haiku};
    \node[anchor=east, font=\tiny, rotate=30] at (axis cs:1900,1.15) {Claude 3 Sonnet};
    \node[anchor=east, font=\tiny, rotate=30] at (axis cs:2700,0.8) {Claude 3.5 Sonnet};
    \node[anchor=west, font=\tiny, rotate=30] at (axis cs:1900,20.49) {Claude 3 Opus};
    \node[anchor=east, font=\tiny, rotate=30] at (axis cs:1500,33.53) {Gemma 2b};
    \node[anchor=west, font=\tiny, rotate=30] at (axis cs:1500,1.88) {Gemma 7b};
    \node[anchor=east, font=\tiny, rotate=30] at (axis cs:2900,0.52) {Gemma 2 9b};
    \node[anchor=east, font=\tiny, rotate=30] at (axis cs:2900,0.35) {Gemma 2 27b};
    \node[anchor=east, font=\tiny, rotate=30] at (axis cs:1300,1.37) {Gemini 1.0 Pro};
    \node[anchor=west, font=\tiny, rotate=30] at (axis cs:3300,0.01) {GPT-4o mini};
    \node[anchor=west, font=\tiny, rotate=30] at (axis cs:-100,0.37) {GPT-3.5-Turbo};
    \node[anchor=west, font=\tiny, rotate=30] at (axis cs:2300,0.22) {GPT-4o};
    \node[anchor=east, font=\tiny, rotate=30] at (axis cs:700,1.64) {GPT-4-Turbo};
    \node[anchor=west, font=\tiny, rotate=30] at (axis cs:100,0.02) {GPT-4};
    \node[anchor=west, font=\tiny, rotate=30] at (axis cs:2100,2.5) {Llama 3 8b};
    \node[anchor=east, font=\tiny, rotate=30] at (axis cs:300,6.37) {Llama 2 13b};
    \node[anchor=east, font=\tiny, rotate=30] at (axis cs:300,0.43) {Llama 2 70b};
    \node[anchor=east, font=\tiny, rotate=30] at (axis cs:2100,0.37) {Llama 3 70b};
    \node[anchor=west, font=\tiny, rotate=30] at (axis cs:3500,4.82) {Llama 3.1 8b};
    \node[anchor=west, font=\tiny, rotate=30] at (axis cs:3500,0.79) {Llama 3.1 70b};
    \node[anchor=west, font=\tiny, rotate=30] at (axis cs:3500,9) {Llama 3.1 405b};
    \node[anchor=west, font=\tiny, rotate=30] at (axis cs:500,12.29) {Mistral 7b};
    \node[anchor=south, font=\tiny, rotate=30] at (axis cs:3300,0.01) {Mistral NeMo};
    \node[anchor=east, font=\tiny, rotate=30] at (axis cs:900,6.08) {Mixtral 8x7b};
    \node[anchor=east, font=\tiny, rotate=30] at (axis cs:900,12.92) {Mixtral 8x22b};
    \node[anchor=east, font=\tiny, rotate=30] at (axis cs:1700,10.26) {Mistral Medium};
    \node[anchor=west, font=\tiny, rotate=30] at (axis cs:1700,19.68) {Mistral Large};
    \node[anchor=south, font=\tiny, rotate=30] at (axis cs:3700,0.01) {Mistral Large 2};
    \node[anchor=east, font=\tiny, rotate=30] at (axis cs:1100,26.98) {Phi-2};
    \node[anchor=west, font=\tiny, rotate=30] at (axis cs:3100,0.51) {Phi-3-mini};
    \node[anchor=east, font=\tiny, rotate=30] at (axis cs:3100,3.87) {Phi-3-small};
    \node[anchor=south, font=\tiny, rotate=30] at (axis cs:3100,0.07) {Phi-3-medium};
    \node[anchor=west, font=\tiny, rotate=30] at (axis cs:2500,1.46) {Qwen 2 72b};

    \end{axis}
\end{tikzpicture}
\caption{Invalid Response Statistics Sorted by model release date}
\label{fig:invalid_response_percentage}
\end{figure}

\subsection{Disorder Severity Evaluation ({ Task 2})}
\label{subsec:task_2}

\subsubsection{Results of prompt SEV-4 on the \gls{ZS} Severity Evaluation task}
{An analysis of Table \ref{tab:severity_zs_results} reveals that the performance of each model on the severity task.} The GPT family of models exhibits strong performance across multiple tasks, with GPT-4-Turbo leading the pack. This model achieves the highest \gls{BA} of 39.6\% in the DepSeverity task and an outstanding 59.7\% in DEPTWEET, making it the top performer for these datasets. Its \gls{MAE} scores are notably low, with the minimum being 0.455 in DEPTWEET. The average \gls{BA} and \gls{MAE} for GPT-4-Turbo stand at 41.4\% and 1.047, respectively, highlighting its robustness and reliability. GPT-4 also performs admirably, with a notable \gls{BA} of 40.9\% in the RED SAM Test and 29.0\% in SAD, securing the second-best score in the latter. Consistent \gls{MAE} scores, such as 0.827 in DepSeverity and 0.751 in RED SAM, contribute to an average \gls{BA} of 40.48\% and \gls{MAE} of 1.141, affirming its strong performance across various datasets. While GPT-4o doesn't lead in any specific dataset, it demonstrates competitive results with a \gls{BA} of 35.6\% in DepSeverity and 50.1\% in DEPTWEET, complemented by a low \gls{MAE} of 0.825. The GPT-3.5-Turbo model delivers moderate performance, peaking at 43.2\% \gls{BA} in DEPTWEET and maintaining consistent \gls{MAE} scores, including 0.888 in DepSeverity. With an average \gls{BA} of 31.28\% and \gls{MAE} of 1.120, it aligns closely with GPT-4o, though it doesn't reach the top-tier performance seen in the latest models. The GPT-4o mini, a recent addition to the series, falls slightly behind the larger GPT-4o model, achieving an average \gls{BA} of 34.10\% and \gls{MAE} of 1.230, indicating a slight gap in performance despite its more compact architecture.

The Claude models present a diverse range of capabilities. Claude 3 Sonnet stands out, particularly in the SAD dataset, where it achieves the highest \gls{BA} of 29.7\% and the lowest \gls{MAE} of 1.404. It also performs strongly in DEPTWEET with a \gls{BA} of 45.3\% and an \gls{MAE} of 0.683, resulting in an overall average \gls{BA} of 35.73\% and \gls{MAE} of 0.975. This model demonstrates versatility and consistency across different datasets. Claude 3 Opus matches the highest \gls{BA} in SAD (29.7\%) and excels with a \gls{BA} of 50.6\% in DEPTWEET, along with a low \gls{MAE} of 0.547, further emphasizing its effective performance. In contrast, Claude 3 Haiku shows generally poor results, with low average \gls{BA} and high \gls{MAE}, positioning it among the weaker models in this analysis. Claude 3.5 Sonnet, while consistent in some areas, does not match the performance of Claude 3 Opus or Sonnet in critical metrics, indicating a need for further refinement.

The Gemini and Gemma models offer notable performances, with Gemini 1.0 Pro achieving significant scores, including a \gls{BA} of 54.2\% in DEPTWEET and a second-best \gls{MAE} of 0.729 in the RED SAM Test. Its average scores of 36.63\% \gls{BA} and 1.122 \gls{MAE} reflect a strong overall showing. The Gemma models, particularly Gemma 2 9b, maintain consistent and competitive performance, with an average \gls{BA} of 33.80\% and \gls{MAE} of 1.243, demonstrating their reliability across datasets. Gemma 2 27b also exhibits steady results, achieving an average \gls{BA} of 29.98\% and \gls{MAE} of 1.234, making these models valuable for various applications.

Mistral models present a mixed bag of results, with Mistral Medium standing out for its \gls{BA} of 39.0\% in DepSeverity and the best \gls{MAE} of 0.779. It maintains competitive average scores of 35.45\% \gls{BA} and 1.206 \gls{MAE}, highlighting its consistent performance. Mistral Large and Mistral 7b deliver moderate performances, with \gls{BA} scores of 31.5\% and 31.9\% in DepSeverity, respectively. Mistral NeMo, despite its smaller size, impresses with the best \gls{MAE} of 0.639 in the RED SAM Test and a commendable \gls{MAE} of 1.738 in SAD, resulting in an overall \gls{MAE} of 1.006, which ranks among the best in its class. However, Mistral Large 2 underperforms compared to its predecessors, suggesting a need for optimization.

The Llama models display a broad spectrum of performance levels. Llama 3 70b emerges as the best among them, achieving a \gls{BA} of 33.4\% in DepSeverity and showing competitive scores across other datasets. In contrast, Llama 2 70b underperforms significantly, with a \gls{BA} of just 12.7\% in DepSeverity and high \gls{MAE} scores, indicating challenges in handling the tasks. The newer Llama 3.1 family fails to surpass their Llama 3 counterparts, with the Llama 3.1 405b model performing similarly to the Llama 3.1 70b model, both of which do not meet the performance of the older Llama 3 70b model, reflecting a gap in expected advancements.

The Phi-3 models show varied outcomes. Phi-3-mini demonstrates moderate capabilities with a \gls{BA} of 31.6\% in DepSeverity and reasonable \gls{MAE} scores. Phi-3-small, however, exhibits lower performance, with a \gls{BA} of 25.5\% in the same task. Phi-3-medium performs slightly better, achieving a \gls{BA} of 32.1\% in DepSeverity and maintaining competitive scores across other datasets.

Finally, the Qwen 2 72b model delivers moderate overall performance. While it doesn't achieve standout results in any specific dataset, its performance is comparable to models like Llama 3 70b, Gemma 2 9b, or Phi-3-medium. {This indicates} a competent model that could benefit from further optimization to reach higher performance levels.

\begin{table}[h!]
\caption{Results of Prompt SEV-4 on Task 2 (\gls{ZS}).}
\label{tab:severity_zs_results}
\centering
\begin{tabular}{@{}lrrrrr@{}}
\toprule
\textbf{Model} & \textbf{DepSeverity} & \textbf{DEPTWEET} & \textbf{RED SAM} & \textbf{SAD} & \textbf{Average} \\ \midrule
Claude 3 Haiku & 27.6 / 1.006 & 34.3 / 0.878 & 36.0 / 0.798 & 11.4 / 2.781 & 27.3 / 1.366 \\
Claude 3 Sonnet & 32.8 / 0.960 & 45.3 / 0.683 & 35.1 / 0.853 & \underline{29.7} / \underline{1.404} & 35.7 / \underline{0.975} \\
Claude 3.5 Sonnet & 30.0 / 0.970 & 39.0 / 0.767 & 34.8 / 0.868 & 15.4 / 2.474 & 29.8 / 1.270 \\
Claude 3 Opus & 31.1 / 0.873 & 50.6 / 0.547 & 37.3 / 0.788 & \underline{29.7} / 1.776 & 37.2 / 0.996 \\
Gemma 7b & 30.2 / 0.987 & 34.2 / 0.923 & 33.2 / 0.905 & 15.4 / 3.287 & 28.3 / 1.525 \\
Gemma 2 9b & 33.2 / 0.846 & 50.4 / 0.559 & 34.8 / 0.775 & 16.8 / 2.793 & 33.8 / 1.243 \\ 
Gemma 2 27b & 30.6 / 0.907 & 39.8 / 0.728 & 33.6 / 0.843 & 15.9 / 2.460 & 30.0 / 1.234 \\ 
Gemini 1.0 Pro & 35.6 / 0.794 & 54.2 / 0.496 & 36.1 / 0.729 & 20.6 / 2.468 & 36.6 / 1.122 \\
GPT-4o mini & 35.3 / 0.858 & 45.7 / 0.649 & 36.5 / 0.813 & 18.9 / 2.559 & 34.1 / 1.230 \\
GPT-3.5-Turbo & 28.8 / 0.888 & 43.2 / 0.647 & 34.4 / 0.798 & 18.7 / 2.145 & 31.3 / 1.120 \\
GPT-4o & 35.6 / 0.825 & 50.1 / 0.577 & 38.5 / 0.762 & 18.2 / 2.671 & 35.6 / 1.209 \\
GPT-4-Turbo & \underline{39.6} / 0.805 & \underline{59.7} / \underline{0.455} & 39.5 / 0.757 & 26.8 / 2.171 & \underline{41.4} / 1.047 \\
GPT-4 & 39.3 / 0.827 & 52.7 / 0.554 & \underline{40.9} / 0.751 & 29.0 / 2.430 & 40.5 / 1.141 \\
Llama 3 8b & 26.1 / 0.929 & 37.0 / 0.769 & 31.4 / 0.769 & 21.2 / 1.655 & 28.9 / 1.281 \\
Llama 3.1 8b & 27.3 / 0.951 & 29.4 / 1.055 & 30.1 / 0.816 & 10.3 / 3.669 & 24.3 / 1.623 \\
Llama 2 13b & 26.3 / 1.045 & 32.8 / 0.924 & 30.6 / 0.966 & 8.6 / 3.932 & 24.6 / 1.717 \\
Llama 2 70b & 12.7 / 1.895 & 26.7 / 1.243 & 15.6 / 1.490 & 10.7 / 3.675 & 16.4 / 2.076 \\
Llama 3 70b & 33.4 / 0.819 & 42.3 / 0.674 & 33.5 / 0.832 & 16.1 / 2.648 & 31.3 / 1.243 \\
Llama 3.1 70b & 33.8 / 0.853 & 42.4 / 0.712 & 33.0 / 0.868 & 13.6 / 3.354 & 30.7 / 1.447 \\
Llama 3.1 405b & 33.0 / 0.877 & 41.8 / 0.812 & 31.6 / 0.900 & 13.6 / 2.890 & 30.0 / 1.370 \\
Mistral 7b & 31.9 / 1.055 & 44.4 / 0.767 & 37.6 / 0.797 & 14.6 / 2.811 & 32.1 / 1.358 \\ 
Mistral NeMo & 27.2 / 0.871 & 32.0 / 0.775 & 36.7 / \underline{0.639} & 15.3 / 1.738 & 27.8 / 1.006 \\ 
Mixtral 8x7b & 35.0 / 0.939 & 44.7 / 0.789 & 32.8 / 0.871 & 20.2 / 2.180 & 33.2 / 1.195 \\
Mixtral 8x22b & 33.5 / 0.864 & 44.2 / 0.648 & 36.6 / 0.795 & 23.4 / 2.369 & 34.4 / 1.169 \\
Mistral Medium & 39.0 / \underline{0.779} & 51.9 / 0.556 & 36.8 / 0.770 & 14.1 / 2.720 & 35.5 / 1.206 \\
Mistral Large & 31.5 / 0.898 & 43.2 / 0.680 & 35.2 / 0.807 & 16.8 / 2.566 & 31.7 / 1.238 \\
Mistral Large 2 & 29.6 / 0.865 & 42.1 / 0.666 & 34.1 / 0.811 & 10.4 / 2.942 & 29.1 / 1.321 \\
Phi-3-mini & 31.6 / 1.015 & 40.7 / 0.865 & 35.1 / 0.865 & 13.6 / 3.011 & 30.3 / 1.439 \\
Phi-3-small & 25.5 / 0.897 & 31.9 / 0.786 & 33.6 / 0.757 & 9.6 / 2.691 & 25.2 / 1.282 \\
Phi-3-medium & 32.1 / 0.883 & 39.5 / 0.754 & 33.2 / 0.839 & 12 / 2.283 & 29.2 / 1.190 \\
Qwen 2 72b & 36.7 / 0.921 & 44.5 / 0.734 & 36.4 / 0.857 & 18.5 / 2.556 & 34.0 / 1.267 \\
\bottomrule
\end{tabular} \\
\textbf{Notes:}
The results are written as follows: \gls{BA} / \gls{MAE}. The best \gls{BA} and \gls{MAE} for each dataset are \underline{underlined} separately. Each dataset may have different models underlined for \gls{BA} and \gls{MAE}.
\end{table}

\subsubsection{Results of prompt SEV-4 on the \gls{FS} Severity Evaluation task}
An analysis of Table \ref{tab:severity_fs_results} reveals that most models generally improved on various datasets with the \gls{FS} prompt. However, some models experienced minor decreases. Notably, even if a model improves across all datasets except SAD, the final average change might be negative due to SAD having the most significant impact, given its 10 severity levels.

Claude 3.5 Sonnet showed significant improvements, with an overall average improvement of +6.50\% in \gls{BA} and a reduction of -0.248 in \gls{MAE}. In contrast, Claude 3 Sonnet experienced mixed results, with a +5.30\% increase in \gls{BA} and a -0.058 reduction in \gls{MAE} in DEPTWEET, but a substantial decrease in SAD, resulting in an overall negative change.

Gemma models, particularly Gemma 2 27b, exhibited substantial positive changes, leading to an overall average improvement of +6.05\% in \gls{BA} and a reduction of -0.191 in \gls{MAE}. Similarly, Gemma 2 9b showed moderate improvements with an overall +2.175\% in \gls{BA} and a reduction of -0.131 in \gls{MAE}.

GPT models demonstrated positive trends, with GPT-4o showing consistent improvements across all metrics, resulting in an overall average improvement of +2.975\% in \gls{BA} and a reduction of -0.117 in \gls{MAE}. GPT-4-Turbo had minor positive changes overall. GPT-4o mini experienced the greatest improvement of all the GPT models (+4.025\% and -0.170).

Mistral models, particularly Mistral 7b, showed significant improvements with an overall average improvement of +6.52\% in \gls{BA} and a reduction of -0.345 in \gls{MAE}. Mixtral models, such as Mixtral 8x7b, also performed well with an overall average improvement of +4.96\% in \gls{BA} and a reduction of -0.140 in \gls{MAE}.

Some models experienced mixed or negative overall changes. Claude 3 Opus saw minor improvements but overall negative performance due to decreases in the RED SAM Test and SAD, leading to an overall average decrease of -2.70\% in \gls{BA} and an increase of +0.096 in \gls{MAE}. Llama 2 13b suffered a major decrease in DepSeverity, resulting in an overall negative change.

Phi-3 models performed well with the \gls{FS} prompt. Phi-3-mini showed significant improvements across all metrics, leading to an overall average improvement of +8.18\% in \gls{BA} and a reduction of -0.488 in \gls{MAE}. Phi-3-medium also demonstrated consistent improvements.

Overall, utilizing the \gls{FS} prompt with three examples generally enhanced performance across most models and datasets. While incorporating \gls{FS} prompts can yield significant improvements, the degree of enhancement varies depending on the specific models and datasets. SAD, with its 10 severity levels, can disproportionately influence the final average, sometimes counteracting improvements observed in other datasets. Models like Claude 3.5 Sonnet, Gemma 2 27b, and Mixtral 8x7b demonstrated the most consistent and substantial enhancements. Notably, due to the inclusion of three additional example posts (in addition to the one provided for evaluation), the overall token count per evaluation is typically multiplied by 3-4.

With these enhancements, GPT-4-Turbo now excels in \gls{MAE} across all datasets except RED SAM, where Gemini 1.0 Pro became the best. In terms of \gls{BA}, the notable changes include Gemini 1.0 Pro's dominance in RED SAM and Mistral 7b's improved performance, making it the leading model for the SAD dataset.

\begin{table}[h!]
\caption{Results of Prompt SEV-4 on Task 2 (FS)}
\label{tab:severity_fs_results}
\centering
\begin{tabular}{@{}lrrrrr@{}}
\toprule
\textbf{Model} & \textbf{DepSeverity} & \textbf{DEPTWEET} & \textbf{RED SAM} & \textbf{SAD} & \textbf{Average} \\ \midrule
Claude 3 Haiku & \textcolor{OliveGreen}{+0.30} / \textcolor{OliveGreen}{-0.059} & \textcolor{BrickRed}{-1.50} / \textcolor{OliveGreen}{-0.032} & \textcolor{OliveGreen}{+0.10} / \textcolor{OliveGreen}{-0.080} & \textcolor{BrickRed}{-2.10} / \textcolor{BrickRed}{+0.345} & \textcolor{BrickRed}{-0.80} / \textcolor{BrickRed}{+0.043} \\
Claude 3 Sonnet & \textcolor{OliveGreen}{+1.80} / \textcolor{OliveGreen}{-0.042} & \textcolor{OliveGreen}{+5.30} / \textcolor{OliveGreen}{-0.058} & \textcolor{OliveGreen}{+0.40} / \textcolor{OliveGreen}{-0.055} & \textcolor{BrickRed}{-10.4} / \textcolor{BrickRed}{+1.101} & \textcolor{BrickRed}{-0.73} / \textcolor{BrickRed}{+0.237} \\
Claude 3.5 Sonnet & \textcolor{OliveGreen}{+1.50} / \textcolor{OliveGreen}{-0.089} & \textcolor{OliveGreen}{+17.2} / \textcolor{OliveGreen}{-0.274} & \textcolor{BrickRed}{-0.40} / \textcolor{OliveGreen}{-0.053} & \textcolor{OliveGreen}{+7.70} / \textcolor{OliveGreen}{-0.575} & \textcolor{OliveGreen}{+6.50} / \textcolor{OliveGreen}{-0.248} \\
Claude 3 Opus  & \textcolor{OliveGreen}{+0.90} / \textcolor{BrickRed}{+0.025} & \textcolor{OliveGreen}{+0.80} / \textcolor{OliveGreen}{-0.003} & \textcolor{BrickRed}{-3.40} / \textcolor{BrickRed}{+0.031} & \textcolor{BrickRed}{-9.10} / \textcolor{BrickRed}{+0.332} & \textcolor{BrickRed}{-2.70} / \textcolor{BrickRed}{+0.096} \\
Gemma 7b & \textcolor{BrickRed}{-1.00} / \textcolor{OliveGreen}{-0.060} & \textcolor{OliveGreen}{+5.50} / \textcolor{OliveGreen}{-0.1530} & \textcolor{BrickRed}{-1.170} / \textcolor{BrickRed}{+0.029} & \textcolor{OliveGreen}{+6.33} / \textcolor{OliveGreen}{-0.596} & \textcolor{OliveGreen}{+2.415} / \textcolor{OliveGreen}{-0.195} \\
Gemma 2 9b & \textcolor{OliveGreen}{+1.00} / \textcolor{OliveGreen}{-0.008} & \textcolor{OliveGreen}{+5.80} / \textcolor{OliveGreen}{-0.068} & \textcolor{OliveGreen}{+2.50} / \textcolor{OliveGreen}{-0.045} & \textcolor{BrickRed}{-0.60} / \textcolor{OliveGreen}{-0.419} & \textcolor{OliveGreen}{+2.175} / \textcolor{OliveGreen}{-0.131} \\
Gemma 2 27b & \textcolor{OliveGreen}{+3.50} / \textcolor{OliveGreen}{-0.119} & \textcolor{OliveGreen}{+12.5} / \textcolor{OliveGreen}{-0.196} & \textcolor{OliveGreen}{+3.10} / \textcolor{OliveGreen}{-0.083} & \textcolor{OliveGreen}{+5.10} / \textcolor{OliveGreen}{-0.366} & \textcolor{OliveGreen}{+6.05} / \textcolor{OliveGreen}{-0.191} \\
Gemini 1.0 Pro & \textcolor{OliveGreen}{+2.70} / \textcolor{BrickRed}{+0.007} & \textcolor{BrickRed}{-1.40} / \textcolor{BrickRed}{+0.009} & \textcolor{OliveGreen}{+7.4} / \textcolor{OliveGreen}{-0.086} & \textcolor{OliveGreen}{+2.60} / \textcolor{OliveGreen}{-0.604} & \textcolor{OliveGreen}{+2.825} / \textcolor{OliveGreen}{-0.168} \\
GPT-4o mini & \textcolor{OliveGreen}{+1.70} / \textcolor{OliveGreen}{-0.031} & \textcolor{OliveGreen}{+7.10} / \textcolor{OliveGreen}{-0.113} & \textcolor{OliveGreen}{+3.40} / \textcolor{OliveGreen}{-0.060} & \textcolor{OliveGreen}{+3.90} / \textcolor{OliveGreen}{-0.475} & \textcolor{OliveGreen}{+4.025} / \textcolor{OliveGreen}{-0.170} \\
GPT-3.5-Turbo & \textcolor{OliveGreen}{+4.20} / \textcolor{OliveGreen}{-0.001} & \textcolor{BrickRed}{-1.30} / \textcolor{BrickRed}{+0.043} & \textcolor{OliveGreen}{+3.00} / \textcolor{BrickRed}{+0.003} & \textcolor{OliveGreen}{+0.00} / \textcolor{BrickRed}{+0.239} & \textcolor{OliveGreen}{+1.475} / \textcolor{BrickRed}{+0.071} \\
GPT-4o & \textcolor{OliveGreen}{+2.90} / \textcolor{OliveGreen}{-0.023} & \textcolor{OliveGreen}{+5.50} / \textcolor{OliveGreen}{-0.047} & \textcolor{OliveGreen}{+2.30} / \textcolor{OliveGreen}{-0.044} & \textcolor{OliveGreen}{+1.20} / \textcolor{OliveGreen}{-0.354} & \textcolor{OliveGreen}{+2.975} / \textcolor{OliveGreen}{-0.117} \\
GPT-4-Turbo & \textcolor{OliveGreen}{+3.60} / \textcolor{OliveGreen}{-0.022} & \textcolor{OliveGreen}{+1.20} / \textcolor{OliveGreen}{-0.015} & \textcolor{OliveGreen}{+3.50} / \textcolor{OliveGreen}{-0.083} & \textcolor{BrickRed}{-5.20} / \textcolor{OliveGreen}{-0.724} & \textcolor{OliveGreen}{+0.78} / \textcolor{OliveGreen}{-0.211} \\
GPT-4 & \textcolor{OliveGreen}{+2.30} / \textcolor{OliveGreen}{-0.042} & \textcolor{OliveGreen}{+8.50} / \textcolor{OliveGreen}{-0.100} & \textcolor{OliveGreen}{+2.40} / \textcolor{OliveGreen}{-0.090} & \textcolor{BrickRed}{-10.5} / \textcolor{OliveGreen}{-0.244} & \textcolor{OliveGreen}{+0.68} / \textcolor{OliveGreen}{-0.119} \\
Mistral 7b & \textcolor{OliveGreen}{+3.90} / \textcolor{OliveGreen}{-0.103} & \textcolor{OliveGreen}{+7.97} / \textcolor{OliveGreen}{-0.131} & \textcolor{OliveGreen}{+1.32} / \textcolor{OliveGreen}{-0.078} & \textcolor{OliveGreen}{+12.9} / \textcolor{OliveGreen}{-1.066} & \textcolor{OliveGreen}{+6.52} / \textcolor{OliveGreen}{-0.345} \\
Mistral NeMo & \textcolor{BrickRed}{-1.50} / \textcolor{BrickRed}{+0.045} & \textcolor{OliveGreen}{+3.50} / \textcolor{OliveGreen}{-0.066} & \textcolor{BrickRed}{-0.60} / \textcolor{BrickRed}{+0.007} & \textcolor{OliveGreen}{+0.40} / \textcolor{BrickRed}{+0.257} & \textcolor{OliveGreen}{+0.45} / \textcolor{BrickRed}{+0.061} \\
Mixtral 8x7b & \textcolor{BrickRed}{-0.10} / \textcolor{OliveGreen}{-0.077} & \textcolor{OliveGreen}{+8.00} / \textcolor{OliveGreen}{-0.169} & \textcolor{OliveGreen}{+8.10} / \textcolor{OliveGreen}{-0.154} & \textcolor{OliveGreen}{+3.83} / \textcolor{OliveGreen}{-0.159} & \textcolor{OliveGreen}{+4.96} / \textcolor{OliveGreen}{-0.140} \\
Mixtral 8x22b & \textcolor{OliveGreen}{+2.10} / \textcolor{OliveGreen}{-0.035} & \textcolor{OliveGreen}{+8.90} / \textcolor{OliveGreen}{-0.093} & \textcolor{OliveGreen}{+2.92} / \textcolor{OliveGreen}{-0.045} & \textcolor{BrickRed}{-0.37} / \textcolor{OliveGreen}{-0.539} & \textcolor{OliveGreen}{+3.39} / \textcolor{OliveGreen}{-0.178} \\
Mistral Medium & \textcolor{BrickRed}{-1.80} / \textcolor{BrickRed}{+0.049} & \textcolor{BrickRed}{-3.60} / \textcolor{BrickRed}{+0.035} & \textcolor{OliveGreen}{+5.60} / \textcolor{OliveGreen}{-0.095} & \textcolor{OliveGreen}{+8.80} / \textcolor{OliveGreen}{-0.836} & \textcolor{OliveGreen}{+2.25} / \textcolor{OliveGreen}{-0.212} \\
Mistral Large & \textcolor{OliveGreen}{+1.20} / \textcolor{OliveGreen}{-0.027} & \textcolor{OliveGreen}{+7.70} / \textcolor{OliveGreen}{-0.126} & \textcolor{OliveGreen}{+2.20} / \textcolor{OliveGreen}{-0.077} & \textcolor{BrickRed}{-4.10} / \textcolor{OliveGreen}{-0.094} & \textcolor{OliveGreen}{+1.75} / \textcolor{OliveGreen}{-0.081} \\
Mistral Large 2 123b & \textcolor{OliveGreen}{+0.80} / \textcolor{OliveGreen}{-0.033} & \textcolor{OliveGreen}{+6.00} / \textcolor{OliveGreen}{-0.105} & \textcolor{OliveGreen}{+3.00} / \textcolor{OliveGreen}{-0.098} & \textcolor{OliveGreen}{+2.90} / \textcolor{OliveGreen}{-0.621} & \textcolor{OliveGreen}{+3.175} / \textcolor{OliveGreen}{-0.214} \\
Llama 3 8b & \textcolor{OliveGreen}{+0.10} / \textcolor{OliveGreen}{-0.046} & \textcolor{BrickRed}{-5.09} / \textcolor{BrickRed}{+0.097} & \textcolor{OliveGreen}{+1.90} / \textcolor{OliveGreen}{-0.096} & \textcolor{BrickRed}{-1.97} / \textcolor{OliveGreen}{-0.023} & \textcolor{BrickRed}{-1.27} / \textcolor{OliveGreen}{-0.017} \\
Llama 3.1 8b & \textcolor{OliveGreen}{+1.80} / \textcolor{OliveGreen}{-0.083} & \textcolor{OliveGreen}{+3.50} / \textcolor{OliveGreen}{-0.173} & \textcolor{OliveGreen}{+4.10} / \textcolor{OliveGreen}{-0.129} & \textcolor{OliveGreen}{+3.00} / \textcolor{OliveGreen}{-0.876} & \textcolor{OliveGreen}{+3.10} / \textcolor{OliveGreen}{-0.315} \\
Llama 2 13b & \textcolor{BrickRed}{-24.4} / \textcolor{BrickRed}{+1.379} & \textcolor{OliveGreen}{+6.80} / \textcolor{OliveGreen}{-0.077} & \textcolor{OliveGreen}{+0.16} / \textcolor{OliveGreen}{-0.049} & \textcolor{OliveGreen}{+3.17} / \textcolor{BrickRed}{+0.179} & \textcolor{BrickRed}{-3.57} / \textcolor{BrickRed}{+0.358} \\
Llama 2 70b & \textcolor{BrickRed}{-5.90} / \textcolor{BrickRed}{+0.363} & \textcolor{OliveGreen}{+4.00} / \textcolor{OliveGreen}{-0.113} & \textcolor{BrickRed}{-3.19} / \textcolor{BrickRed}{+0.115} & \textcolor{OliveGreen}{+9.61} / \textcolor{OliveGreen}{-1.274} & \textcolor{OliveGreen}{+1.13} / \textcolor{OliveGreen}{-0.227} \\
Llama 3 70b & \textcolor{OliveGreen}{+1.10} / \textcolor{OliveGreen}{-0.023} & \textcolor{OliveGreen}{+9.00} / \textcolor{OliveGreen}{-0.129} & \textcolor{OliveGreen}{+3.22} / \textcolor{OliveGreen}{-0.089} & \textcolor{OliveGreen}{+7.32} / \textcolor{OliveGreen}{-0.687} & \textcolor{OliveGreen}{+5.16} / \textcolor{OliveGreen}{-0.232} \\
Llama 3.1 70b & \textcolor{OliveGreen}{+3.10} / \textcolor{OliveGreen}{-0.062} & \textcolor{OliveGreen}{+10.00} / \textcolor{OliveGreen}{-0.182} & \textcolor{OliveGreen}{+3.10} / \textcolor{OliveGreen}{-0.077} & \textcolor{OliveGreen}{+4.80} / \textcolor{OliveGreen}{-0.808} & \textcolor{OliveGreen}{+5.25} / \textcolor{OliveGreen}{-0.282} \\
Llama 3.1 405b & \textcolor{OliveGreen}{+2.00} / \textcolor{OliveGreen}{-0.078} & \textcolor{OliveGreen}{+8.60} / \textcolor{OliveGreen}{-0.203} & \textcolor{OliveGreen}{+4.10} / \textcolor{OliveGreen}{-0.150} & \textcolor{OliveGreen}{+3.60} / \textcolor{OliveGreen}{-0.524} & \textcolor{OliveGreen}{+4.58} / \textcolor{OliveGreen}{-0.239} \\
Phi 3 Mini & \textcolor{OliveGreen}{+6.80} / \textcolor{OliveGreen}{-0.202} & \textcolor{OliveGreen}{+9.70} / \textcolor{OliveGreen}{-0.238} & \textcolor{OliveGreen}{+5.40} / \textcolor{OliveGreen}{-0.213} & \textcolor{OliveGreen}{+10.8} / \textcolor{OliveGreen}{-1.298} & \textcolor{OliveGreen}{+8.18} / \textcolor{OliveGreen}{-0.488} \\
Phi 3 Small & \textcolor{OliveGreen}{+5.00} / \textcolor{OliveGreen}{-0.034} & \textcolor{OliveGreen}{+8.00} / \textcolor{OliveGreen}{-0.122} & \textcolor{OliveGreen}{+2.80} / \textcolor{OliveGreen}{-0.032} & \textcolor{OliveGreen}{+3.90} / \textcolor{BrickRed}{+0.253} & \textcolor{OliveGreen}{+4.93} / \textcolor{BrickRed}{+0.016} \\
Phi 3 Medium & \textcolor{OliveGreen}{+3.70} / \textcolor{OliveGreen}{-0.046} & \textcolor{OliveGreen}{+6.50} / \textcolor{OliveGreen}{-0.107} & \textcolor{OliveGreen}{+3.50} / \textcolor{OliveGreen}{-0.061} & \textcolor{OliveGreen}{+6.70} / \textcolor{OliveGreen}{-0.128} & \textcolor{OliveGreen}{+5.10} / \textcolor{OliveGreen}{-0.086} \\ 
Qwen 2 72b & \textcolor{OliveGreen}{+4.90} / \textcolor{OliveGreen}{-0.113} & \textcolor{OliveGreen}{+7.70} / \textcolor{OliveGreen}{-0.158} & \textcolor{OliveGreen}{+1.80} / \textcolor{OliveGreen}{-0.039} & \textcolor{OliveGreen}{+7.80} / \textcolor{OliveGreen}{-0.347} & \textcolor{OliveGreen}{+5.55} / \textcolor{OliveGreen}{-0.164} \\
\bottomrule
\end{tabular} \\
\textbf{Notes:} The results are written as follows: change to \gls{BA} / \gls{MAE}.
\end{table}

\subsection{Psychiatric Knowledge Assessment ({ Task 3})}
\label{subsec:task_3}
\begin{figure}[h!]
\centering
\includegraphics[width=1\linewidth]{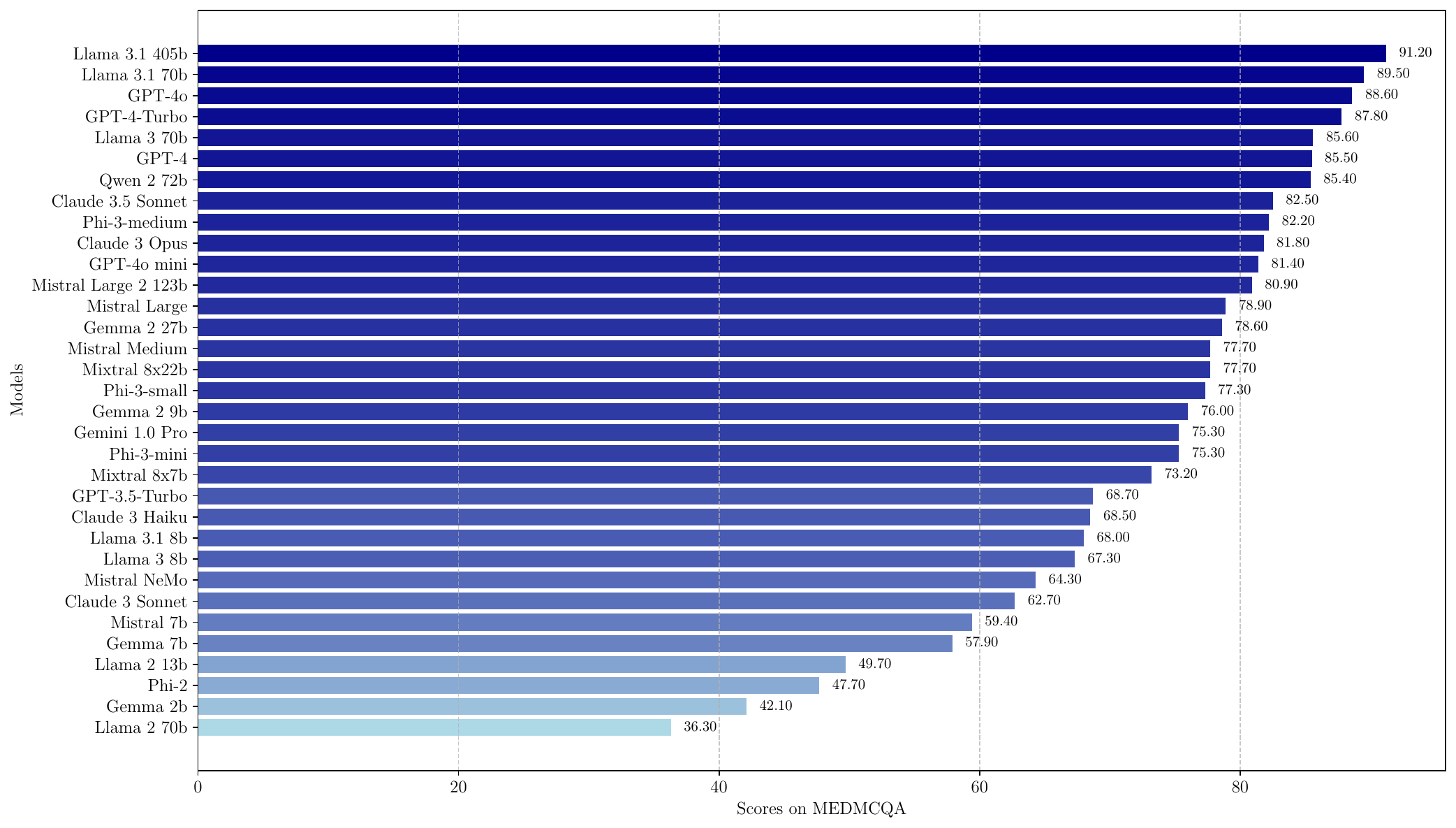}
\caption{Results of all models on Task 3 { on the MedMCQA dataset}}
\label{fig:task_3_results}
\end{figure}

The previous experiments have highlighted each model's effectiveness in classifying mental illnesses from social media posts. However, we still lack a fundamental understanding of each model's inherent knowledge of psychiatry and various mental disorders. The results in figure \ref{fig:task_3_results} demonstrate the foundational psychiatric knowledge that each model possesses.

Llama 3.1 405b tops the performance chart, achieving an impressive accuracy of 91.2\%, followed closely by Llama 3.1 70b with an accuracy of 89.50\%. The smaller Llama 3.1 8b, however, lags behind with an accuracy of 68.0\%. Despite this, it outperforms other models in its size category (under 14 billion parameters) such as Mistral 7b, Gemma 7b, and Phi-2, with a lead ranging from 8\% to 20\%. The older Llama 3 models all perform slightly worse compared to the more recently released models (e.g. Llama 3 70b actually beats GPT-4 even if only by 0.10\%). Interestingly, it's only in this task, which includes raw factual querying of the models, that the strength of the {Llama} 3.1 family and {specifically} the {Llama} 3.1 405b models stand out. 

GPT-4o follows behind the {Llama} 3.1 models with an accuracy 88.6\%, followed closely by GPT-4-Turbo at 87.8\%, and GPT-4 at 85.5\%. It's clear that each successive iteration of the model, even while smaller and more efficient, is better trained and has a better overall understanding of psychiatric knowledge.

Claude 3.5 Sonnet follows, nearly matched by the Phi-3-medium model with accuracies of 82.5\% and 82.2\%, respectively. Both models are impressive for their size, but the Phi-3-medium model is especially notable, considering its relatively small size of 14 billion parameters, yet matching much larger models.

Claude 3 Opus slightly surpasses Mistral Large with an accuracy of 81.8\% compared to 78.9\%. The Gemma 2 27b follows closely with an accuracy of 78.6\%, nearly matching Mistral Large despite its much smaller size.

Interestingly, the Mistral Medium model follows closely with an accuracy of 77.7\%, matching the Mixtral 8x22b. Mistral Large 2 beats the previous Mistral Large and medium models but still loses out to the Claude 3 Opus model, and even to the GPT-4o mini model.

Phi-3-small, Gemma 2 9b, and Phi-3-mini follow in that order with accuracy scores of 77.3\%, 76.0\%, and 75.3\%. Interestingly, Gemini 1.0 Pro's accuracy matches that of Phi-3-mini at 75.3\%, despite its much larger size.

Other models such as Mixtral 8x7b and GPT-3.5-Turbo show competitive performance, with accuracies of 75.3\%, 73.2\%, and 68.7\%, respectively.

Claude 3 Haiku achieves an accuracy of 68.5\%, while Claude 3 Sonnet performs worse with 62.7\%. The Llama 2 models, particularly the 13b and 70b variants, struggle significantly in the knowledge task, achieving accuracies of 49.7\% and 36.3\%, respectively. Phi-2 also shows a subpar performance with an accuracy of 47.7\%.

The general trend observed from these results is that while model size matters, more recent models perform significantly better than their older counterparts, even when there is a massive size differential (e.g., Phi-3-medium and Llama 3.1 70b, or the many other examples mentioned earlier). For models released around the same time, model size appears to be the deciding factor, with inconsistencies observed earlier (such as Gemma 2 9b outperforming Gemma 2 27b or Phi-3-mini outperforming Phi-2-small) being nearly absent in this task. It is likely that the more recent models were trained on larger and more diverse datasets, especially those including psychiatric data, thus enhancing their foundational knowledge in psychiatry.

\begin{figure}[ht]
\centering
\begin{tikzpicture}
    \begin{axis}[
        width=1\linewidth,
        height=0.6\linewidth,
        xlabel={Release Date},
        ylabel={Accuracy},
        xmin=-250, xmax=3900, 
        ymin=30, ymax=100,
        grid=both,
        grid style={line width=0.1pt, draw=gray!20, dash pattern=on 1pt off 1pt},
        xtick={-100, 100, 300, 500, 700, 900, 1100, 1300, 1500, 1700, 1900, 2100, 2300, 2500, 2700, 2900, 3100, 3300, 3500, 3700}, 
        xticklabels={2022-11-28, 2023-03-14, 2023-07-18, 2023-09-27, 2023-11-14, 2023-12-09, 2023-12-12, 2023-12-23, 2024-02-21, 2024-02-26, 2024-03-14, 2024-04-18, 2024-05-13, 2024-06-06, 2024-06-20, 2024-06-27, 2024-07-02, 2024-07-18, 2024-07-23, 2024-07-24},
        x tick label style={rotate=60, anchor=east}, 
        scatter/classes={
            claude={mark=pentagon,draw=cyan,fill=cyan},
            gemma={mark=o,draw=blue,fill=blue},       
            gpt={mark=square,draw=red,fill=red},      
            llama={mark=triangle,draw=green,fill=green}, 
            mistral={mark=diamond,draw=purple,fill=purple}, 
            phi={mark=star,draw=orange,fill=orange},  
            qwen={mark=x,draw=brown,fill=brown}       
        },
        legend style={at={(0.83,0.36125)}, anchor=north west},
            legend cell align={left},
            legend entries={Claude, Gemma/Gemini, GPT, Llama, Mistral, Phi, Qwen},
        every axis/.append style={
            font=\small,
            label style={font=\small},
            tick label style={font=\small}
        }
    ]
    \addplot[only marks,scatter,scatter src=explicit symbolic] 
    coordinates {
        (1900,68.5)[claude]   
        (1900,62.7)[claude]   
        (2700,82.5)[claude]   
        (1900,81.8)[claude]   
        (1500,42.1)[gemma]   
        (1500,57.9)[gemma]   
        (2900,76.0)[gemma]   
        (2900,78.6)[gemma]   
        (1300,75.3)[gemma]   
        (3300,81.4)[gpt]     
        (-100,68.7)[gpt]     
        (2300,88.6)[gpt]     
        (700,87.8)[gpt]     
        (100,85.5)[gpt]        
        (2100,67.3)[llama]   
        (300,49.7)[llama]    
        (300,36.3)[llama]    
        (2100,85.6)[llama]   
        (3500,68.0)[llama]   
        (3500,89.5)[llama]   
        (3500,91.2)[llama]   
        (500,59.4)[mistral] 
        (3300,64.3)[mistral] 
        (900,73.2)[mistral] 
        (900,77.7)[mistral] 
        (1700,77.7)[mistral] 
        (1700,78.9)[mistral] 
        (3700,80.9)[mistral] 
        (1100,47.7)[phi]     
        (3100,75.3)[phi]     
        (3100,77.3)[phi]     
        (3100,82.2)[phi]     
        (2500,85.4)[qwen]    
    };

    Add curved, colored, and dashed lines connecting the models
    \addplot[cyan, dashed, smooth, opacity=0.35, tension=0.2] coordinates {(1900, 62.7) (1900, 68.5) (1900, 81.8) (2700, 82.5)};
    \addplot[blue, dashed, smooth, opacity=0.35, tension=0.2] coordinates {(1300, 75.3) (1500, 42.1) (1500, 57.9) (2900, 76.0) (2900, 78.6)};
    \addplot[red, dashed, smooth, opacity=0.35, tension=0.2] coordinates {(-100, 68.7) (100, 85.5) (700, 87.8)  (2300, 88.6) (3300, 81.4)};
    \addplot[green, dashed, smooth, opacity=0.35, tension=0.2] coordinates {(300, 36.3) (300, 49.7) (2100, 67.3) (2100, 85.6) (3500, 68.0) (3500, 89.5) (3500, 91.2)};
    \addplot[purple, dashed, smooth, opacity=0.35, tension=0.2] coordinates {(500, 59.4) (900, 73.2) (900, 77.7) (1700, 77.7) (1700, 78.9) (3300, 64.3) (3700, 80.9)};
    \addplot[orange, dashed, smooth, opacity=0.35, tension=0] coordinates {(1100, 47.7) (3100, 75.3) (3100, 77.3) (3100, 82.2)};

    \node[anchor=west, font=\tiny, rotate=30] at (axis cs:1900,68.5) {Claude 3 Haiku};
    \node[anchor=north, font=\tiny, rotate=30] at (axis cs:1900,62.7) {Claude 3 Sonnet};
    \node[anchor=east, font=\tiny, rotate=30] at (axis cs:2700,82.5) {Claude 3.5 Sonnet};
    \node[anchor=north, font=\tiny, rotate=30] at (axis cs:1900,81.8) {Claude 3 Opus};
    \node[anchor=east, font=\tiny, rotate=30] at (axis cs:1500,42.1) {Gemma 2b};
    \node[anchor=east, font=\tiny, rotate=30] at (axis cs:1500,57.9) {Gemma 7b};
    \node[anchor=east, font=\tiny, rotate=30] at (axis cs:2900,76.0) {Gemma 2 9b};
    \node[anchor=east, font=\tiny, rotate=30] at (axis cs:2900,78.6) {Gemma 2 27b};
    \node[anchor=south, font=\tiny, rotate=30] at (axis cs:1300,75.3) {Gemini 1.0 Pro};
    \node[anchor=west, font=\tiny, rotate=30] at (axis cs:3300,81.4) {GPT-4o mini};
    \node[anchor=west, font=\tiny, rotate=30] at (axis cs:-100,68.7) {GPT-3.5-Turbo};
    \node[anchor=west, font=\tiny, rotate=30] at (axis cs:2300,88.6) {GPT-4o};
    \node[anchor=east, font=\tiny, rotate=30] at (axis cs:700,87.8) {GPT-4-Turbo};
    \node[anchor=west, font=\tiny, rotate=30] at (axis cs:100,85.5) {GPT-4};
    \node[anchor=west, font=\tiny, rotate=30] at (axis cs:2100,67.3) {Llama 3 8b};
    \node[anchor=east, font=\tiny, rotate=30] at (axis cs:300,49.7) {Llama 2 13b};
    \node[anchor=east, font=\tiny, rotate=30] at (axis cs:300,36.3) {Llama 2 70b};
    \node[anchor=south, font=\tiny, rotate=30] at (axis cs:2100,85.6) {Llama 3 70b};
    \node[anchor=west, font=\tiny, rotate=30] at (axis cs:3500,68.0) {Llama 3.1 8b};
    \node[anchor=west, font=\tiny, rotate=30] at (axis cs:3500,89.5) {Llama 3.1 70b};
    \node[anchor=west, font=\tiny, rotate=30] at (axis cs:3500,91.2) {Llama 3.1 405b};
    \node[anchor=west, font=\tiny, rotate=30] at (axis cs:500,59.4) {Mistral 7b};
    \node[anchor=east, font=\tiny, rotate=30] at (axis cs:3300,64.3) {Mistral NeMo};
    \node[anchor=east, font=\tiny, rotate=30] at (axis cs:900,73.2) {Mixtral 8x7b};
    \node[anchor=east, font=\tiny, rotate=30] at (axis cs:900,77.7) {Mixtral 8x22b};
    \node[anchor=east, font=\tiny, rotate=30] at (axis cs:1700,77.7) {Mistral Medium};
    \node[anchor=south, font=\tiny, rotate=30] at (axis cs:1700,78.9) {Mistral Large};
    \node[anchor=east, font=\tiny, rotate=30] at (axis cs:3700,80.9) {Mistral Large 2};
    \node[anchor=east, font=\tiny, rotate=30] at (axis cs:1100,47.7) {Phi-2};
    \node[anchor=north, font=\tiny, rotate=30] at (axis cs:3100,75.3) {Phi-3-mini};
    \node[anchor=south, font=\tiny, rotate=30] at (axis cs:3100,77.3) {Phi-3-small};
    \node[anchor=south, font=\tiny, rotate=30] at (axis cs:3100,82.2) {Phi-3-medium};
    \node[anchor=north, font=\tiny, rotate=30] at (axis cs:2500,85.4) {Qwen 2 72b};
    \end{axis}
\end{tikzpicture}
\caption{Results of all models on Task 3, sorted by model release date}
\label{fig:task_3_resutls_by_model_release_date}
\end{figure}
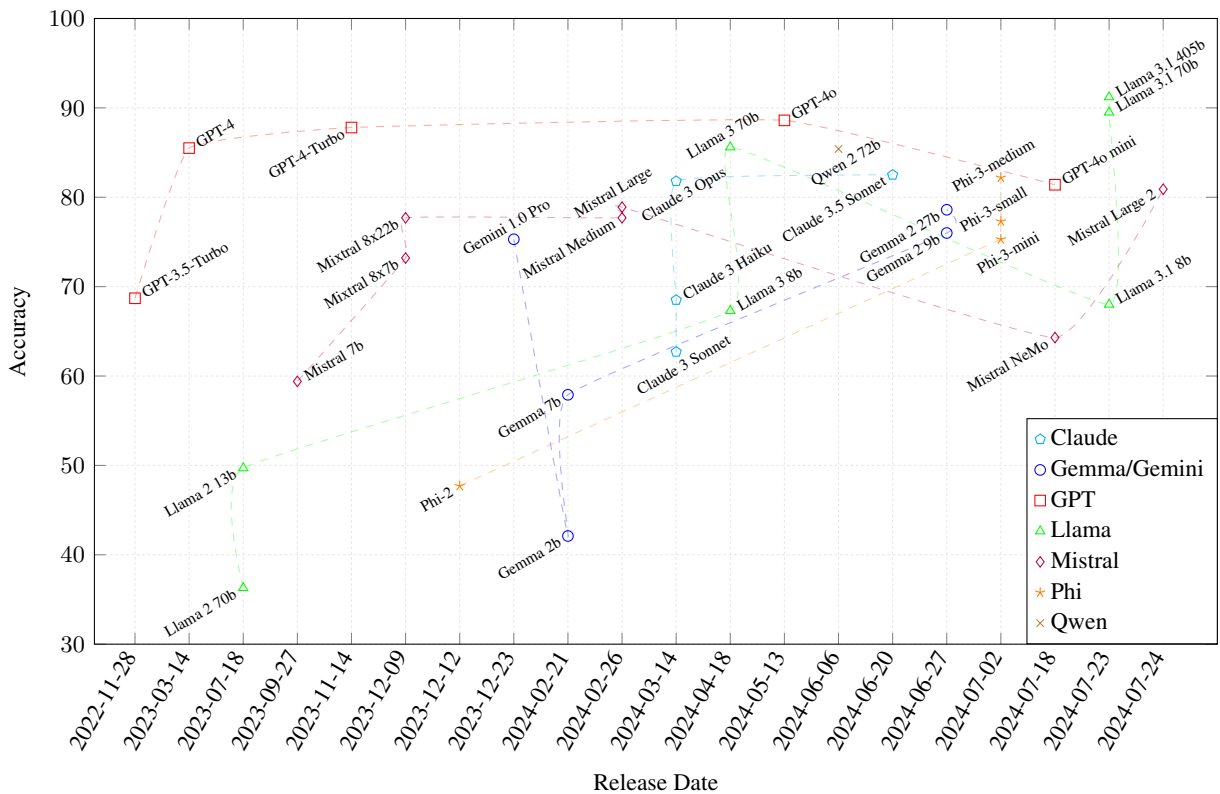

Figure {\ref{fig:task_3_resutls_by_model_release_date}} further confirms this finding by illustrating the relationship between model accuracy and release date. The graph demonstrates a general upward trend in accuracy for models released over time, highlighting the ongoing advancements in model architecture, training data, and methodologies.

Notably, models like GPT-4 and GPT-4 Turbo, released in 2023, continue to perform exceptionally well despite their relative age. This success can be attributed to their substantial model sizes and the extensive datasets used for training. Interestingly, GPT-4o, although smaller than both GPT-4 and GPT-4 Turbo, outperforms them, likely due to enhancements in training techniques and optimization.

A similar trend is observed across other model families. The Llama 3.1 models show a marked improvement over the Llama 3 models, which themselves significantly outperform the older Llama 2 models. This progression underscores the importance of continued refinements in model development. Similarly, Mistral NeMo demonstrates a performance edge over earlier iterations, and Claude 3.5 Sonnet surpasses the previous Claude 3 models, reflecting the positive impact of recent advancements.

The graph also reinforces the observation that more recently released models generally outperform older, larger models. For instance, the Phi-3-medium model outperforms larger models such as Mistral Large and Mixtral 8x22b, suggesting that improvements in training quality and data diversity play a crucial role in model effectiveness.

However, within the same release period, model size remains a critical factor when training data quality is comparable. This is evident in the consistent performance gains seen with larger models, highlighting that, while newer training techniques and data can enhance smaller models, the additional capacity provided by larger models often leads to superior accuracy.

{
\subsection{Fine-tuning Exploration}
\label{subsec:finetuning}
This section presents and discusses the results of our initial fine-tuning explorations. Our primary goal here is to examine the in-domain and out-of-domain performance impact of fine-tuning GPT-4o-mini and Phi-3.5-mini on Disorder Severity Assessment.

To initially probe the benefits of fine-tuning, we conducted targeted experiments with both GPT-4o-mini and Phi-3.5-mini. Focusing on GPT-4o-mini, fine-tuning on a 1600-sample, balanced DEPTWEET subset yielded encouraging in-domain gains, achieving a notable 27.5 percentage point improvement in \gls{BA} on DEPTWEET itself. Evaluating out-of-domain transfer, fine-tuned GPT-4o-mini also demonstrated positive performance shifts on DepSeverity in fine-tuned settings, with a 5.6\% increase in \gls{BA} accompanied by a decrease of 0.028 in \gls{MAE}. Similarly, in-domain fine-tuning on DEPTWEET resulted in a decreased \gls{MAE} of 0.297 (decrease in \gls{MAE} of 0.352). However, this positive trend was not universally observed; on the RED SAM dataset, fine-tuning on DEPTWEET led to a decrease in performance (-3.4\% \gls{BA} and an increase of +0.321 \gls{MAE}). 

A similar mixed trend emerged with Phi-3.5-mini, where we explored parameter-efficient fine-tuning employing LoRA in 4-bit quantization (Q4). Fine-tuning Phi-3.5-mini on DEPTWEET yielded substantial in-domain improvements (+45.5\% \gls{BA} \& -0.595 \gls{MAE} on DEPTWEET) and positive transfer to DepSeverity (+12.7\% \gls{BA} \& -0.215 \gls{MAE}) and Red Sam (+7.7\% \gls{BA} \& -0.082 \gls{MAE}), though a slight degradation was observed on SAD (-6\% \gls{BA} \& +0.25 \gls{MAE}). Interestingly, fine-tuning Phi-3.5-mini on the stress-focused SAD dataset, instead of DEPTWEET, resulted in a different performance profile, with more modest in-domain gains on SAD (+30.2\% \gls{BA} \& -1.766 \gls{MAE}), but also smaller, yet still positive transfer effects on DEPTWEET (+14.4\% \gls{BA} \& -0.184 \gls{MAE}), DepSeverity (+4.8\% \gls{BA} \& -0.088 \gls{MAE}), and RED SAM (+4\% \gls{BA} \& -0.037 \gls{MAE}).

Fine-tuning was performed using the LoRA method implemented in the Hugging Face Transformers library, with an Adam optimiser, a learning rate of 2e-5, a batch size of 16, and a maximum of 3 epochs. We used early stopping if validation performance, measured by balanced accuracy for classification tasks and MAE for severity estimation, did not improve for two consecutive epochs. Validation sets comprised 20\% of the training data, stratified by outcome class or severity score. All experiments were run on NVIDIA A100 GPUs. We tracked both training cost (GPU-hours) and inference latency before and after fine-tuning. 
}

{
\subsection{Limitations}
\label{subsec:limitations_discussion}
This study demonstrates the potential of applying \gls{LLM}s to mental health tasks, but several limitations should be considered when interpreting the findings. First, the reliance on social media data introduces concerns regarding data quality, representativeness, and labeling subjectivity, all of which may bias results. In addition, practical constraints such as computational cost, context-length restrictions, \gls{API} filtering, prompt sensitivity, and the varying degrees of ``stubbornness'' observed across different \gls{LLM}s challenge both the scalability and generalizability of these methods. Beyond technical factors, the inherent subjectivity of psychiatric evaluation, coupled with ethical considerations related to data privacy, algorithmic bias, and model transparency, further complicates deployment in real-world settings. While the results highlight encouraging opportunities, they should be interpreted cautiously, with recognition of these challenges and the need for continued research. A more detailed discussion of these limitations and their implications for generalizability is presented in Section~\ref{sec:limitations}.

A further limitation lies in the absence of comprehensive statistical significance testing when comparing model performance across tasks and prompt conditions. Although prior work has underscored the importance of such testing to ensure robust interpretation---particularly in cases of modest performance differences---implementing it here was infeasible. This limitation stemmed from the large number of models evaluated ($\sim$35), restricted access to several \gls{API}s, and the prohibitive computational and financial costs of repeated experimental runs. Future work should prioritize integrating statistical significance testing frameworks as model access stabilizes and benchmarking resources expand, thereby enabling more rigorous and reliable comparative analysis.

}

{
\section{Implications of the Study}
\label{sec:implications}
This study provides a comprehensive look at how well different \gls{LLM}s can perform on mental health tasks. The findings have important implications for using \gls{AI} in psychiatry. In this section, we discuss the practical and ethical implications of deploying \gls{LLM}s in this sensitive domain. THe implications can be summarized int he following points
\begin{itemize}
\item \textbf{Model Selection:}
There's no "one-size-fits-all" model for mental health tasks. Performance varies significantly across datasets. Developers need to carefully choose models based on their specific needs and the type of data they're working with. However, some models demonstrated more consistent performance across tasks than others. Notably, the OpenAI GPT family consistently ranked among the top performers in binary classification, severity estimation, and knowledge assessment. Similarly, the Mistral family, particularly Mistral NeMo and Mistral Medium, showed strong and consistent performance on certain datasets. As such, developers should test different models on data similar to theirs to find the best fit.

\item\textbf{Closed vs Open-Source Models:} 
While OpenAI's models were the best overall models for many tasks, they remain closed-source, without the possibility of the stricter monitoring and transparency necessary for model deployment in this field. As such, open-source models offer a viable and scalable alternative for many mental health applications, especially considering the sensitive nature of the domain. Models like Llama 3 70b are comparable to closed-source models like GPT-4 in binary disorder detection, while being much smaller. Mistral NeMo, an even smaller model, is great for quick and efficient processing. They can be customized (i.e. via fine-tuning) and deployed more easily, especially when resources are limited. Furthermore, both open-source and closed-source models (where providers allow) can be fine-tuned on specific datasets. This can further improve performance and tailor the models to particular mental health tasks or populations.

\item\textbf{Prompt Engineering is Key:}
Prompt engineering significantly improve model performance, especially for smaller models. The results presented in this study highlight the significant gains that can be achieved through careful prompt design and refinement. Rather than simply selecting a model and deploying it directly, developers should invest effort in systematically crafting and testing various prompts to identify those that result in the most accurate and reliable responses. As demonstrated by the substantial improvement of Gemma 7b, where prompt engineering elevated its accuracy on the Dreaddit dataset to 76.70\% (an improvement of 22.58\%) using the BIN-4 prompt, this process can unlock the full potential of models, even those with limited computational resources. Furthermore, our experiments, as shown in Table \ref{tab:binary_repetition_line_experiment} in Section \ref{sec:additional_investigations}, suggest that incorporating a repetition line—a restatement of the core instruction—within the prompt can also enhance performance, potentially by reinforcing the desired behavior in the model.

\item\textbf{\gls{FS} Learning Improves Accuracy:}
\gls{FS} learning generally improves accuracy. This was evident across nearly all models and datasets tested, such as the significant performance boost observed in Phi-3-mini, achieving an average improvement of +8.18\% in \gls{BA} and a reduction of -0.488 in \gls{MAE} across the datasets tested. While the degree of improvement varied, the trend was clear: providing examples generally led to better results. However, it's important to remember that \gls{FS} learning typically increases computational cost, often using 3-4x more tokens during inference (with three example posts provided, along with the one for which we require evaluation). As such, while \gls{FS} learning presents a powerful technique for enhancing model performance, developers must carefully weigh the benefits of improved accuracy against the increased computational demands, especially when deploying models in resource-constrained environments or when dealing with high-volume, real-time applications.

\item\textbf{LLMs Can Estimate Severity, but with Limits:}
\gls{LLM}s demonstrate a capability to estimate the severity of mental health conditions, but their accuracy is not yet at a level where they can be solely relied upon for critical decisions. While models like GPT-4-Turbo achieved a relatively low MAE of 0.455 on the DEPTWEET dataset, other models and datasets showed higher average errors. For instance, on the DepSeverity dataset, Mistral Medium had an MAE of 0.779, and on the SAD dataset, even the best-performing model, Claude 3 Sonnet, had an MAE of 1.404, which is relatively impressive on a 10 point severity scale. These results suggest that while \gls{LLM}s can provide a general estimate of severity, often within one level of the actual value on a multi-level scale, they should be used in conjunction with, not in place of, human clinical judgment. The potential for error, as indicated by the \gls{MAE}, should always be taken into account when interpreting model outputs.

\item\textbf{Newer Models are Generally Better for Knowledge Tasks: }
Newer \gls{LLM}s, such as the Llama 3.1 family, demonstrate a significantly improved understanding of psychiatric knowledge compared to their predecessors. In the psychiatric knowledge assessment task, Llama 3.1 405b achieved an impressive accuracy of 91.2\% (exceeding GPT-4o), followed by Llama 3.1 70b at 89.50\%. Even the smaller Llama 3.1 8b model outperformed many older, larger models. This suggests that newer models may have been trained on more comprehensive datasets or have architectures better suited for knowledge retention and retrieval. Therefore, for applications requiring an accurate and broad psychiatric knowledge base, such as question-answering, prioritizing the use of the most recent \gls{LLM}s is likely to yield the best results. Developers should consider regularly updating their systems to incorporate advancements in newer model releases. Furthermore, techniques like \gls{RAG}, while not explored in this study, could potentially further enhance performance by providing models with access to external, up-to-date knowledge sources.

\item\textbf{Trade-offs Between Safety and Accuracy:}
Models with stricter ethical guidelines, like Llama 3.1 and Gemini, sometimes performed worse on mental health tasks. While these safeguards are intended to prevent harm and ensure responsible use, they can also limit the ability of \gls{LLM}s to process and respond to sensitive information that may be crucial for accurate assessments. For example, the Llama 3.1 models exhibited a higher rate of invalid responses, potentially due to their heightened sensitivity to prompts related to mental health issues. As such, developers and researchers need to find a balance between making \gls{AI} safe and making it useful for mental health. Overly strict rules might limit the ability of \gls{LLM}s to help with sensitive topics, while lax rules could lead to unintended negative consequences.

\item\textbf{Bias is a Concern:}
While our experiments did not touch on this point, \gls{LLM}s can inherit biases from their training data, potentially leading to unfair or inaccurate assessments, particularly for underrepresented groups. Although our results cannot directly determine whether a model's incorrect answer stems from bias or other factors, developers must be aware of this risk and strive to mitigate potential biases.

\item\textbf{Explainability is Important:}
It is often difficult to understand how \gls{LLM}s arrive at their answers. This lack of transparency can be problematic in mental healthcare, where clinicians and patients need to understand the reasoning behind \gls{AI}-generated assessments or recommendations. While true explainability remains a challenge, research into techniques like \gls{CoT} prompting and attention visualization may offer insights into model decision-making.

\item\textbf{Data Privacy Must be Protected:}
\gls{LLM}s for mental health often use sensitive personal information. Thus, protecting data privacy is essential. Strict security measures, data anonymization techniques, and robust ethical guidelines are needed when collecting, storing, and using data for \gls{LLM} training and deployment in mental healthcare.
\end{itemize}

In summary, this study demonstrates that while \gls{LLM}s offer exciting possibilities for improving mental healthcare, realizing their full potential requires a balanced approach, careful model selection, thorough prompt engineering, and attention to ethical boundaries.

\section{Additional Investigations}
\label{sec:additional_investigations}
In this section, we discuss side experiments that we performed while on the course of this larger analysis that are interesting but not core to this study.

\subsection{Repetition Line Experiment}
\label{subsec:repetition_line_experiment}
In this experiment, we evaluate the effect of the repetition line presented in the SEV-4 prompt (Figure \ref{fig:severity_prompts}). The effect of the line is observed in a binary task in isolation of any other modifications. The line is appended to the end of the binary BIN-1 prompt as showcased in Table \ref{tab:binary_1.1_prompt}. 

\begin{table}[h!]
\caption{Prompt BIN-1.1}
\label{tab:binary_1.1_prompt}
\centering
\begin{tabular}{@{}l@{}} 
\toprule
{\textbf{Prompt BIN-1.1}} \\ 
\midrule
As a psychologist, analyze the provided social media post(s). \\ 
Is the poster depressed? \\
(Answer with Yes or No only without explaining your reasoning) \\
\textbf{(Please return only the label (Yes or No) and no other text)} \\
\bottomrule
\end{tabular}
\end{table}

The results in Table \ref{tab:binary_repetition_line_experiment} reveal a consistent trend of marked improvements in nearly all models with the improvements ranging from 1\% to an {utmost} of 14\%. Interestingly, the models that had the largest improvements (>10\%) are all Mistral-based, which suggests that this particular family of models, whether for training reasons or how they were initially trained, seem to prefer reiterating the instructions or the important information in a prompt. Other model families generally still had respectable improvements of around 1-2\%, which suggests that even other models prefer reiterating certain parts of the prompt, with only Claude 3 Sonnet and GPT-3.5-Turbo having a decrease in performance. 

\begin{table}[h!]
\caption{Effect of Repetition Line in Binary Prompt BIN-1 on DepSeverity (Binary - 1000 Samples)}
\label{tab:binary_repetition_line_experiment}
\centering
\begin{tabular}{@{}lrr@{}}
\toprule
\textbf{Model} & \textbf{BIN-1} & \textbf{BIN-1.1} \\ \midrule
Claude 3 Haiku  &69.20\%         &  \textcolor{BrickRed}{69.10\%  (-0.10\%)} \\
Claude 3 Sonnet & 69.10\%         & \textcolor{BrickRed}{66.40\% (-2.70\%)}                  \\
Gemma 7b        & 54.10\%         & \textcolor{OliveGreen}{55.00\% (+0.90\%)}                  \\
Gemma 2 9b      & 68.50\%        & \textcolor{BrickRed}{67.60\% (-0.90\%)} \\
Gemma 2 27b     & 66.00\%        & \textcolor{OliveGreen}{68.50\% (+2.50\%)} \\
Gemini 1.0 Pro       & 69.00\%         & \textcolor{OliveGreen}{71.40\% (+2.40\%)}                  \\
GPT-3.5-Turbo         & 71.40\%         & \textcolor{BrickRed}{67.00\% (-4.40\%)}                  \\
Llama 3 8b      & 61.48\%        & \textcolor{OliveGreen}{66.80\% (+5.32\%)}                 \\
Llama 3.1 8b    & 63.30\%        & \textcolor{OliveGreen}{66.20\% (+2.9\%)} \\
Llama 2 70b     & 72.70\%         & \textcolor{OliveGreen}{75.00\% (+2.30\%)}                  \\
Llama 3 70b     & 70.49\%        & \textcolor{OliveGreen}{71.90\% (+1.41\%)}                 \\
Llama 3.1 70b    & 69.80\%        & \textcolor{OliveGreen}{69.90\% (+0.1\%)} \\
Llama 3.1 405b    & 67.10\%        & \textcolor{OliveGreen}{68.70\% (+1.6\%)} \\
Mistral 7b      & 52.98\%        & \textcolor{OliveGreen}{64.40\% (+11.42\%)}                \\
Mistral NeMo    & 64.30\%        &  \textcolor{BrickRed}{61.10\% (-3.20\%)} \\
Mixtral 8x7b    & 58.75\%        & \textcolor{OliveGreen}{69.30\% (+10.55\%)}                \\
Mixtral 8x22b   & 57.81\%        & \textcolor{OliveGreen}{71.90\% (+14.09\%)}                \\
Mistral Medium  & 59.70\%         & \textcolor{OliveGreen}{71.10\% (+11.40\%)}                 \\
Mistral Large 2 & 54.20\%          & \textcolor{OliveGreen}{72.50\% (+18.30\%)} \\
Phi-3-mini  & 70.50\%         & \textcolor{BrickRed}{70.10\% (-0.40\%)} \\
Phi-3-small  & 63.40\%         & \textcolor{OliveGreen}{70.20\% (+6.80\%)} \\
Phi-3-medium  & 71.60\%         & \textcolor{OliveGreen}{72.50\% (+0.90\%)} \\
Qwen 2 72b  & 72.80\%         & \textcolor{OliveGreen}{75.10\% (+2.3\%)} \\
\bottomrule
\end{tabular}
\end{table}

\subsection{Knowledge Reinforcement Experiment}
In this experiment, we sought to determine whether providing models with information they had previously generated would enhance their performance. This idea stemmed from an earlier study (not included in this paper) where we provided models with the \gls{DSM-V} criteria for depression. Interestingly, some models improved in accuracy, while others experienced a decline. To investigate this phenomenon more systematically, we designed the Knowledge Reinforcement Experiment.

The experiment was conducted in two steps. First, each \gls{LLM} was prompted to list symptoms it associates with depression. In this step, we used "Give me a detailed list of symptoms that indicate whether a person is depressed or not." as a prompt.

In the second step, the list of symptoms provided by each model was appended to a modified version of the BIN-1 prompt, as shown in Table \ref{tab:zs_reiteartion_prompt_2}. This modified prompt, which we call the Knowledge Reinforcement Prompt, included a repetition line designed to help the models adhere more closely to the instructions, particularly since the detailed list of symptoms significantly expanded the prompt. The longer prompts occasionally caused models (notably the Mistral Large and Claude 3 Opus) to disregard some instructions, such as providing a simple "Yes" or "No" answer without explanation.

\begin{table}[h!]
\caption{Prompt BIN-1 + Disorder symptoms extracted from the same \gls{LLM} that is being prompted}
\label{tab:zs_reiteartion_prompt_2}
\centering
\begin{tabular}{@{}l@{}} 
\toprule
\textbf{Knowledge Reinforcement Prompt} \\ 
\midrule
As a psychologist, analyze the provided social media post(s). \\
Is the poster depressed? \\
(Answer with Yes or No only without explaining your reasoning) \\
(Please return only the label (Yes or No) and no other text) \\
List of depression symptoms: \\
... \\
\bottomrule
\end{tabular}
\end{table}

Table \ref{tab:zs_reiteartion_results} presents the results of this experiment. The main goal is to observe whether a model would benefit from being provided information it very clearly knows, given that it provided us that information to begin with. The results lack a clear pattern, with some models showing a moderate increase in accuracy, notably GPT-3.5-Turbo and Claude 3 Sonnet. However, most smaller \gls{LLM}s experienced a decrease in accuracy, some quite significantly. We are unsure why smaller \gls{LLM}s were negatively affected, as the additional information should have been neutral or even beneficial. One hypothesis is that the lengthy prompts with extensive information might have distracted them from the main instructions and task, potentially leading to poorer performance.

\begin{table}[h!]
\caption{Results of the Knowledge Reinforcement experiment compared to the baseline {BIN-1.1} prompt on DepSeverity (Binary - 1000 Samples)} 
\label{tab:zs_reiteartion_results}
\centering
\begin{tabular}{@{}lrr@{}}
\toprule
\textbf{Model} & \textbf{Prompt BIN-1.1 (baseline)} & \textbf{Knowledge Reinforcement Prompt}  \\ \midrule
Claude 3 Haiku & 69.1\% & \textcolor{BrickRed}{-9.7\%} \\
Claude 3 Sonnet & 66.4\% & \textcolor{OliveGreen}{+4.2\%} \\
Gemma 7b & 55\% & 0\% \\
Gemini 1.0 Pro & 71.4\% & \textcolor{OliveGreen}{+0.2\%} \\
GPT-3.5-Turbo & 67\% & \textcolor{OliveGreen}{+2.8\%} \\
Llama 3 8b & 66.8\% & \textcolor{BrickRed}{-8.5\%} \\
Llama 2 70b & 75\% & \textcolor{BrickRed}{-4.1\%} \\
Llama 3 70b & 71.9\% & \textcolor{OliveGreen}{+0.9\%} \\
Mistral 7b & 64.4\% & \textcolor{BrickRed}{-0.6\%} \\ 
Mixtral 8x7b & 69.3\% & \textcolor{BrickRed}{-4.3\%} \\
Mixtral 8x22b & 71.9\% & \textcolor{BrickRed}{-3.2\%} \\
Mistral Medium & 71.1\% & \textcolor{OliveGreen}{+1.3\%} \\ \bottomrule
\end{tabular}
\end{table}

{
\subsection{Clinical Interpretation of Model Performance}
While our evaluation primarily reports quantitative performance metrics, the patterns observed also yield clinically relevant insights into why certain models outperform others. These factors extend beyond raw accuracy and should inform model selection for psychiatric applications.
Models such as GPT-4o and GPT-4 consistently demonstrated both higher adherence to task instructions and superior accuracy across multiple datasets, particularly in structured classification tasks (e.g., 85.0\% accuracy on DEPTWEET using BIN-4). This suggests that strong instruction-following finetuning an approach that explicitly optimizes for compliance with structured prompts may align more closely with the rigid formats used in psychiatric assessments and screening tools. In a clinical workflow, this could translate to more reliable completion of standardized diagnostic questionnaires or risk assessments without unintended deviations.
Our results challenge the assumption that larger models automatically yield better clinical performance. For example, the Phi-3-mini (3.8B) model outperformed its larger sibling, Phi-3-small (7B), on the RED SAM dataset (57.8\% vs. 57.0\%), and Gemma 2 9B outperformed Gemma 2 27B on 5 out of 6 datasets, including an advantage on DEPTWEET (80.4\% vs. 77.9\%). These findings highlight that architecture, optimization, and fine-tuning objectives can outweigh parameter count, especially when tasks require precise adherence to specialized prompts rather than broad world knowledge.
Although specific training datasets for proprietary models are undisclosed, performance patterns suggest that exposure to medical or scientific text either explicitly or through general web-scale sources may enhance psychiatric task performance. Models with likely greater medical content coverage, such as GPT-4o and Qwen 2 72B, showed consistently strong results in psychiatric knowledge assessments (e.g., 91.2\% on MedMCQA for Llama 3.1 405B, which is likely influenced by targeted domain adaptation). This reinforces the value of domain-specific fine-tuning or continual pretraining when deploying LLMs for clinical reasoning tasks.
We observed that certain models, such as Llama 3.1 and Gemini, refused to process a notable portion of psychiatric task prompts, likely due to strengthened safety alignment. While such refusals limit benchmarking completeness, they may be desirable in real-world triage contexts especially for high-risk content like active suicidality where containment and safe handoff to a human clinician are preferable to automated classification. Clinically, the optimal balance between responsiveness and safety will depend on deployment context: high-throughput population screening may tolerate fewer refusals, whereas acute care settings may benefit from stricter safeguards.
In summary, model selection for psychiatric applications should weigh not only accuracy but also instruction-following ability, efficiency relative to size, potential domain adaptation, and the intended trade-off between safety and coverage.

Given the clinical importance and complexity of severity estimation, we conducted a targeted error analysis to better understand where models failed, focusing on symptom subtlety, contextual sensitivity, and diagnostic bias. Across models, the most common failure mode was under-recognition of mild or subthreshold symptom presentations; for example, depressive symptoms conveyed indirectly (“It’s been harder to get out of bed lately, but I’m fine”) were often misclassified as minimal severity even by high-performing models such as GPT-4o and Claude 3.5, highlighting the challenge of lexical subtlety and indirect phrasing. Contextual cues such as sarcasm, humor, or cultural idioms also led to misclassifications; in one illustrative case (“Guess I’m winning the ‘most anxious’ award this year”), models like Phi-3-mini interpreted the statement literally as mild anxiety rather than recognizing hyperbolic humor indicating moderate to severe symptoms, whereas contextually stronger models like GPT-4o and Qwen 2 72B were less affected but still occasionally erred. We also observed overpathologizing tendencies, particularly among smaller or less instruction-tuned architectures, where casual mentions of stress (“This week has been rough at work”) were classified as moderate depression or anxiety, potentially leading to unnecessary escalation. These patterns carry distinct clinical risks: missed mild symptoms may delay intervention, while overpathologizing can inflate false positives and burden services. Mitigating these errors may require context-aware pre-screening layers or hybrid human--AI review, especially for borderline cases.

\begin{table}[h]
\centering
\caption{Error breakdown in disorder severity estimation. Frequencies represent the proportion of misclassified cases in the error analysis subset and sum to 100\%.}
\begin{tabular}{p{2.5cm}p{4.5cm}p{3cm}p{2cm}p{3.5cm}}
\hline
\textbf{Error Type} & \textbf{Example Case} & \textbf{Commonly Affected Models} & \textbf{Frequency (\%)} & \textbf{Primary Risk in Deployment} \\
\hline
Missed Mild Symptoms & ``It’s been harder to get out of bed lately, but I’m fine.'' & GPT-4o, Claude 3.5, Phi-3-mini & 42 & Under-intervention; delayed detection of early symptoms \\ \hline
Context Misinterpretation & ``Guess I’m winning the ‘most anxious’ award this year.'' & Phi-3-mini, Gemma 2 9B, Mistral NeMo & 31 & Misclassification; inaccurate triage in subtle cases \\ \hline
Overpathologizing & ``This week has been rough at work.'' & Phi-3-mini, Gemma 7B, smaller Llama 3 & 27 & False positives; unnecessary escalation and resource strain \\
\hline
\end{tabular}
\end{table}

\subsection{Nuanced Interpretation of Model Refusals}
While refusal behavior in some models (e.g., Llama 3.1 family, Gemini) reduced the number of analyzable responses in our benchmarking, such behavior should not be interpreted solely as a limitation. In real-world psychiatric triage, refusal to answer may play a protective role preventing the generation of potentially harmful, misleading, or unsafe content.

We distinguish between undesirable refusals and desirable refusals:

\begin{itemize}
    \item Undesirable refusals occur when models decline to respond to benign evaluative prompts, such as classification of anonymized social media posts for research purposes. These refusals can hinder technical evaluation and limit model utility in controlled research settings.
    \item Desirable refusals occur when the model detects that a task may produce unsafe or harmful outputs. For example, during an acute suicide risk scenario, a refusal to provide direct advice could allow the model to instead suggest immediate contact with emergency services or transfer to a human clinician. This aligns with clinical best practices, where safety and containment take precedence over diagnostic completeness.
\end{itemize}
From a deployment perspective, the optimal balance between responsiveness and safety will depend on context. Population-level screening tools may benefit from minimal refusals to ensure coverage, whereas crisis intervention systems may prioritize refusal in situations where any automated response could cause harm.
}

{
\section{Limitations and Future Directions}

Although this study offers valuable information on the capabilities of \gls{LLM}s for mental health tasks, it is important to recognize several limitations that may affect the interpretation and generalizability of our findings. In addition to the future directions that can help achieve higher performance.

\subsection{Limitations}
\label{sec:limitations}

These limitations arise from the inherent characteristics of the datasets we used, the constraints of the \gls{LLM}s themselves, and the broader challenges associated with evaluating mental health through social media.

\subsubsection{Dataset-Related Limitations}

\paragraph{Data Source and Quality:}
In this study, we relied on publicly available social media datasets to train and evaluate the \gls{LLM}s. While these datasets offer a valuable resource for exploring mental health language, they lack the rigor of clinical reports (e.g., standardized diagnostic criteria). Though we prioritized datasets with human annotations \cite{turcan2019dreaddit, mauriello2021sad, kabir2023deptweet}, label subjectivity remains a concern – particularly for nuanced conditions like depression, where annotator agreement varies significantly \cite{kabir2023deptweet}. Additionally, social media demographics over-represent younger, tech-literate populations, limiting generalizability to clinical populations. Our exclusive focus on English-language datasets further restricts cross-cultural applicability. 

\paragraph{Limited Number of Disorders:}
Our study focused on a limited set of mental health conditions, specifically stress, depression, and suicidal ideation, as detailed in Section \ref{subsec:datasets}. While these conditions are prevalent and important, they do not represent the full spectrum of mental health disorders. This limitation is primarily due to the availability of suitable datasets; datasets focusing on other mental health conditions that met our quality criteria were scarce. Consequently, our findings may not generalize to other mental health conditions, and further research is needed to evaluate the performance of \gls{LLM}s across a broader range of disorders.

\subsubsection{Model-Related Limitations}
\label{subsec:model_related_limitations}
\paragraph{Context Length Restrictions:}
Several models we evaluated, such as Phi-2, have limitations in the length of text they can process at once, constrained by their fixed context window sizes. While this was not a major limitation in our study as most datasets we used did not have particularly long text, it did prevent us from evaluating the CSSRS dataset. This limitation may pose challenges for future research involving longer social media posts or extensive patient histories.

\paragraph{API Filtering and Ethical Safeguards:}
API filtering (e.g., Gemini 1.0 Pro blocking self-harm prompts) and built-in safeguards (e.g., Llama 3.1’s refusal rates) limited our ability to probe critical mental health scenarios. While ethically justified (to prevent misuse), these restrictions – exemplified by Llama 3.1’s 22\% invalid response rate (Section \ref{subsec:invalid_response_analysis}) – create evaluation blind spots that may underestimate model risks in real-world deployment. This is the primary reason for excluding Gemini 1.5 for this study.

\paragraph{Model Stubbornness:}
In our experiments, Some models (notably Mistral \& Claude variants) frequently ignored formatting instructions (e.g., "answer concisely"), requiring iterative prompt adjustments (Section \ref{sebsec:severity_prompts}). This stubbornness complicated fair comparisons, as task performance metrics for these models may reflect compliance issues rather than true capability gaps. 

\subsubsection{Methodological Limitations}

\paragraph{Focus on Specific Tasks:}
Our study focused on a limited set of tasks: binary disorder detection, severity evaluation, and psychiatric knowledge assessment. While these tasks are important, they do not encompass the full spectrum of mental health applications. Our findings, therefore, may not be generalizable to other tasks, such as assessing empathy and engaging in therapeutic dialogue. Further research is needed to evaluate \gls{LLM}s on a wider range of tasks to fully understand their capabilities and limitations in the context of mental healthcare.

\paragraph{Exploration of Prompt Space:}
While we extensively explored various prompt engineering techniques, including role-playing and structured prompts, the vast space of possible prompts makes it impossible to guarantee that we identified the optimal ones for each model and task. Our results revealed that \gls{LLM} performance is highly sensitive to prompt wording and structure, with minor changes leading to significant result variations.

\paragraph{Computational Cost and Scalability:}
Using large \gls{LLM}s, such as GPT-4 and Claude 3 Opus, presented a high computational cost constraint in our study. To manage expenses, we employed sampling techniques, as discussed in Section \ref{subsec:sampling}, and relied on Hoeffding's inequality (Equation \ref{eqn:1}) to determine appropriate sample sizes. While this allowed for a diverse range of models to be included, it limited the amount of data processed for each, meaning our findings are based on representative samples rather than complete datasets.

\paragraph{Lack of Transparency and Explainability:}
The "black box" nature of \gls{LLM}s limited our ability to investigate the reasoning behind model outputs. While this is a known methodological constraint of current architectures, it raises broader ethical concerns about deploying such systems in mental health contexts. Future research should focus on explainable \gls{AI} techniques that can provide more transparent and interpretable results in the context of mental health evaluations. 
}

\subsection{Future Directions}
\label{sec:future}
Our analysis identifies six key priorities for advancing \gls{LLM} applications in mental health summarized in the following points:

\begin{itemize}
    \item First, improving model transparency is essential given the "black box" nature of current architectures. Implementing \gls{CoT} prompting could mitigate this by enforcing explicit reasoning chains, given our finding that simple prompt changes caused 22.6\% accuracy swings. This would help clinicians verify if models correctly weigh symptoms versus contextual factors in posts from platforms like Reddit.

\item Second, since smaller models (12B-14B parameters) matched or outperformed larger counterparts, future work should focus on integrating \gls{RAG} with clinical foundations. Linking models like Phi-3 and Mistral NeMo to diagnostic criteria from \gls{DSM-V} or anonymized hospital case notes (in closed clinical settings) would ground predictions in medical reality while enabling verifiable citations. This approach avoids costly parameter scaling while improving the clinical relevance.

\item Third, domain-specific fine-tuning using clinically vetted data is essential for accurate assessments. While our general-purpose models achieved up to 85\% detection accuracy on certain datasets, their performance varied across datasets and disorders. Training open-source architectures on therapist-annotated case histories (with rigorous de-identification) could reduce these inconsistencies.

\item Fourth, expanding testing to non-English contexts across diverse demographics is crucial for clinical utility. While our work focused on English social media datasets, validating models on localized text (e.g., Arabic tweets or Korean forum posts) would assess cross-cultural robustness. Collaborations with multilingual healthcare providers are needed to create ethically sourced benchmarks without compromising patient anonymity.

\item Fifth, future research should assess multimodal \gls{LLM}s and reasoning models such as o1, o3 and Deepseek-R1. Evaluating audio features (speech rhythm in depression) and video cues (facial affect recognition) alongside textual analysis may better approximate clinical evaluations.

Finally, the field must expand beyond detection tasks to evaluate therapeutic uses. Systematic testing of \gls{LLM} capabilities in emotional support, empathy scoring, crisis de-escalation (e.g., suicide hotline simulations), and cognitive behavioural {therapy} style dialogue guidance will determine clinical readiness.
\end{itemize}
}

{
\section{Conclusion}
\label{sec:conclusion}
Our systematic evaluation of 33 \gls{LLM}s across three clinically-relevant tasks—binary detection, severity estimation, and psychiatric knowledge assessment—reveals several key insights for mental health applications.

\textbf{First}, modern \gls{LLM}s demonstrate strong binary classification capabilities, with GPT-4 achieving 85.2\% accuracy on the SAD dataset and Mistral NeMo (12B parameters) outperforming larger architectures including GPT-4 (74.5\% vs 72\% on Dreaddit). However, reliability varies substantially across datasets and clinical conditions, with no single model family achieving universal superiority. Notably, the OpenAI ecosystem—particularly GPT-4—maintained the highest consistency, ranking as the top or near-top performer across all evaluation tasks.

\textbf{Second}, parameter count proves unreliable as a performance predictor. The 14B-parameter Phi-3-medium matched 80\% of GPT-4's cross-task capabilities, and Mistral NeMo dominated on the RED SAM dataset. Additionally, the 405B Llama 3.1 underperformed its older Llama 3 70B counterpart — a disparity we trace to excessive content filtering (22\% valid query rejection rate) in the newer variant.

\textbf{Third}, methodological choices dramatically impact outcomes. Prompt engineering induced 22.6\% accuracy variance in Gemma 7B, while three-example few-shot learning reduced Phi-3-mini's severity estimation errors by 20\%. These findings caution against comparing models without strict protocol standardization.

\textbf{Fourth}, in knowledge retrieval, more recently released models consistently outperformed predecessors, regardless of size, possibly due to architectural improvements or higher quality training data provided. The Llama 3.1 series dominated this category (405B: 91.2\%, 70B: 89.5\%), while Claude 3.5 Sonnet (82.5\%) surpassed the older and larger Claude 3 Opus (81.8\%). Even the modestly sized 14B Phi-3-medium achieved competitive performance (82.2\%).

Three practical considerations emerge for implementing \gls{LLM}s in mental healthcare: (1) Model architecture matters more than size, as shown by smaller models (e.g., Mistral NeMo) competing with much larger models – developers should prioritize task-specific evaluation over defaulting to the largest available models. (2) Standardized methods are crucial for reliable results, given that performance fluctuated significantly with different prompting techniques – the field needs consistent testing protocols to ensure valid model comparisons. (3) Ethical and practical needs should drive model choices – while closed-source systems like GPT-4 deliver strong benchmarks, open-source alternatives (e.g., Llama 3) offer needed transparency/customization despite balancing acts between safety controls and clinical effectiveness.

It is important to acknowledge the limitations of our study. While we focused on prevalent mental health conditions, our datasets, derived from social media, lack clinical validation and may reflect demographic biases. Evaluation constraints, such as \gls{API} filtering and ethical guards (e.g., Llama 3.1’s query rejections) and model stubbornness (e.g., formatting issues with Claude and Mistral), could obscure real-world performance variations. Our English-only focus and computational sampling approach, driven by cost constraints, restrict the applicability of findings to diverse cultural contexts. Notably, the "black box" nature of \gls{LLM}s complicates clinical interpretability despite their strong benchmarks.
}

\bibliographystyle{unsrt}  
\bibliography{references}

\end{document}

%% file: Abbreviations.tex
\newacronym{AI}{AI}{Artificial Intelligence}

\newacronym{LLM}{LLM}{Large Language Model}

\newacronym{LLMs}{LLMs}{Large Language Models}

\newacronym{MCQ}{MCQ}{Multiple Choice Question}

\newacronym{API}{API}{Application Programming Interface}

\newacronym{BA}{BA}{Balanced Accuracy}

\newacronym{MAE}{MAE}{Mean Absolute Error}

\newacronym{ML}{ML}{Machine Learning}

\newacronym{NLP}{NLP}{Natural Language Processing}

\newacronym{BERT}{BERT}{Bidirectional Encoder Representations from Transformers}

\newacronym{DSM-V}{DSM-V}{Diagnostic and Statistical Manual of Mental Disorders-V}

\newacronym{CoT}{CoT}{chain-of-thought}

\newacronym{ZS}{ZS}{zero-shot}

\newacronym{FS}{FS}{few-shot}

\newacronym{SoTA}{SoTA}{state-of-the-art}

\newacronym{BDI}{BDI}{Beck Depression Inventory}

\newacronym{fMRI}{fMRI}{functional magnetic resonance imaging}

\newacronym{EEG}{EEG}{electroencephalogram}

\newacronym{ECG}{ECG}{electrocardiogram}

\newacronym{PHQ}{PHQ}{Patient Health Questionnaire}

\newacronym{PTSD}{PTSD}{Post-traumatic stress disorder}

\newacronym{EHR}{EHR}{Electronic Health Records}

\newacronym{RAG}{RAG}{Retrieval-Augmented Generation}

%% file: Summary_table.tex
\small
\begin{landscape}
\begin{longtable}{@{}p{3cm}p{3cm}p{3cm}p{3cm}p{3cm}p{2cm}p{3cm}@{}} 
\caption{Summary of Human Annotated Mental Health Datasets} \label{tab:datasets} \\

\toprule 
\textbf{Name} &
\textbf{Task} &
\textbf{Disorder(s)} &
\textbf{Source} &
\textbf{\makecell[l]{Annotator \\ Experience}} &
\textbf{Sample Size} &
\textbf{Dataset Access} \\ 
\midrule
\endfirsthead

\multicolumn{7}{c}%
{{\tablename\ \thetable{} -- continued from previous page}} \\
\midrule
\textbf{Name} &
\textbf{Task} &
\textbf{Disorder(s)} &
\textbf{Source} &
\textbf{\makecell[l]{Annotator \\ Experience}} &
\textbf{Sample Size} &
\textbf{Dataset Access} \\ 
\midrule
\endhead

\hline \multicolumn{7}{r}{{Continued on next page}} \\ \endfoot

\hline
\endlastfoot
Dreaddit \cite{turcan2019dreaddit} &
\makecell[l]{Binary Stress \\ Classification} &
Stress &
Reddit &
Crowdsourced &
1000 posts &
Publicly Available \\

DepSeverity \cite{naseem2022early} &
\makecell[l]{Depression Severity \\ Classification} &
Depression (4 levels)	 &
Reddit &
Professionals &
3,553 posts	 &
Publicly Available \\
\makecell[l]{Stress-Annotated-\\Dataset (SAD)} \cite{mauriello2021sad} &
\makecell[l]{Stress Severity \\ Classification} &
\makecell[l]{Daily Stressors \\ (9 categories)} &
\makecell[l]{SMS-like conversations, \\ crowdsourced data, \\ LiveJournal} &
Crowdsourced  &
\makecell[l]{6850 \\ sentences} &
Publicly Available \\

DEPTWEET \cite{kabir2023deptweet} &
\makecell[l]{Depression Severity \\ Classification} &
Depression (4 levels) &
X (formerly Twitter) &
Crowdworkers (trained) &
40,191 tweets &
\makecell[l]{Available Upon \\ Request} \\

CSSRS-Suicide \cite{gaur2019reddit} &
Suicide Risk Severity Assessment &
\makecell[l]{Suicide Risk \\ (5 levels)} &
Reddit (r/SuicideWatch) &
\makecell[l]{Professional \\ psychiatrists} &
\makecell[l]{500 \\ Reddit users} &
Publicly Available \\

SDCNL \cite{haque2021deep} &
\makecell[l]{Suicide vs. Depression \\ Classification} &
\makecell[l]{Depression, \\ Suicidal Ideation} &
Reddit (r/SuicideWatch, r/Depression) &
N/A &
1,895 posts &
Publicly Available \\
PsyQA \cite{sun2021psyqa} &
\makecell[l]{Generating Long \\ Counseling Text} &
\makecell[l]{Various Mental \\ Health Disorders} &
\makecell[l]{Mental Health \\ Service Platform \\ (Yixinli)} &
\makecell[l]{Trained volunteers or \\ professional counselors} &
\makecell[l]{22K questions, \\ 56K answers} &
\makecell[l]{Available Upon \\ Request} \\
\makecell[l]{MedMCQA \\ (Psychiatry Subset)} \cite{pal2022medmcqa} &
\makecell[l]{Medical Question \\ Answering} &
\makecell[l]{Various psychiatric \\ disorders} &
\makecell[l]{Medical entrance \\ exams \\ (AIIMS \& NEET PG)} &
\makecell[l]{Questions created by \\ exam boards} &
\makecell[l]{4442 (\textasciitilde 5\%)} &
Publicly Available \\
PsySym \cite{zhang2022symptom} & 
  \makecell[l]{Symptom \\ Identification} &
  \makecell[l]{7 mental disorders \\ (Depression, Anxiety, \\ ADHD, Disorder, \\ OCD , Eating Disorder)} &
  Reddit &
  \makecell[l]{Trained volunteers \\ and  psychiatrists} &
  \makecell[l]{8,554 \\ sentences} &
  \makecell[l]{Available Upon \\ Request} \\
\makecell[l]{MedQA \\ (Psychiatry Subset)} \cite{jin2021disease} &
  Medical Question Answering &
  \makecell[l]{Various clinical \\ topics} &
  \makecell[l]{Medical entrance \\ exams \\ (AIIMS \& NEET PG)} &
  Exam boards &
  \makecell[l]{47,457 Total, \\ \textasciitilde 800 psychiatry \\ related} &
  Publicly Available \\
LGBTeen \cite{lissak2024colorful} & 
  \makecell[l]{Emotional Support \\ and Information} &
  Queer-related topics &
  Reddit (r/LGBTeens) &
  \makecell[l]{Human Redditors \\ and \gls{LLM}s} &
  \makecell[l]{1,000 posts, \\ 11,320 \gls{LLM} \\ responses} &
  Publicly Available \footnote{While the author states that this dataset is available on {\color{red}GitHub}, the dataset is not present in the {\color{red}GitHub} repository provided by the author, and the author did not respond to our queries regarding the availability of the dataset} \\
ESConv \cite{liu2021towards} & 
  \makecell[l]{Emotional Support \\ Conversation} &
  \makecell[l]{Various Mental \\ Disorders } & Crowdsourced &
  Crowdworkers (trained) &
  \makecell[l]{1,053 \\ conversations} &
  Publicly Available \\
Ji et al.\cite{ji2022suicidal} & 
  \makecell[l]{Suicidal Ideation and \\ Mental Disorder \\ Detection} &
  \makecell[l]{Suicidal Ideation, \\ Depression, Anxiety, \\ Bipolar} &
  \makecell[l]{Reddit (SuicideWatch \\ and Mental 
 \\ Health-related \\ subreddits)} &
  Experts &
  4800 posts &
  \makecell[l]{Available Upon \\ Request} \\
Cascalheira et al. \cite{cascalheira2023models} & 
  \makecell[l]{Gender Dysphoria \\ Identification} &
  Gender Dysphoria &
  Reddit (r/GenderDysphoria) &
  \makecell[l]{Clinicians and Students \\ with transgender and \\ non-binary experience} &
  1,573 posts &
  \makecell[l]{Available Upon \\ Request} \\
Hua et al. \cite{hua2023deep} &
  \makecell[l]{Transgender and \\ Gender Diverse \\ Patient Identification} &
  \makecell[l]{Transgender and \\ Gender Diverse} &
  \gls{EHR}s &
  Clinicians &
  3,350 patients  &
  \makecell[l]{Available upon request \\ pending institutional \\ reviews and approvals} \\
Mukherjee et al. \cite{mukherjee2023early} &
  \makecell[l]{Autism Spectrum \\ Disorder Symptom \\ Identification} &
  \makecell[l]{Autism Spectrum \\ Disorder} &
  \makecell[l]{Social networks (X), \\ special children's \\ organizations} &
  Researchers &
  Not Specified &
  Not mentioned \\
DATD \cite{owen2020towards} &
  \makecell[l]{Depression and \\ Anxiety Identification} &
  \makecell[l]{Depression and \\ Anxiety} &
  X &
  Not Specified &
  1,050 tweets &
  Publicly Available \\
PAN12 \cite{inches2012overview} & 
  \makecell[l]{Sexual Predator \\ Identification}	 &
  Online Grooming	 &
  \makecell[l]{Perverted Justice \\ Website, Omegle, \\ IRC logs}	 &
  One Expert  &
  Hundreds of thousands of conversations  &
  Publicly Available \\
PJZC \cite{milon2022take} & 
  \makecell[l]{Grooming \\ Identification}	 &
  Online Grooming	 &
  \makecell[l]{Perverted Justice \\ Website, \\ \#ZIG IRC Channel, \\ Chit-Chat Dataset} &
  Researchers &
  21,070 chats &
  Publicly Available \\
DAIC-WOZ \cite{gratch2014distress} & 
  \makecell[l]{Distress \\ Identification} &
  \makecell[l]{Depression, \\ PTSD, Anxiety} &
  Interviews & 
  Experts &
  \makecell[l]{421 \\ interviews} &
  \makecell[l]{Available Upon \\ Request} \\
D4 \cite{yao2022d4} & 
  \makecell[l]{Response Generation, \\ Topic Prediction, \\ Dialogue Summary, \\ Severity Classification} &
  \makecell[l]{Depression, \\ Suicide Risk} &
  \makecell[l]{Simulated \\ conversations based \\ on real patient \\ portraits} &
  \makecell[l]{Psychiatrists, \\ Crowdworkers} &
  \makecell[l]{1,339 \\ conversations} &
  \makecell[l]{Available Upon \\ Request} \\
DialogueSafety \cite{qiu2023benchmark}  & 
  \makecell[l]{Dialogue Safety \\ Classification} &
  \makecell[l]{Various (Nonsense, \\ Humanoid Mimicry, \\ Linguistic Neglect, \\ etc.)} & 
  \makecell[l]{Chinese Online \\ Counseling Platform, \\ Yixinli QA Column} &
  \makecell[l]{Experienced \\ Supporters, \\ Counseling \\ Psychologists, \\ Linguistic Experts} &
  \makecell[l]{7,935 \\ dialogues} &
  \makecell[l]{Available Upon \\ Request} \\
CLPsych15 \cite{coppersmith2015clpsych} & 
  \makecell[l]{Depression and \\ PTSD \\ Identification} &
  \makecell[l]{Depression, \\ PTSD} &
  X &
  Experts &
  \makecell[l]{1,746 tweets} &
  \makecell[l]{Available Upon \\ Request pending \\ institutional reviews \\ and approvals}  \\
RED SAM \cite{sampath2022data, kayalvizhi2022findings} & 
  \makecell[l]{Depression Level \\ Identification} &
  Depression (3 levels) &
  Reddit &
  Crowdworkers  &
  16,632 posts  &
  Publicly Available \\
  \bottomrule
\end{longtable}
\end{landscape}